\documentclass[10pt,twocolumn,letterpaper]{article}

\usepackage{cvpr}
\usepackage{times}
\usepackage{epsfig}
\usepackage{graphicx}
\usepackage{amsmath}
\usepackage{amssymb}

\usepackage{subcaption}
\usepackage{float}

\usepackage[pagebackref=true,breaklinks=true,letterpaper=true,colorlinks,bookmarks=false]{hyperref}

\newcommand{\myparagraph}[1]{\par\vspace{5pt}\noindent\textbf{#1} }

\newcommand\blfootnote[1]{%
  \begingroup
  \renewcommand\thefootnote{}\footnote{#1}%
  \addtocounter{footnote}{-1}%
  \endgroup
}

\cvprfinalcopy 

\def\cvprPaperID{1542} 

\ifcvprfinal\fi
\begin{document}

\title{Cross-modal Deep Face Normals with Deactivable Skip Connections}
\author{Victoria Fern\'andez Abrevaya$^{*1}$, Adnane Boukhayma$^{*\dagger 2}$, Philip H. S. Torr$^3$, Edmond Boyer$^1$ \\
$^1$ Inria, Univ. Grenoble Alpes, CNRS, Grenoble INP, LJK, France\\
$^2$ Inria, Univ. Rennes, CNRS, IRISA, M2S, France\\ 
$^3$ University of Oxford, UK\\
{\tt\small \{victoria.fernandez-abrevaya, adnane.boukhayma, edmond.boyer\}@inria.fr}\\
{\tt\small philip.torr@eng.ox.ac.uk}
}

\maketitle
\ifcvprfinal\thispagestyle{empty}\fi


%


\begin{abstract}
\blfootnote{* Authors contributed equally}
\blfootnote{$\dagger$ This work was done while the author was at University of Oxford}
We present an approach for estimating surface normals from in-the-wild color images of faces. While data-driven strategies have been proposed for single face images,  limited available ground truth data makes this problem difficult. To alleviate this issue, we propose a method that can leverage all available image and normal data, whether paired or not,  thanks to a novel cross-modal learning architecture. In particular, we enable additional training with single modality data, either color or normal, by using two encoder-decoder networks with a shared latent space. The proposed architecture also enables face details to be transferred between the image and normal domains, given paired data, through skip connections between the image encoder and normal decoder. Core to our approach is a novel module that we call deactivable skip connections, which allows integrating both the auto-encoded  and image-to-normal branches within the same architecture that can be trained end-to-end. This allows learning of a rich latent space that can accurately capture the normal information.
We compare against state-of-the-art methods and show that our approach can achieve significant improvements, both quantitative and qualitative, with natural face images.
\end{abstract}

\section{Introduction}

\begin{figure}[t]
\centering
\begin{tabular}{c c l}
\includegraphics[height=2.2cm]{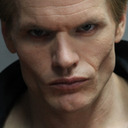} &
\includegraphics[height=2.2cm]{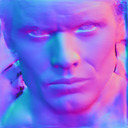} &
\includegraphics[height=2.2cm,trim={2.5cm 1cm 1.5cm 0.0125cm},clip]{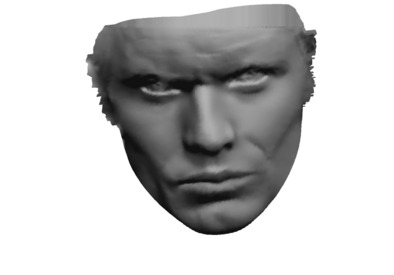} \\
\includegraphics[height=2.2cm]{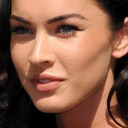} &
\includegraphics[height=2.2cm]{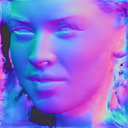} &
\includegraphics[height=2.2cm,trim={1.5cm 0.925cm 1.5cm 0.225cm},clip]{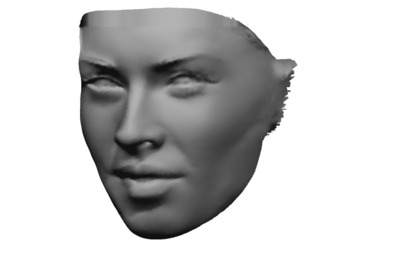} \\
\includegraphics[height=2.2cm]{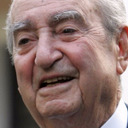} &
\includegraphics[height=2.2cm]{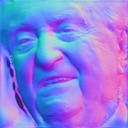} &
\includegraphics[height=2.2cm,trim={1.75cm 0.4cm 1.5cm 0.5cm},clip]{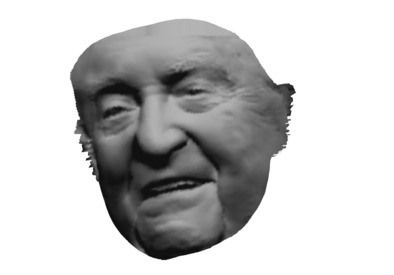}
\end{tabular}
\caption{Our model predicts accurate normals from a single input image that can be used to enhance a coarse geometry (e.g. PRN~\cite{feng18}).}
\end{figure}%


3D reconstruction of the human face  is a long-standing problem in computer vision, with a wide range of applications including biometrics, forensics, animation, gaming, and human digitalization. In many of these applications monocular inputs are considered in order to limit the acquisition constraints, hence enabling uncontrolled environments as well as  efficient information usage for \eg facial telecommunication and entertainment. Although significant  progress has been recently made by the scientific community, recovering detailed 3D face models given only single images is still an open problem. 

\smallskip
Monocular face reconstruction is in essence an ill-posed problem which requires strong prior knowledge. Assuming a simple shading model, seminal shape-from-shading (SfS) approaches~\cite{Horn89, Zhang99}  were estimating shape normals by considering local pixel intensity variations. Fine scale surface details can be recovered using this strategy, however the applicability to in-the-wild images is limited by the simplified image formation model that is assumed. Later on,  a more global strategy was proposed with parametric face models~\cite{Blanz99, Vlasic05}. They allow fitting a template face controlled by only a few coefficients, resulting hence in improved robustness. While being widely adopted, parametric models are inherently restricted in expressiveness and have difficulties in recovering small surface details, as a consequence of their low dimensional representation. Recently, deep learning methods that exploit large-scale face image datasets have been investigated with the aim of a better generalization. While most work in this category are trained to estimate the coefficients of a parametric model~\cite{Tewari17, Tewari18, Genova18, Kim18, Sanyal19}, a few other approaches infer directly per-pixel depth~\cite{Sela17}, UV position maps~\cite{feng18} or surface normals~\cite{Trigeorgis17, Sengupta18}.

As observed in previous work~\cite{Smith19, Zhang18}, regressing depth information alone can lead to suboptimal results, especially detail-wise, as the inherent scale ambiguity with single images can make convergence difficult for neural networks. On the other hand, the estimation of normals  appears to be an easier task for such networks, given that normals are strongly correlated to pixel intensities and depend mostly on local information, a fact already exploited by SfS techniques. Still, only a few approaches have been proposed in this line for facial images~\cite{Shu17, Sengupta18}, mostly due to the limited available ground-truth data.  We propose here a method that overcomes this limitation and can leverage all data available  through the use of cross-modal learning. Our experiments demonstrate that this strategy can estimate more accurate and sharper facial surface normals from single images.

\medskip
The proposed approach recovers accurate normals corresponding to the facial region within  an RGB image, with the goal of enhancing an existing coarse reconstruction, \cite{feng18} in our experiments. 
We cast the problem as a color-to-normal image translation, which can be in principle solved by combining an image encoder $E_I$ with a normal decoder $D_N$ as in~\cite{Trigeorgis17}, and including skip connections between $E_I$ and $D_N$~\cite{Ronneberger15}  in order to transfer details from the image domain to the normals domain. 
However, training such a network can prove difficult unless a large dataset of image/normal pairs, that ideally contains images in-the-wild, is available. In practice few such datasets are currently publicly available, \eg~\cite{Zafeiriou11}, which were  moreover captured under controlled conditions. To improve generalization, we propose to augment the architecture with a normal encoder $E_N$ and an image decoder $D_I$, where all encoders/decoders share the same latent space. This augmented architecture provides additional constraints on the latent space with the auto-encoded image-to-image and normal-to-normal branches, allowing therefore for a much wider range of training datasets.  In order to keep advantage of the skip connections between $E_I$ and $D_N$, while avoiding the resulting bonded connections between $E_N$ with $D_N$ that hamper the architecture, we introduce the \emph{deactivable skip connections}.  This allows skip connections to be turned on and off during training according to the type of data.

\medskip
In summary, this work contributes (1) a framework that leverages cross-modal learning for the estimation of normals from a single face image in-the-wild; (2) the introduction of the \emph{deactivable skip connection}; and (3) an extensive evaluation that shows that our approach outperforms state-of-the-art methods on the Photoface~\cite{Zafeiriou11} and Florence~\cite{Bagdanov11} datasets, with up to nearly $10\%$ improvements in angular error on the Florence dataset, as well as visually compelling reconstructions.
\section{Related Work}

We focus the discussion below on methods that consider 3D face reconstruction, or normal estimation,  given single RGB images.

\myparagraph{Reconstruction with Parametric Models} 3D reconstruction from a single image is  ill-posed and many methods resort therefore to strong priors with parametric face models such as blendshape~\cite{Garrido13,  Cao15,Thomas16} or statistical models, typically the 3D Morphable Model (3DMM)~\cite{Blanz99}.  These models are commonly used within an analysis-by-synthesis optimization~\cite{Romdhani05, Huber16, Egger18, Booth18, Gecer19} or, more recently, using deep learning to regress model parameters~\cite{Richardson16, Richardson17, Tewari17, Tuan17, Genova18, feng18, Kim18, Tewari19, Sanyal19}, or alternatively to regress other face information using 3DMM training data, for instance   volumetric information~\cite{Jackson17}, UV position map~\cite{feng18}, normal map~\cite{Trigeorgis17}, depth map~\cite{Sela17}, or the full image decomposition~\cite{Shu17, Sengupta18, Kim18b}. This strategy has proven robustness, however it is constrained by the parametric representation that offers limited expressiveness and fails  in recovering fine scale details. 

In order to improve the quality of the reconstructions several works have proposed to add medium-scale correctives on top of the parametric model~\cite{Li13, Garrido16, Tewari18}, to train a local wrinkle regressor~\cite{Cao15}, or to learn deep non-linear 3DMM~\cite{Tran19, Zhou19} that can capture higher-frequency details. Our method also enables to enhance a face prediction through the estimation of more accurate normals. 

\myparagraph{Normal Estimation with Shape from Shading} Shape-from-shading (SfS)~\cite{Horn89, Zhang99} is a well-studied technique that aims at  recovering detailed 3D surfaces from a single image based on shading cues. It estimates surface normals using the image irradiance equation, as well as illumination model parameters when these are unknown. SfS is inherently limited by the simplified image formation model assumed but has inspired numerous works that build on the correlation between pixel intensity and normals, either explicitly or implicitly.  For instance, a  few works on faces combined SfS with a data-driven model, \eg~\cite{Smith06, Kemelmacher10, Snape14}, which helps to avoid some of the limitations such as ill-posedeness and ambiguities \eg~\cite{Belhumeur99}.  The recent works of Shu \etal~\cite{Shu17} and Sengupta \etal~\cite{Sengupta18} use deep neural networks to decompose in-the-wild facial images into surface normals, albedo and shading, assuming Lambertian reflectance and using a semi-supervised learning approach inspired by SfS. Our work follows a similar direction and  estimates the normal information from a single image, but unlike~\cite{Shu17, Sengupta18} we do not rely on an image formation model and let instead the network learn such a transformation from real data.

\myparagraph{Normal Estimation with Deep Networks} Closely related to our work are methods that recover surface normals from an image using deep neural networks, \eg~\cite{Wang15, Eigen15, Yoon16, Kokkinos17, Bednarik18, Qi18, Zhang18, Qiu19, Du19, Smith19, Alldieck19}. Yoon \etal~\cite{Yoon16} and Bansal \etal~\cite{bansal16} focus on the normal prediction task in order to recover detailed surfaces. Eigen and Fergus~\cite{Eigen15} simultaneously regress depth, normal and semantic segmentation using a multi-scale approach. Zhang and Funkhouser~\cite{Zhang18} predict surface normal and occlusion boundaries to later optimize for depth completion; a similar direction was followed by~\cite{Qiu19} for outdoor scenes. Trigeorgis \etal~\cite{Trigeorgis17} estimate facial normals with a supervised approach trained on synthetic data. Our approach differs from the aforementioned methods with a new architecture that enables cross-modal learning, hence improving performances in monocular 3D face normal estimation.  

\myparagraph{Geometry Enhancement using Deep Networks}
Methods have been proposed  that directly  enhance face models using deep neural networks. Richardson~\etal~\cite{Richardson17} use two networks where the first estimates a coarse shape and the second one refines the depth map from the previous branch, using an SfS-inspired unsupervised loss function. Sela \etal~\cite{Sela17} recover the depth and correspondence maps coupled with an off-line refinement step. The works of~\cite{Yamaguchi18, Huynh18} estimate high frequency details by training with very accurate ground-truth data, which requires a careful acquisition process and high-quality inputs. Tran \etal~\cite{Tran18} estimate a per-pixel bump map, where the ground-truth data is obtained by applying an SfS method offline. The work of~\cite{Chen19} 
learns to estimate a geometric proxy and a displacement map for details primarily for high resolution images ($2048 \times 2048$). While they mention limitations with low resolution images, we show results with resolutions as low as $256 \times 256$. 
\section{Method}

\begin{figure}[t]
\includegraphics[width=0.7\linewidth]{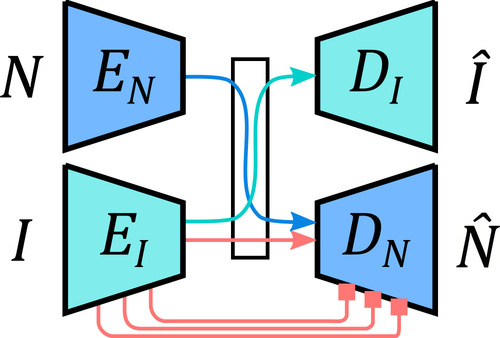}
\centering
\caption{Overview of the proposed approach. Our cross-modal architecture allows exploitation of paired and unpaired image/normal data for image-to-normal translation (Red), by means of further image-to-image (Green) and normal-to-normal (Blue) regularizations during training. The \textit{deactivable skip connections} allow to transfer details from the image encoder $E_I$ to the normal decoder $D_N$ without having to link the normal encoder $E_N$ to the normal decoder $D_N$.}
\label{fig:pipe}
\end{figure}

We propose to predict face normals from a single color image using a deep convolutional encoder-decoder network. A natural solution to this purpose is to combine an image encoder $E_I$ with a normal decoder $D_N$, as in \eg~\cite{Trigeorgis17}. However training such an architecture requires  pairs of normal and color images in correspondence. Although  a few public datasets are available that contain high-quality 3D or normal ground-truth information for faces, for instance ICT-3DRFE~\cite{stratou11} or Photoface~\cite{Zafeiriou11}, they were obtained under controlled conditions and do not, therefore,  really cover the distribution of images in-the-wild. On the other hand, numerous large datasets of natural images are publicly available, for example CelebA~\cite{liu15} and AffectNet~\cite{mollahosseini17}, yet without the associated accurate and detailed ground-truth normal values. Whereas other works have approached this by augmenting the training corpus with synthetic ground-truth~\cite{Trigeorgis17, Sengupta18}, we  propose instead a method based on cross-modal learning that can leverage all available data, even unpaired. 

\subsection{Cross-modal Architecture}
\label{sec:cross-modal}

As depicted in Fig.~\ref{fig:pipe}, we use two encoder/decoder networks, one for images $E_I / D_I$ and one for normals $E_N/D_N$, sharing the same latent space. This architecture is trained with image-to-image, normal-to-normal and image-to-normal supervision simultaneously in order to obtain a rich and robust latent representation. To this purpose,  we exploit paired images of normal and color information on faces, as available from~\cite{stratou11,Zafeiriou11}, in addition to individual images of either color or normal information, from \eg CelebA-HQ \cite{Karras17}  and  BJUT-3D~\cite{bjut}. To improve the overall performance we augment this architecture with long skip connections between $E_I$ and $D_N$, as it favors the transfer of details between   the image and normal domains, and since it has been shown to significantly increase performance in several image translation tasks  \eg~\cite{Isola17}. In practice we use a U-Net+ResNet ~\cite{Ronneberger15,he16} architecture that combines the benefits of both short and long skip connections.

Training such an architecture end-to-end raises an obstacle: the skip connections from $E_I$ to $D_N$ ($E_I \rightarrow D_N$), which are based on concatenating feature maps, impose, by construction, to also have skip connections between the encoder and decoder of the normal modality, i.e. $E_N \rightarrow D_N$. This is counterproductive in practice: by setting skip connections within the same modality, it is in fact easier for the normal autoencoder to transfer features from the earliest layers of its encoder to the last layers of its decoder through the skip connection, thus depriving the deeper layers of any meaningful gradients during training. Not only will this fail to improve the latent face representation, but it will also alter the coefficients of the normal decoder for the image-to-normal inference task.

For this reason, we introduce the \emph{deactivable skip connections} as shown in Fig.~\ref{fig:skip} and detailed in Sec.~\ref{sec:method_arch}. This allows us to train the framework end-to-end by setting long connections solely between $E_I$ and $D_N$, thus learning a rich latent space that encodes facial features from both color and normal images while profiting from all available data.

\subsection{Deactivable Skip Connections}
\label{sec:method_arch}

\begin{figure}[t]

\begin{subfigure}{\linewidth}
\includegraphics[height=2.2cm]{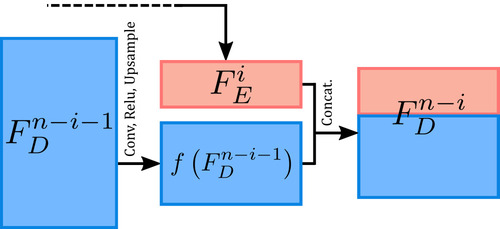}
\centering
\caption{Standard skip connection}
\label{fig:skip1}
\end{subfigure}
\begin{subfigure}{\linewidth}
\includegraphics[height=2.2cm]{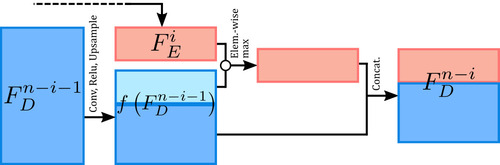}
\centering
\caption{Deactivable skip connection}
\label{fig:skip2}
\end{subfigure}

\caption{Instead of concatenating the encoder features (red) and decoder features (blue), as with standard skip connections, we fuse the encoder features  with part of the decoder features (light blue), to be able to deactivate this operation when needed.}
\label{fig:skip}
\end{figure}

As mentioned earlier, skip connections are well suited to our problem as they allow sharing of low-level information at multiple scales while still preserving the general structure. In the implementation of standard skip connections, as in~\cite{Ronneberger15,Isola17}, the decoder features at the $({n-i})^{th}$ layer $F_{D}^{n-i}$ are the concatenation of the processed previous layer features $f(F_{D}^{n-i-1})$ and the encoder features at layer $i$, $F_{E}^{i}$, where $n$ is the total number of layers (see Fig.\ref{fig:skip1}).

Let $m\left(F_{E_I}^{i}\right)$ be the number of feature maps at the $i^{th}$ layer of $E_I$. The proposed architecture (Fig.~\ref{fig:pipe}) requires to set connections from the image encoder $E_I$ to the normal decoder $D_N$, and as a consequence, each layer features $F_{D_N}^{n-i}$ of $D_N$ are expected to always have an additional $m\left(F_{E_I}^{i}\right)$ channels. In order to gain generalization over each domain,  both the color and the normal images can be auto-encoded during training. However, since the concatenation is expected during training on the decoder $D_N$ side, features of the normal encoder $E_N$ must be concatenated as well, which is as discussed detrimental to our model.

The \emph{Deactivable Skip Connections} are designed such that, during training, the transfer of feature maps from encoders to decoders can be selectively activated or deactivated.
Compared to a decoder equipped with standard skip connections, the processed features $f\left(F_{D}^{n-i-1}\right)$ of our decoder include $m\left(F_{E}^{i}\right)$ extra channels (light blue in Fig.\ref{fig:skip2}). During a normal-to-normal pass, the skip connections are deactivated and the $(n-i)^{th}$ layer features of the normal decoder correspond to the processed previous layer features \eg $F_{D}^{n-i}=f\left(F_{D}^{n-i-1}\right)$. 
During an image-to-normal pass, the skip connection is activated: we first perform an element-wise max-pooling between the $i^{th}$ layer features of the encoder $F_{E}^{i}$ and the last $m\left(F_{E}^{i}\right)$ channels of the processed $(n-i-1)^{th}$ layer features of the decoder $f\left(F_{D}^{n-i-1}\right)$, as illustrated in Fig.~\ref{fig:skip2}. The result is stacked back with the remaining of the processed previous layer features thus forming the final $(n-i)^{th}$ decoder layer features $F_{D}^{n-i}$. Doing so allows to transfer the information from encoder to decoder without degrading performances when the transfer operation does not occur, as when auto-encoding normals.

\subsection{Training}

We train the framework end-to-end using both supervised and unsupervised data, where the latter includes individual image and normal datasets. During  training, the skip connections are deactivated when doing a normal-to-normal pass. For the supervised case, and for unsupervised normals, the loss function is the cosine distance between the output and the ground-truth, which in our experiments gave better results than the $L1/L2$ norm:

\begin{equation}
\mathcal{L}_{nrm}(N,\hat{N}) = 1 - \frac{1}{|N|} \sum_{(i,j)} \frac{N(i,j)^\top \cdot \hat{N}(i,j)}{||N(i,j)||_2 ||\hat{N}(i,j)||_2},
\end{equation}

\noindent where $N(i,j)$ and $\hat{N}(i,j)$ are the normal vectors at pixel $(i,j)$ in the ground-truth and output normal images $N$ and $\hat{N}$ respectively, and $|N|$ is the number of pixels in $N$. For unsupervised image data we use the $L2$ loss:
\begin{equation}
\mathcal{L}_{img}(I,\hat{I}) = || I - \hat{I} ||^2_2,
\end{equation}

\noindent where $\hat{I}$ is the output color image and $I$ the ground-truth. In both cases, the loss is applied only on facial regions segmented using  masks obtained as explained in Sec.\ref{sec:impl}. 

In practice, as we can only perform a training iteration for one input modality at a time, either an input batch of images or normals, we train our model as follows: when loading a batch of images with image/normal ground-truth pairs, we perform a normal-to-normal iteration first, followed by an image-to-normal plus image-to-image iteration, where both losses in the latter iteration are summed with equal weights.
When loading a batch of images only, we perform an image-to-image iteration. Finally,  with a batch of normals only, we naturally proceed with a normal-to-normal iteration alone.

\section{Evaluation}
\label{sec:results}

\newcommand{\sza}{0.10} 
\newcommand{\szadouble}{0.2} 
\newcommand{\szb}{3}
\newcommand{\szc}{0.55}



We report below on the accuracy of the normals estimated with our approach on standard datasets~\cite{Zafeiriou11, Bagdanov11}. We compare against state-of-the-art methods on normal estimation and 3D reconstruction, and show significant improvements in terms of normal prediction accuracy. This is supported by compelling reconstructions of images in-the-wild from 300-W~\cite{Sagonas13}, as can be seen in Figs.~\ref{fig:comparison_norm} and ~\ref{fig:comparison_mesh}.

Following previous works~\cite{Sengupta18, Trigeorgis17}, we evaluate with the mean angular error between the output and the ground-truth normals, as well as percentage of pixels within the facial region with an angular error of less than $20^\circ$, $25^\circ$ and $30^\circ$. For qualitative comparisons we show both the output normal map, as well as the mesh results obtained by enhancing the output of PRN~\cite{feng18} using normal mapping~\cite{Cohen98}: we append the predicted normals to the the PRN meshes pixel-wise
thus rendering enhanced geometric shading.

\subsection{Implementation Details}
\label{sec:impl}
The framework was implemented in PyTorch~\cite{pytorch}, and all experiments were run on a  GTX TITAN Black. The networks were trained for $40$ epochs using ADAM solver~\cite{kingma14} with a learning rate of $10^{-4}$. We use a ResNet-18~\cite{he16} architecture and set five skip connections, one at the output of the initial layer and the rest at the output of each of the four residual blocks. Each mini-batch during training consists of data of the same type, \ie images only, normals only or image-normal pairs, as this worked best for us empirically.

Similar to prior work, input images are crops of fixed size around the face. We extract 2D keypoints with a face detector~\cite{dlib} and create masks on the facial region by finding the tightest square of edge size $l$ around the convex hull of the points. The images are then cropped with a square patch of size $1.2\times l$ centered at the same 2D location as the previously detected box, and subsequently resized to $256 \times 256$. The code will be made publicly available.

\subsection{Datasets}

Our training set comprises multiple datasets: ICT-3DRFE~\cite{stratou11} and Photoface~\cite{Zafeiriou11} which provide image/normal pairs, CelebA-HQ \cite{Karras17} which only contains 2D images, and BJUT-3D~\cite{bjut}, which consists of high-quality 3D scans.

We generated $8625$ image/normal pairs from ICT-3DRFE by randomly rotating the $345$ 3D models and relighting them using the provided albedos. We sampled random rotation axes and angles in $[-\pi/4,\pi/4]$, random lighting directions with positive $z$, and random intensities in $[0,2]$. For Photoface, following the setting in~\cite{Trigeorgis17,Sengupta18}, we randomly selected a training subset of $353$ people resulting in $9478$ image/normal pairs. We also generated $5000$ high resolution facial images from CelebA-HQ, which is used to train the image-to-image branch exclusively. In addition, we render $3000$ normal images from the $500$ scans of BJUT-3D, rotated with random axes and angles in $[-\pi/4,\pi/4]$. We only render normal images from this dataset as the original scan color images are not provided.

For evaluation purposes we use the remaining testing subset of Photoface, which consists of $100$ subjects not seen during training and $1489$ image/normal pairs. This subset challenges the reconstruction with very severe lighting conditions. Following the work of~\cite{feng18}, we create an additional evaluation set by rendering $530$ color and normal facial images from the $53$ 3D models of the Florence dataset~\cite{Bagdanov11}, rotated with random axes and angles in $[-\pi/4,\pi/4]$. This allows to evaluate on a completely unseen dataset. Finally, we use the 300-W dataset~\cite{Sagonas13} of 2D face images to assess qualitative performances in-the-wild. Note that for both training and testing, we limited ourselves to 3D face datasets of high quality and details.

\begin{table}[h!]
\begin{center}

\setlength{\tabcolsep}{0.5em}
\begin{tabular}{ c|c|c|c|c| } 
\cline{2-5}
& Mean$\pm$std & $<20^\text{o}$ & $<25^\text{o}$ & $<30^\text{o}$  \\
\hline
\multicolumn{1}{|c|}{Pix2V \cite{Sela17}}  & 33.9$\pm$5.6 & 24.8\% & 36.1\% & 47.6\% \\
\multicolumn{1}{|c|}{Extreme \cite{Tran18}} & 27.0$\pm$6.4 & 37.8\% & 51.9\% & 64.5\% \\
\multicolumn{1}{|c|}{3DMM \cite{Trigeorgis17}} & 26.3$\pm$10.2 & 4.3\% & 56.1\% & {\bf 89.4}\% \\

\multicolumn{1}{|c|}{3DDFA \cite{Zhu17}}  & 26.0$\pm$7.2 & 40.6\% & 54.6\% & 66.4\% \\
\multicolumn{1}{|c|}{SfSNet \cite{Sengupta18}}  & 25.5$\pm$9.3 & 43.6\% & 57.5\% & 68.7\% \\
\multicolumn{1}{|c|}{PRN \cite{feng18}}  & 24.8$\pm$6.8 & 43.1\% & 57.4\% & 69.4\% \\
\multicolumn{1}{|c|}{Ours} & {\bf 22.8$\pm$6.5} & {\bf 49.0\%} & {\bf 62.9\%} & 74.1\% \\

\hline
\multicolumn{1}{|c|}{UberNet \cite{Kokkinos17}} & 29.1$\pm$11.5 & 30.8\% & 36.5\% & 55.2\% \\
\multicolumn{1}{|c|}{NiW \cite{Trigeorgis17}} & 22.0$\pm$6.3 & 36.6\% & 59.8\% & 79.6\%\\
\multicolumn{1}{|c|}{Marr Rev \cite{bansal16}} & 28.3$\pm$10.1 & 31.8\% & 36.5\% & 44.4\% \\
\multicolumn{1}{|c|}{SfSNet-ft \cite{Sengupta18}} & 12.8$\pm$5.4 & 83.7\% & 90.8\% & 94.5\%\\
\multicolumn{1}{|c|}{Ours-ft} & {\bf 12.0$\pm$5.3} & {\bf 85.2\%} & {\bf 92.0\%}& {\bf 95.6\%} \\
\hline

\end{tabular}
\end{center}
\vspace{-5pt}
\caption{Quantitative comparisons on the Photoface dataset~\cite{Zafeiriou11} with mean angular errors (degrees) and percentage of errors below $20^\circ$, $25^\circ$ and $30^\circ$. \texttt{-ft} means that the method was fine-tuned on Photoface.}
\label{tab:quant_ph}
\end{table}

\begin{table}[h!]
\begin{center}

\setlength{\tabcolsep}{0.5em}
\begin{tabular}{ c|c|c|c|c| } 
 \cline{2-5}
    & Mean$\pm$std & $<20^\text{o}$ & $<25^\text{o}$ & $<30^\text{o}$  \\
  \hline 
\multicolumn{1}{|c|}{Extreme \cite{Tran18}} & 19.2$\pm$2.2 & 64.7\% & 75.9\% & 83.3\% \\
\multicolumn{1}{|c|}{SfSNet \cite{Sengupta18}} & 18.7$\pm$3.2 & 63.1\% & 77.2\% & 86.7\% \\
\multicolumn{1}{|c|}{3DDFA \cite{Zhu17}}  & 14.3$\pm$2.3 & 79.7\% & 87.3\% & 91.8\% \\
\multicolumn{1}{|c|}{PRN \cite{feng18}}  & 14.1$\pm$2.16 & 79.9\% & 88.2\% & 92.9\% \\
\multicolumn{1}{|c|}{Ours} & {\bf 11.3$\pm$1.5} & {\bf 89.3\%} & {\bf 94.6\%} & {\bf 96.9\%} \\

\hline
\end{tabular}
\end{center}
\vspace{-5pt}
\caption{Quantitative comparisons on the Florence dataset~\cite{Bagdanov11} with mean angular errors (degrees) and percentage of errors below $20^\circ$, $25^\circ$ and $30^\circ$.}
\label{tab:quant_fl}
\end{table}

\subsection{Comparisons}

\begin{figure*}[h!]
\centering

\begin{subfigure}{\sza\linewidth}
\includegraphics[width=\linewidth]{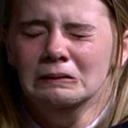}
\centering
\end{subfigure}
\begin{subfigure}{\sza\linewidth}
\includegraphics[width=\linewidth]{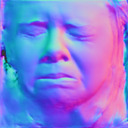}
\centering
\end{subfigure}
\begin{subfigure}{\sza\linewidth}
\includegraphics[width=\linewidth]{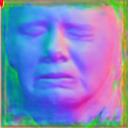}
\centering
\end{subfigure}
\begin{subfigure}{\sza\linewidth}
\includegraphics[width=\linewidth]{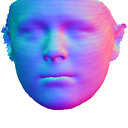}
\centering
\end{subfigure}
\begin{subfigure}{\sza\linewidth}
\includegraphics[width=\linewidth]{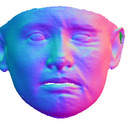}
\centering
\end{subfigure}
\begin{subfigure}{\sza\linewidth}
\includegraphics[width=\linewidth]{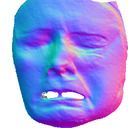}
\centering
\end{subfigure}
\begin{subfigure}{\sza\linewidth}
\includegraphics[width=\linewidth]{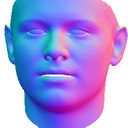}
\centering
\end{subfigure}

\begin{subfigure}{\sza\linewidth}
\includegraphics[width=\linewidth]{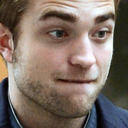}
\centering
\end{subfigure}
\begin{subfigure}{\sza\linewidth}
\includegraphics[width=\linewidth]{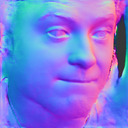}
\centering
\end{subfigure}
\begin{subfigure}{\sza\linewidth}
\includegraphics[width=\linewidth]{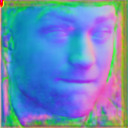}
\centering
\end{subfigure}
\begin{subfigure}{\sza\linewidth}
\includegraphics[width=\linewidth]{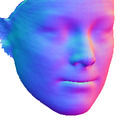}
\centering
\end{subfigure}
\begin{subfigure}{\sza\linewidth}
\includegraphics[width=\linewidth]{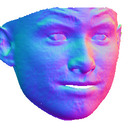}
\centering
\end{subfigure}
\begin{subfigure}{\sza\linewidth}
\includegraphics[width=\linewidth]{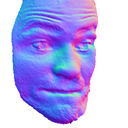}
\centering
\end{subfigure}
\begin{subfigure}{\sza\linewidth}
\includegraphics[width=\linewidth]{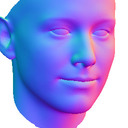}
\centering
\end{subfigure}

\begin{subfigure}{\sza\linewidth}
\includegraphics[width=\linewidth]{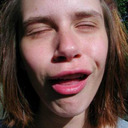}
\centering
\end{subfigure}
\begin{subfigure}{\sza\linewidth}
\includegraphics[width=\linewidth]{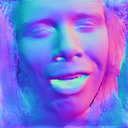}
\centering
\end{subfigure}
\begin{subfigure}{\sza\linewidth}
\includegraphics[width=\linewidth]{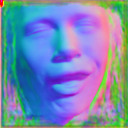}
\centering
\end{subfigure}
\begin{subfigure}{\sza\linewidth}
\includegraphics[width=\linewidth]{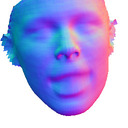}
\centering
\end{subfigure}
\begin{subfigure}{\sza\linewidth}
\includegraphics[width=\linewidth]{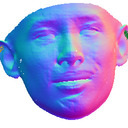}
\centering
\end{subfigure}
\begin{subfigure}{\sza\linewidth}
\includegraphics[width=\linewidth]{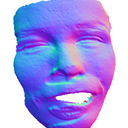}
\centering
\end{subfigure}
\begin{subfigure}{\sza\linewidth}
\includegraphics[width=\linewidth]{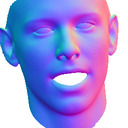}
\centering
\end{subfigure}

\begin{subfigure}{\sza\linewidth}
\includegraphics[width=\linewidth]{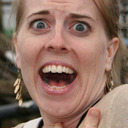}
\centering
\end{subfigure}
\begin{subfigure}{\sza\linewidth}
\includegraphics[width=\linewidth]{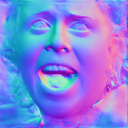}
\centering
\end{subfigure}
\begin{subfigure}{\sza\linewidth}
\includegraphics[width=\linewidth]{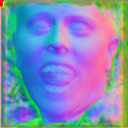}
\centering
\end{subfigure}
\begin{subfigure}{\sza\linewidth}
\includegraphics[width=\linewidth]{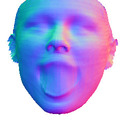}
\centering
\end{subfigure}
\begin{subfigure}{\sza\linewidth}
\includegraphics[width=\linewidth]{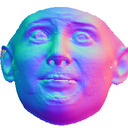}
\centering
\end{subfigure}
\begin{subfigure}{\sza\linewidth}
\includegraphics[width=\linewidth]{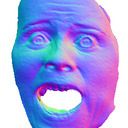}
\centering
\end{subfigure}
\begin{subfigure}{\sza\linewidth}
\includegraphics[width=\linewidth]{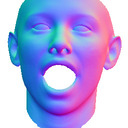}
\centering
\end{subfigure}

\begin{subfigure}{\sza\linewidth}
\includegraphics[width=\linewidth]{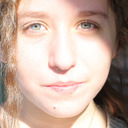}
\centering
\end{subfigure}
\begin{subfigure}{\sza\linewidth}
\includegraphics[width=\linewidth]{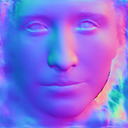}
\centering
\end{subfigure}
\begin{subfigure}{\sza\linewidth}
\includegraphics[width=\linewidth]{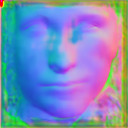}
\centering
\end{subfigure}
\begin{subfigure}{\sza\linewidth}
\includegraphics[width=\linewidth]{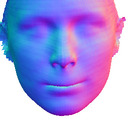}
\centering
\end{subfigure}
\begin{subfigure}{\sza\linewidth}
\includegraphics[width=\linewidth]{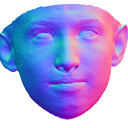}
\centering
\end{subfigure}
\begin{subfigure}{\sza\linewidth}
\includegraphics[width=\linewidth]{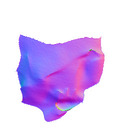}
\centering
\end{subfigure}
\begin{subfigure}{\sza\linewidth}
\includegraphics[width=\linewidth]{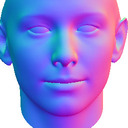}
\centering
\end{subfigure}

\begin{subfigure}{\sza\linewidth}
\includegraphics[width=\linewidth]{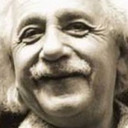}
\centering
\end{subfigure}
\begin{subfigure}{\sza\linewidth}
\includegraphics[width=\linewidth]{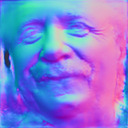}
\centering
\end{subfigure}
\begin{subfigure}{\sza\linewidth}
\includegraphics[width=\linewidth]{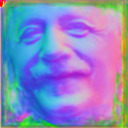}
\centering
\end{subfigure}
\begin{subfigure}{\sza\linewidth}
\includegraphics[width=\linewidth]{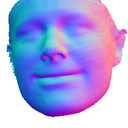}
\centering
\end{subfigure}
\begin{subfigure}{\sza\linewidth}
\includegraphics[width=\linewidth]{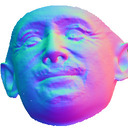}
\centering
\end{subfigure}
\begin{subfigure}{\sza\linewidth}
\includegraphics[width=\linewidth]{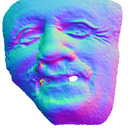}
\centering
\end{subfigure}
\begin{subfigure}{\sza\linewidth}
\includegraphics[width=\linewidth]{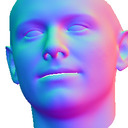}
\centering
\end{subfigure}

\begin{subfigure}{\sza\linewidth}
\includegraphics[width=\linewidth]{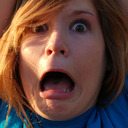}
\centering
\caption{{\footnotesize input}}
\end{subfigure}
\begin{subfigure}{\sza\linewidth}
\includegraphics[width=\linewidth]{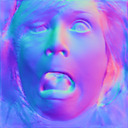}
\centering
\caption{{\footnotesize Ours}}
\end{subfigure}
\begin{subfigure}{\sza\linewidth}
\includegraphics[width=\linewidth]{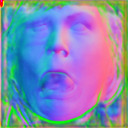}
\centering
\caption{{\footnotesize SfSnet}}
\end{subfigure}
\begin{subfigure}{\sza\linewidth}
\includegraphics[width=\linewidth]{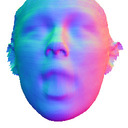}
\centering
\caption{PRN}
\end{subfigure}
\begin{subfigure}{\sza\linewidth}
\includegraphics[width=\linewidth]{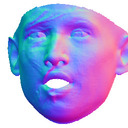}
\centering
\caption{Extreme}
\end{subfigure}
\begin{subfigure}{\sza\linewidth}
\includegraphics[width=\linewidth]{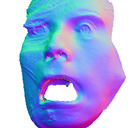}
\centering
\caption{Pix2V}
\end{subfigure}
\begin{subfigure}{\sza\linewidth}
\includegraphics[width=\linewidth]{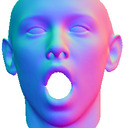}
\centering
\caption{3DDFA}
\end{subfigure}

\caption{Qualitative comparisons on normals in the 300-W dataset~\cite{Sagonas13}.}
\label{fig:comparison_norm}
\end{figure*}

\begin{figure*}[h!]
\centering
\begin{subfigure}[b]{\sza\linewidth}
\includegraphics[height=\szc\textheight]{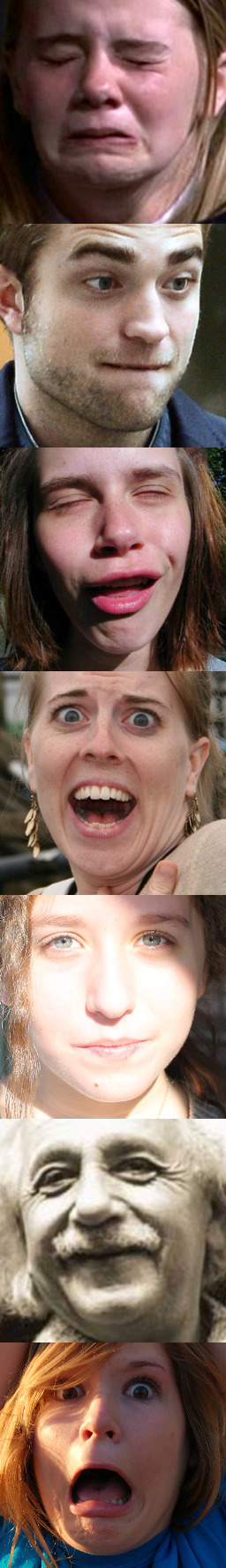}
\centering
\caption{{\scriptsize Input}}
\end{subfigure}
\qquad
\begin{subfigure}[b]{\szadouble\linewidth}
\includegraphics[height=\szc\textheight]{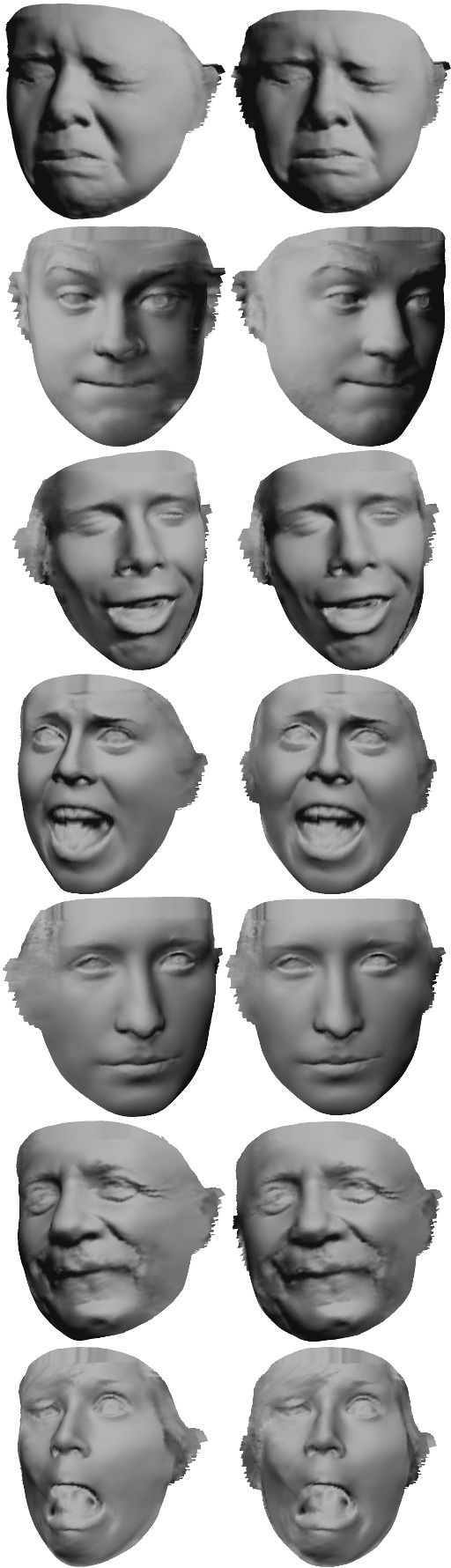}
\centering
\caption{{\scriptsize Ours+PRN}}
\end{subfigure}
\qquad
\begin{subfigure}[b]{\sza\linewidth}
\includegraphics[height=\szc\textheight]{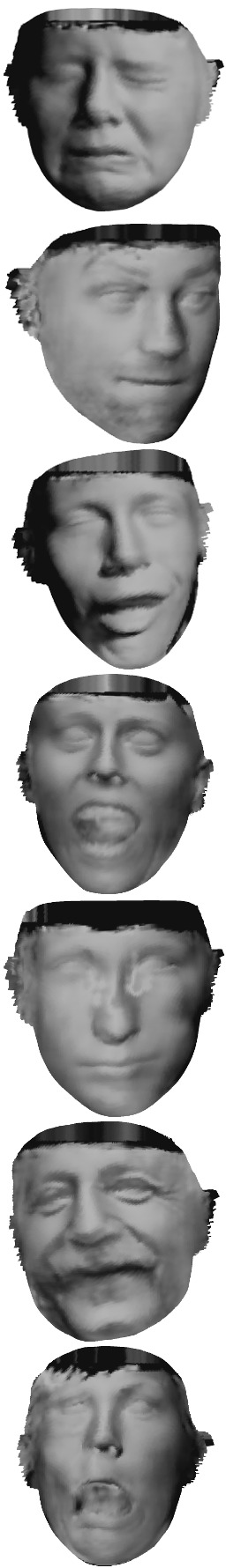}
\centering
\caption{{\scriptsize SfSNet+PRN}}
\end{subfigure}
\quad
\begin{subfigure}[b]{\sza\linewidth}
\includegraphics[height=\szc\textheight]{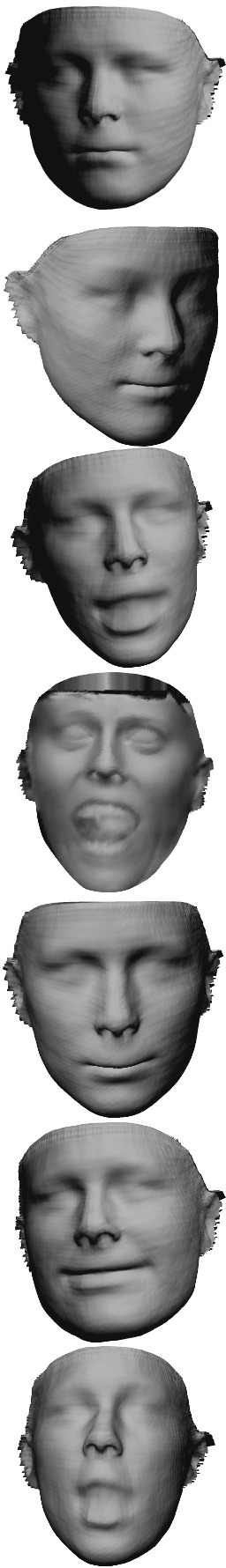}
\centering
\caption{{\scriptsize PRN}}
\end{subfigure}
\quad
\begin{subfigure}[b]{\sza\linewidth}
\includegraphics[height=\szc\textheight]{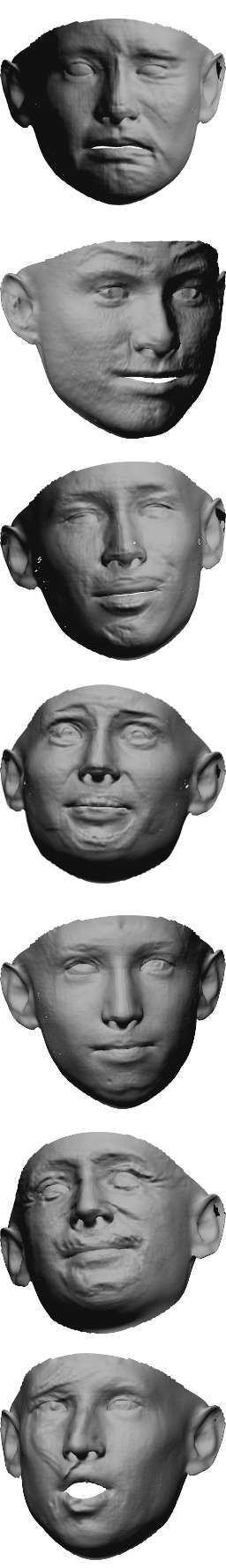}
\centering
\caption{{\scriptsize Extreme}}
\end{subfigure}
\quad
\begin{subfigure}[b]{\sza\linewidth}
\includegraphics[height=\szc\textheight]{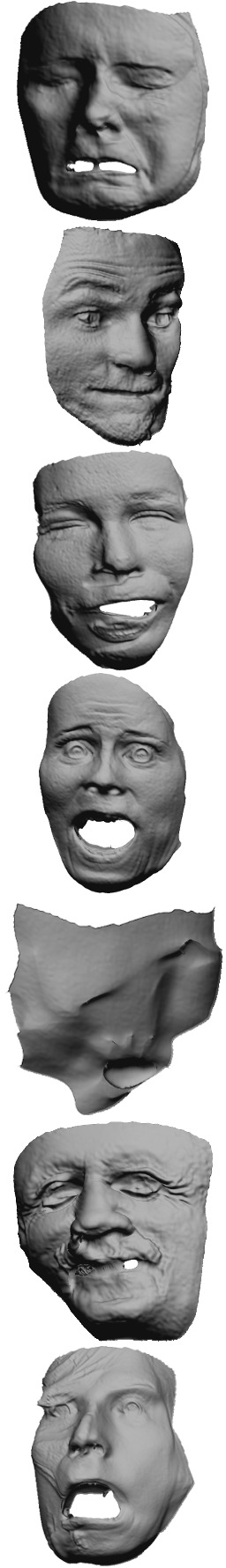}
\centering
\caption{{\scriptsize Pix2Vertex}}
\end{subfigure}
\quad
\begin{subfigure}[b]{\sza\linewidth}
\includegraphics[height=\szc\textheight]{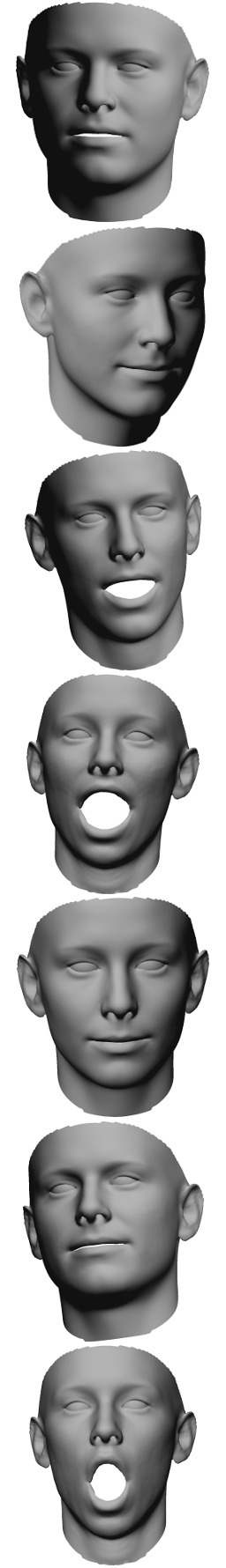}
\centering
\caption{{\scriptsize 3DDFA}}
\end{subfigure}

\caption{Qualitative comparisons on geometries in the 300-W dataset~\cite{Sagonas13}.}
\label{fig:comparison_mesh}
\end{figure*}

%
%
%

We compare our results to methods that explicitly recover surface normals, either for facial images (SfSNet~\cite{Sengupta18}, NiW~\cite{Trigeorgis17}) or for general scenes (Marr Rev~\cite{bansal16}, UberNet~\cite{Kokkinos17}). We also compare against state-of-the-art approaches for 3D face reconstruction, namely the classic 3DMM fitting method used in~\cite{Trigeorgis17}, 3DDFA~\cite{Zhu17}, the bump map regression based approach of \cite{Tran18} and the combined regression+shape-from-shading approach of~\cite{Sela17}. 

Quantitative results can be found in Table~\ref{tab:quant_ph} for Photoface and Table~\ref{tab:quant_fl} for Florence datasets. We show results of our method both with (\texttt{Ours-ft}) and without (\texttt{Ours}) fine-tuning of the training split of Photoface in the upper and lower parts of Table~\ref{tab:quant_ph} respectively. The same is done with SfSNet. The error values on Photoface for the methods of \cite{Sengupta18, Trigeorgis17, Sela17, bansal16, Kokkinos17} are as reported in~\cite{Sengupta18}, and we use the publicly available implementations of \cite{Tran18, Zhu17, feng18} for the others. 
For the Florence dataset we use the publicly available implementations. Note that, to be able to evaluate the per-pixel normal accuracy, we can only compare to 3D reconstruction methods whose output is aligned with the image. For a fair comparison, all methods were given facial images of size $256 \times 256$ as input, resized if necessary.

The proposed approach shows the best values both in mean angular error and percentage under $20^\circ$, $25^\circ$ and $30^\circ$ degrees, only outperformed by 3DMM on errors under $30^\circ$. As noted by the authors in~\cite{Trigeorgis17}, 3DMM fitting performs well under $30^\circ$ because of the coarseness of the model and the keypoint supervision, but its performance on tighter angles drops drastically as it lacks precision. We found that, although \cite{Sela17, Tran18} usually provide seemingly detailed reconstructions, the actual normals of these methods lack accuracy as witnessed by their numbers.

Our good performance is also confirmed by qualitative comparisons over images in-the-wild in various head poses and under arbitrary lighting conditions as can be seen in Figs.~\ref{fig:comparison_norm} and~\ref{fig:comparison_mesh}. For comparisons with mesh results (Fig.~\ref{fig:comparison_mesh}), we show for both our approach and SfSNet~\cite{Sengupta18} the normal mapping over the same base mesh, obtained using PRN~\cite{feng18}, and we refer to these as Ours+PRN and SfSNet+PRN respectively. We show our meshes from two views to illustrate that the output is not optimized for a
particular viewpoint, a known limitation with SfS. Compared to SfSNet we recover much more refined details that significantly enhance the base mesh. Compared to Extreme~\cite{Tran18} our approach does not include unnecessary additional  noise. As observed by other authors, Pix2Vertex~\cite{Sela17} cannot handle difficult poses or illuminations, and sometimes simply fails to converge. Both PRN and 3DDFA can correctly recover the general structure of the face, although their goal  was not to recover surface details as we do.

We believe our improved results are due to the fact that we do not rely on a parametric model for training data generation, as was done in \eg~\cite{Sengupta18}, as well as the strongly regularized latent space that is learned through the two encoder/decoder networks, in addition to the skip connections that can transfer the necessary details.



\subsection{Ablation}
\label{sec:abl}

\begin{figure}[h!]
\centering

\begin{subfigure}[t]{0.23\linewidth}
\includegraphics[width=\linewidth]{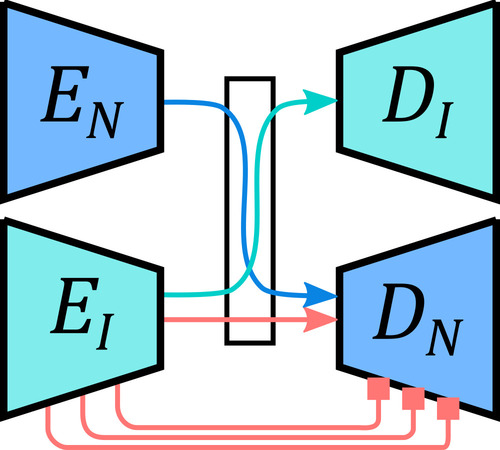}
\centering
\caption{\scriptsize Ours}
\label{fig:archi1}
\end{subfigure}
\hspace{1pt}
\begin{subfigure}[t]{0.23\linewidth}
\includegraphics[width=\linewidth]{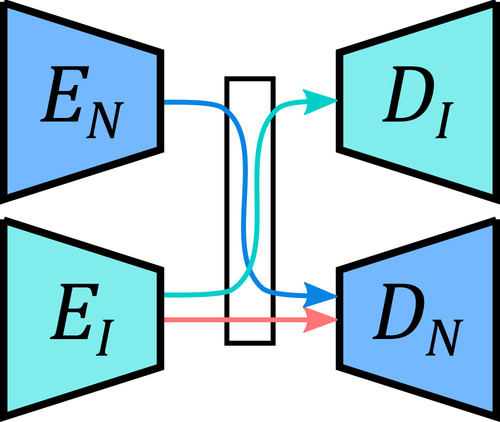}
\centering
\caption{\scriptsize w/o skip co.}
\label{fig:archi2}
\end{subfigure}
\hspace{1pt}
\begin{subfigure}[t]{0.23\linewidth}
\includegraphics[width=\linewidth]{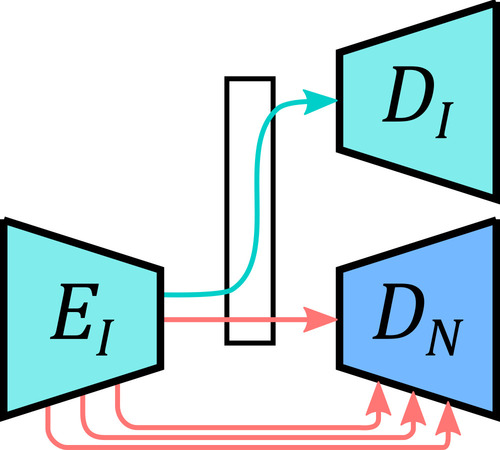}
\centering
\caption{\scriptsize w/o $E_N$}
\label{fig:archi3}
\end{subfigure}
\hspace{1pt}
\begin{subfigure}[t]{0.23\linewidth}
\includegraphics[width=\linewidth]{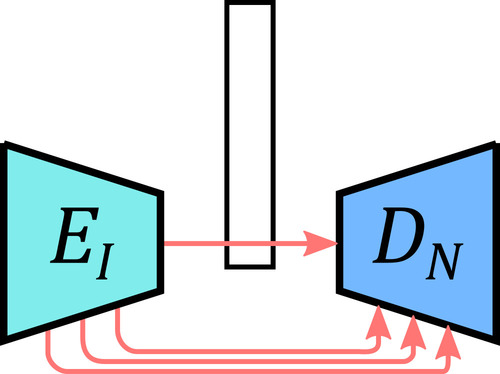}
\centering
\caption{\scriptsize w/o $E_N,D_I$}
\label{fig:archi4}
\end{subfigure}

\caption{Architectures for the ablation test:  (a) our proposed architecture, (b) without skip connections, (c) without the normal encoder  and (d) without the normal encoder and the image decoder.}
\label{fig:archis}
\end{figure}

\begin{figure}[h!]
\centering

\begin{subfigure}[t]{0.18\linewidth}
\includegraphics[width=\linewidth]{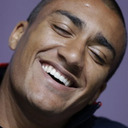}
\includegraphics[width=\linewidth]{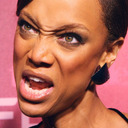}
\includegraphics[width=\linewidth]{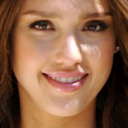}
\includegraphics[width=\linewidth]{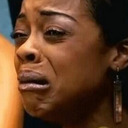}
\centering
\caption{\tiny Input}
\end{subfigure}
\begin{subfigure}[t]{0.18\linewidth}
\includegraphics[trim={1.5cm 0 1.5cm 0},clip,height=\linewidth]{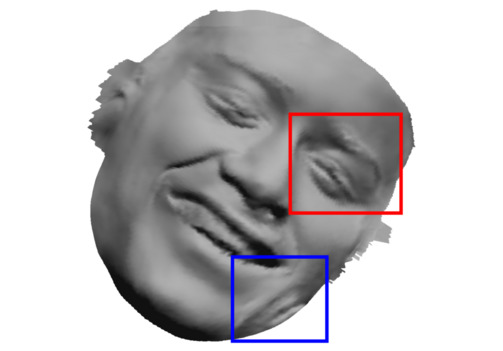}
\includegraphics[trim={1.5cm 0 1.5cm 0},clip,height=\linewidth]{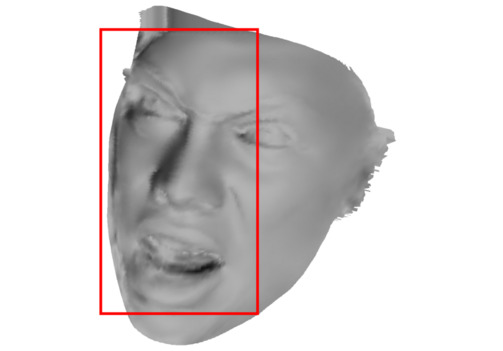}
\includegraphics[trim={1.5cm 0 1.5cm 0},clip,height=\linewidth]{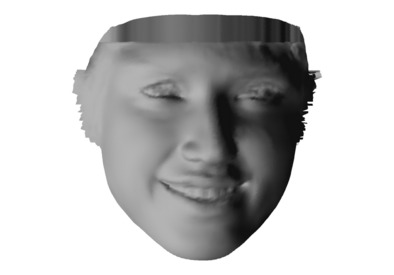}
\includegraphics[trim={1.5cm 0 1.5cm 0},clip,height=\linewidth]{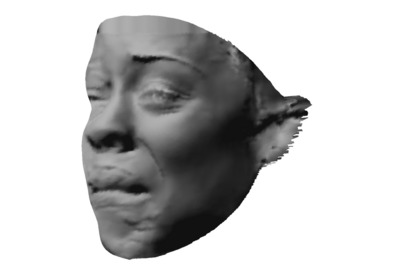}
\centering
\caption{\tiny Ours}
\label{fig:archi1_ex}
\end{subfigure}
\begin{subfigure}[t]{0.18\linewidth}
\includegraphics[trim={1.5cm 0 1.5cm 0},clip,height=\linewidth]{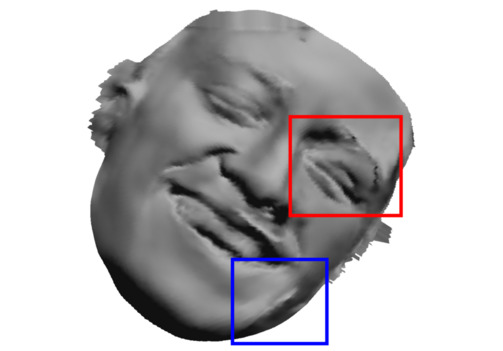}
\includegraphics[trim={1.5cm 0 1.5cm 0},clip,height=\linewidth]{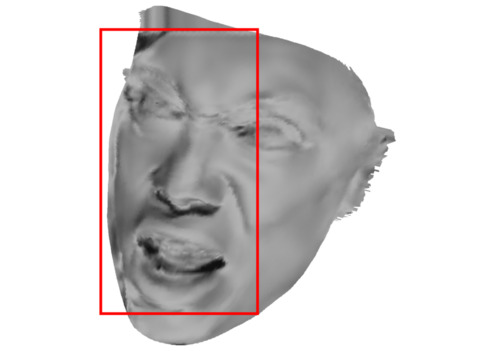}
\includegraphics[trim={1.5cm 0 1.5cm 0},clip,height=\linewidth]{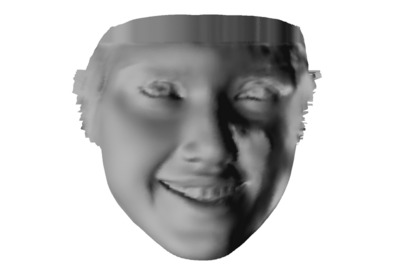}
\includegraphics[trim={1.5cm 0 1.5cm 0},clip,height=\linewidth]{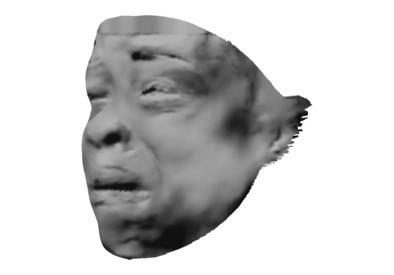}
\centering
\caption{\tiny w/o $E_N$ $D_I$}
\label{fig:archi4_ex}
\end{subfigure}
\begin{subfigure}[t]{0.18\linewidth}
\includegraphics[trim={1.5cm 0 1.5cm 0},clip,height=\linewidth]{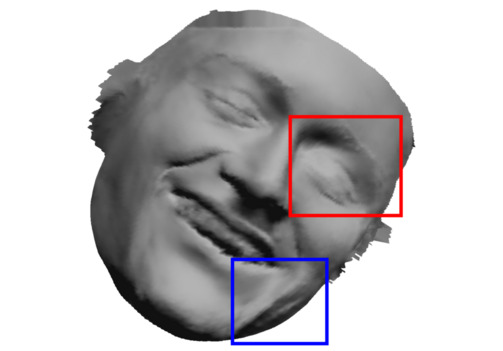}
\includegraphics[trim={1.5cm 0 1.5cm 0},clip,height=\linewidth]{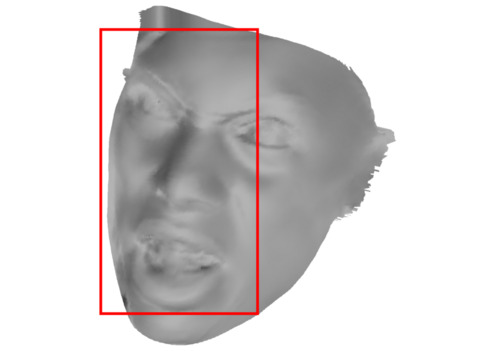}
\includegraphics[trim={1.5cm 0 1.5cm 0},clip,height=\linewidth]{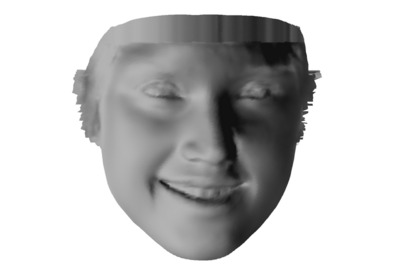}
\includegraphics[trim={1.5cm 0 1.5cm 0},clip,height=\linewidth]{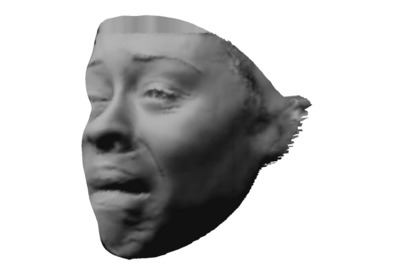}
\centering
\caption{\tiny w/o $E_N$}
\label{fig:archi3_ex}
\end{subfigure}
\begin{subfigure}[t]{0.18\linewidth}
\includegraphics[trim={1.5cm 0 1.5cm 0},clip,height=\linewidth]{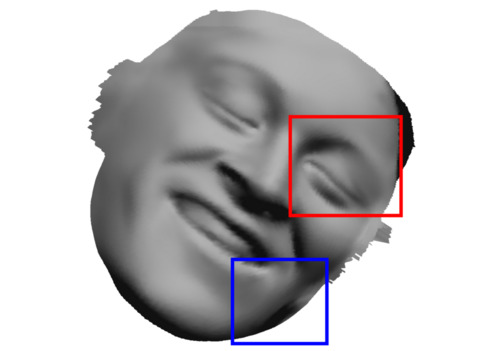}
\includegraphics[trim={1.5cm 0 1.5cm 0},clip,height=\linewidth]{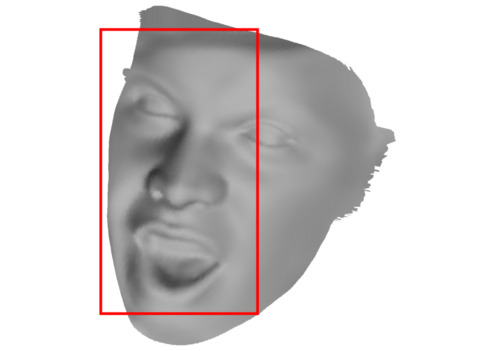}
\includegraphics[trim={1.5cm 0 1.5cm 0},clip,height=\linewidth]{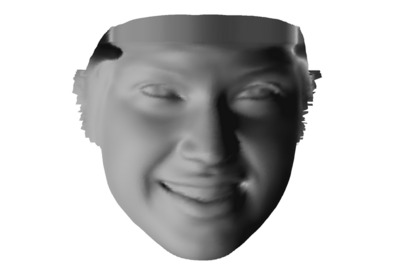}
\includegraphics[trim={1.5cm 0 1.5cm 0},clip,height=\linewidth]{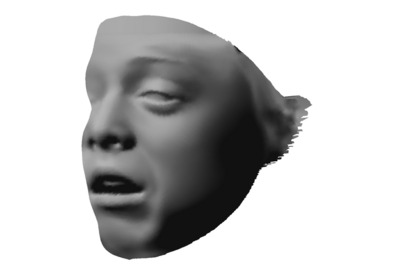}
\centering
\caption{\tiny w/o skip co.}
\label{fig:archi2_ex}
\end{subfigure}

\caption{Qualitative comparisons between architectures: (b) our proposed architecture, (c) without the normal encoder and the image decoder, (d) without the normal encoder, and (e) without skip connections.}
\label{fig:ablation}
\end{figure}

We evaluate here the influence of the proposed architectural components. In particular, we compare against the alternatives shown in Fig.~\ref{fig:archis}: our model without skip connections (Fig.~\ref{fig:archi2}), without the normal encoder $E_N$ (Fig.~\ref{fig:archi3}), and without both the normal encoder $E_N$ and image decoder $D_I$ (Fig.~\ref{fig:archi4}), \ie a basic encoder-decoder architecture. Since there is no need in the last two cases for deactivable skip connections we use standard ones. We show quantitative results in Table~\ref{tab:ablation}, and qualitative examples in Fig.~\ref{fig:ablation}.

\begin{figure}[h!]

\begin{subfigure}{\linewidth}
\centering

	\setlength{\tabcolsep}{0.4em}
	\begin{tabular}{ c|c|c|c|c| } 
	\cline{2-5}
	& Mean$\pm$std & $<20^\text{o}$ & $<25^\text{o}$ & $<30^\text{o}$  \\
	\hline 
	\multicolumn{1}{|c|}{w/o skip co.{\scriptsize (Fig.\ref{fig:archi2})}} & 24.4 $\pm$6.7 & 46.6\% & 60.6\% & 72.0\%\\
		\multicolumn{1}{|c|}{w/o $E_N,D_I${\scriptsize (Fig.\ref{fig:archi4})}}  &  23.3 $\pm$6.3 & 47.7\% & 61.9\% & 73.3\%  \\
	\multicolumn{1}{|c|}{w/o $E_N${\scriptsize (Fig.\ref{fig:archi3})}}  &  23.0 $\pm$6.8 & 47.6\% & 61.5\% & 73.1\%  \\
	\multicolumn{1}{|c|}{Ours {\scriptsize (Fig.\ref{fig:archi1})}} & {\bf 22.8$\pm$6.5} & {\bf 49.0\%} & {\bf 62.9\%} & {\bf 74.1\%} \\
	\hline
	\end{tabular}
	
\caption{On Photoface~\cite{Zafeiriou11}}
\label{tab:ablation_ph}
\end{subfigure}

\vspace{5pt}

\begin{subfigure}{\linewidth}
\centering

	\setlength{\tabcolsep}{0.4em}
	\begin{tabular}{ c|c|c|c|c| } 
	\cline{2-5}
	& Mean$\pm$std & $<20^\text{o}$ & $<25^\text{o}$ & $<30^\text{o}$  \\
	\hline 
	\multicolumn{1}{|c|}{w/o skip co.{\scriptsize (Fig.\ref{fig:archi2})}} & 12.6$\pm$1.4 & 85.8\% & 92.6\% & 95.8\% \\
	\multicolumn{1}{|c|}{w/o $E_N${\scriptsize (Fig.\ref{fig:archi3})}} & 12.4$\pm$1.6 & 86.0\% & 92.6\% & 95.9\% \\
    \multicolumn{1}{|c|}{w/o $E_N,D_I${\scriptsize (Fig.\ref{fig:archi4})}} & 12.0$\pm$1.2 & 87.8\% & 94.1\% & 96.7\% \\	
	\multicolumn{1}{|c|}{Ours {\scriptsize (Fig.\ref{fig:archi1})}} & {\bf 11.3$\pm$1.5} & {\bf 89.3\%} & {\bf 94.6\%} & {\bf 96.9\%} \\
	\hline
	\end{tabular}
	
\caption{On Florence~\cite{Bagdanov11}}
\label{tab:ablation_fl}
\end{subfigure}
\caption{Quantitative comparisons between architectures:  the proposed architecture (\emph{Ours}), without skip connections (\emph{w/o skip co.}),  without the normal encoder (\emph{w/o $E_N$}) and without the normal encoder and the image decoder (\emph{w/o $E_N,D_I$}).}
\label{tab:ablation}
\end{figure}

Our final model outperforms the alternatives both quantitatively and qualitatively which validates the proposed cross-modal architecture design, and the benefit of the introduced deactivable skip connections. 



For example, we can see in the geometric shape of the eyelids in the first row of Fig.~\ref{fig:ablation} and the shading in the second row that our final model gets the best from each of the alternatives. Our correct global shape estimate is comparable to that of the cross-modal model without skip connections, although the latter is smoother and clearly lacks details. 
Additionally we can see that removing the image decoder $D_I$ and normal encoder $E_N$ (\ie a standard encoder-decoder with skip connections) gives poor results for images-in-the-wild, due to the domain gap between training and evaluation. This can be visualized particularly in the artifacts appearing on the third and fourth examples, or the inaccurate shadings of the second example.
Finally, our fine details are comparable to those of the model with skip connections but without the normal encoder $E_N$, which in turn has a reduced ability to represent the shape accurately, since it has not learned an additional prior on the geometric aspects of the face.

\subsection{Limitations}
\label{sec:limitations}

\begin{figure}[h!]
\centering
\begin{subfigure}{.13\linewidth}
\centering
\includegraphics[width=\linewidth]{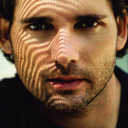}
\includegraphics[width=\linewidth]{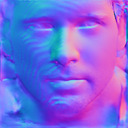}    
\caption{}
\label{fig:failure1}
\end{subfigure}
\begin{subfigure}{.13\linewidth}
\centering
\includegraphics[width=\linewidth]{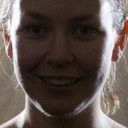}
\includegraphics[width=\linewidth]{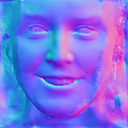}    
\caption{}
\label{fig:failure2}
\end{subfigure}
\begin{subfigure}{.13\linewidth}
\centering
\includegraphics[width=\linewidth]{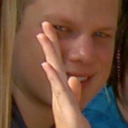}
\includegraphics[width=\linewidth]{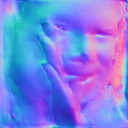}    
\caption{}
\label{fig:failure3}
\end{subfigure}
\begin{subfigure}{.13\linewidth}
\centering
\includegraphics[width=\linewidth]{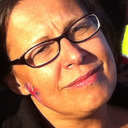}
\includegraphics[width=\linewidth]{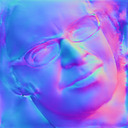}    
\caption{}
\label{fig:failure4}
\end{subfigure}
\begin{subfigure}{.13\linewidth}
\centering
\includegraphics[width=\linewidth]{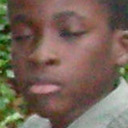}
\includegraphics[width=\linewidth]{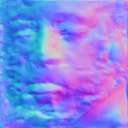}    
\caption{}
\label{fig:failure5}
\end{subfigure}
\begin{subfigure}{.13\linewidth}
\centering
\includegraphics[width=\linewidth]{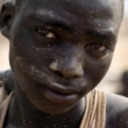}
\includegraphics[width=\linewidth]{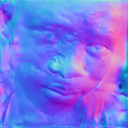}    
\caption{}
\label{fig:failure6}
\end{subfigure}

\vspace{-5pt}
\caption{Failure cases.}
\label{fig:failure}
\end{figure}

The proposed method still has limitations, some of which are shown in Fig.~\ref{fig:failure}. These belong to extreme situations that represent outliers to the training data, including faces in very severe lighting/shades (Fig.\ref{fig:failure1},\ref{fig:failure2}), occlusion (Fig.\ref{fig:failure3},\ref{fig:failure4}), very low quality images (Fig.\ref{fig:failure5}) and unusual facial textures (Fig.\ref{fig:failure6}).

\section{Conclusion}

We presented a novel deep-learning based approach for the estimation of facial normals in-the-wild. Our method is centered on a new architecture that combines the robustness of cross-modal learning and the detail transfer ability of skip connections, enabled thanks to the proposed \emph{deactivable skip connections}. By leveraging both paired and unpaired data of image and normal modalities during training, we achieve state-of-the-art results on angular estimation errors and obtain visually compelling enhanced 3D reconstructions on challenging images in-the-wild. 
%
Among the limitations of our work are the inability to properly handle occlusions (as it is mostly a local method) and to recover finer-details, \eg pore-level details, which are directions that will be tackled in future work.

{\small
\bibliographystyle{ieee_fullname}
\bibliography{paper}
}

\pagebreak
\begingroup
\let\clearpage\relax 
\onecolumn 
\begin{centering}
{\Large \bf Cross-modal Deep Face Normals with Deactivable Skip Connections \\ \vspace{5pt} Supplementary Material}\\
\vspace{12pt}
\end{centering}



\paragraph{Low-cost depth enhancement:}
We can use our model to enhance the appearance of the noisy depth data coming from low-cost RGB-D sensors, e.g. Kinect. We show an example of this using the FaceWarehouse dataset~\cite{Cao13}, where we use the accompanying RGB image to predict normals with our method, and append these normals to the raw depth image pixel-wise using normal mapping~\cite{Cohen98}, thus rendering enhanced geometric shading. In Fig.~\ref{fig:kinect} we show the RGB images in the first row, the raw depth in the second, and the same depth enhanced with our model's predictions in the last one. The ability to recover accurate normals allows to enhance the depth appearance significantly.

\newcommand{\szd}{0.085}

\begin{figure*}[h!]
\centering

\begin{subfigure}{\szd\linewidth}
\includegraphics[width=\linewidth]{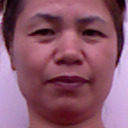}
\includegraphics[trim={3.25cm 1.6cm 3.25cm 1.6cm},clip,width=\linewidth]{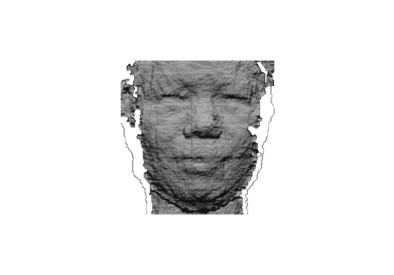}
\includegraphics[trim={3.25cm 1.6cm 3.25cm 1.6cm},clip,width=\linewidth]{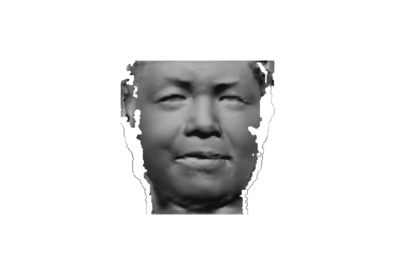}
\end{subfigure}
\begin{subfigure}{\szd\linewidth}
\includegraphics[width=\linewidth]{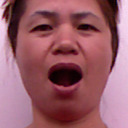}
\includegraphics[trim={3.25cm 1.6cm 3.25cm 1.6cm},clip,width=\linewidth]{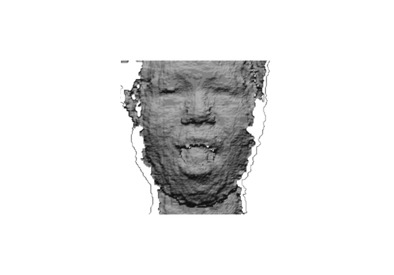}
\includegraphics[trim={3.25cm 1.6cm 3.25cm 1.6cm},clip,width=\linewidth]{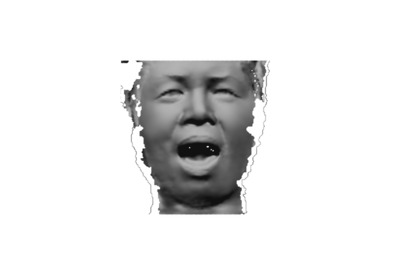}
\end{subfigure}
\begin{subfigure}{\szd\linewidth}
\includegraphics[width=\linewidth]{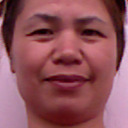}
\includegraphics[trim={3.25cm 1.6cm 3.25cm 1.6cm},clip,width=\linewidth]{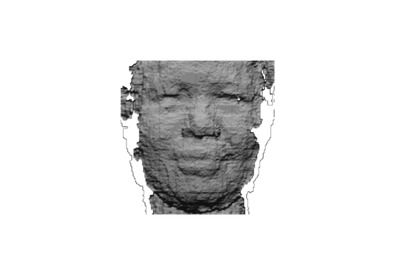}
\includegraphics[trim={3.25cm 1.6cm 3.25cm 1.6cm},clip,width=\linewidth]{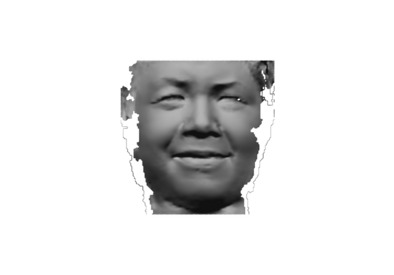}
\end{subfigure}
\begin{subfigure}{\szd\linewidth}
\includegraphics[width=\linewidth]{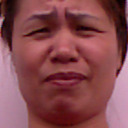}
\includegraphics[trim={3.25cm 1.6cm 3.25cm 1.6cm},clip,width=\linewidth]{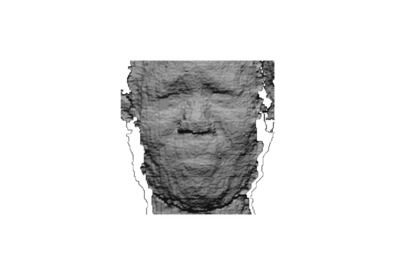}
\includegraphics[trim={3.25cm 1.6cm 3.25cm 1.6cm},clip,width=\linewidth]{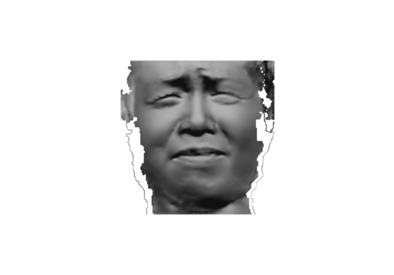}
\end{subfigure}
\begin{subfigure}{\szd\linewidth}
\includegraphics[width=\linewidth]{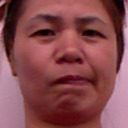}
\includegraphics[trim={3.25cm 1.6cm 3.25cm 1.6cm},clip,width=\linewidth]{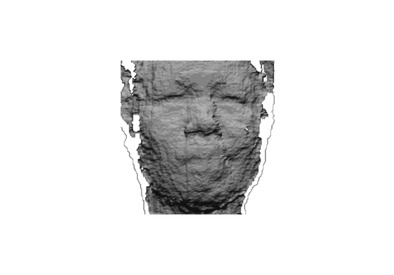}
\includegraphics[trim={3.25cm 1.6cm 3.25cm 1.6cm},clip,width=\linewidth]{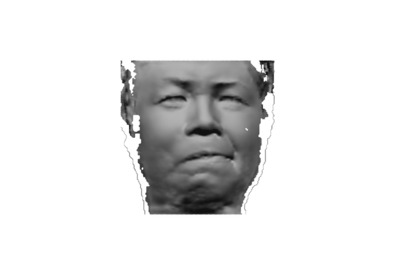}
\end{subfigure}
\begin{subfigure}{\szd\linewidth}
\includegraphics[width=\linewidth]{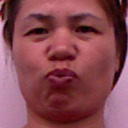}
\includegraphics[trim={3.25cm 1.6cm 3.25cm 1.6cm},clip,width=\linewidth]{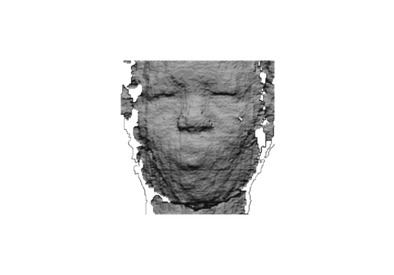}
\includegraphics[trim={3.25cm 1.6cm 3.25cm 1.6cm},clip,width=\linewidth]{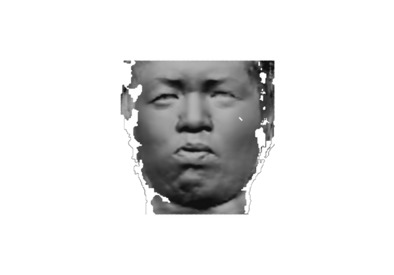}
\end{subfigure}
\begin{subfigure}{\szd\linewidth}
\includegraphics[width=\linewidth]{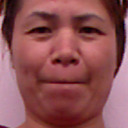}
\includegraphics[trim={3.25cm 1.6cm 3.25cm 1.6cm},clip,width=\linewidth]{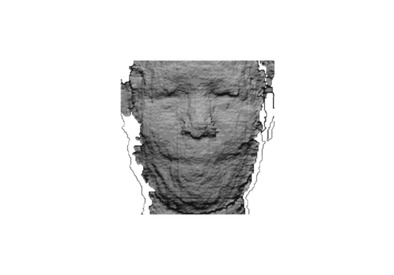}
\includegraphics[trim={3.25cm 1.6cm 3.25cm 1.6cm},clip,width=\linewidth]{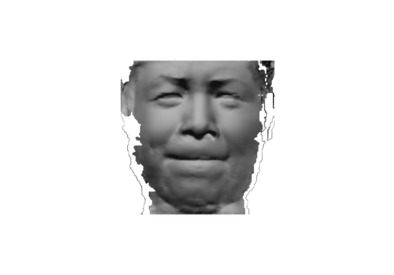}
\end{subfigure}
\begin{subfigure}{\szd\linewidth}
\includegraphics[width=\linewidth]{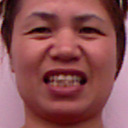}
\includegraphics[trim={3.25cm 1.6cm 3.25cm 1.6cm},clip,width=\linewidth]{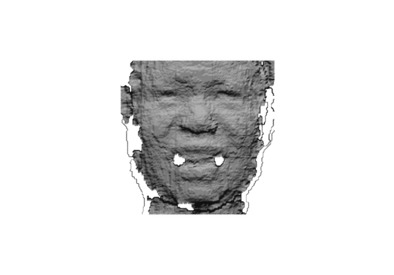}
\includegraphics[trim={3.25cm 1.6cm 3.25cm 1.6cm},clip,width=\linewidth]{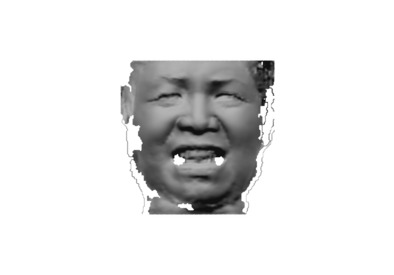}
\end{subfigure}
\begin{subfigure}{\szd\linewidth}
\includegraphics[width=\linewidth]{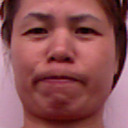}
\includegraphics[trim={3.25cm 1.6cm 3.25cm 1.6cm},clip,width=\linewidth]{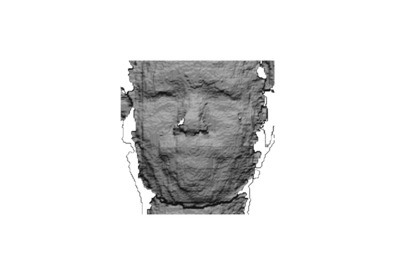}
\includegraphics[trim={3.25cm 1.6cm 3.25cm 1.6cm},clip,width=\linewidth]{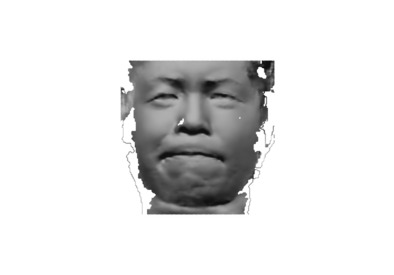}
\end{subfigure}
\begin{subfigure}{\szd\linewidth}
\includegraphics[width=\linewidth]{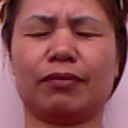}
\includegraphics[trim={3.25cm 1.6cm 3.25cm 1.6cm},clip,width=\linewidth]{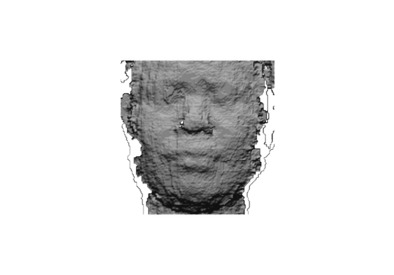}
\includegraphics[trim={3.25cm 1.6cm 3.25cm 1.6cm},clip,width=\linewidth]{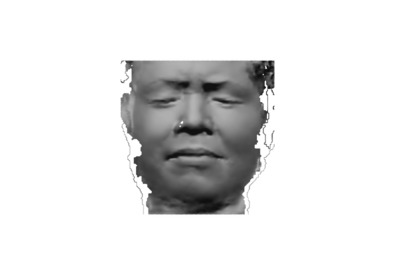}
\end{subfigure}

\vspace{15pt}

\begin{subfigure}{\szd\linewidth}
\includegraphics[width=\linewidth]{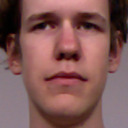}
\includegraphics[trim={1.65cm 0 1.65cm 0},clip,width=\linewidth]{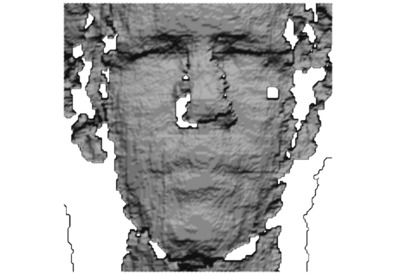}
\includegraphics[trim={1.65cm 0 1.65cm 0},clip,width=\linewidth]{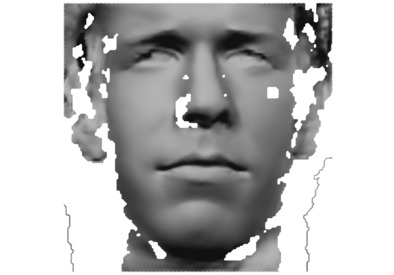}
\end{subfigure}
\begin{subfigure}{\szd\linewidth}
\includegraphics[width=\linewidth]{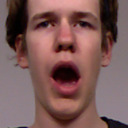}
\includegraphics[trim={1.65cm 0 1.65cm 0},clip,width=\linewidth]{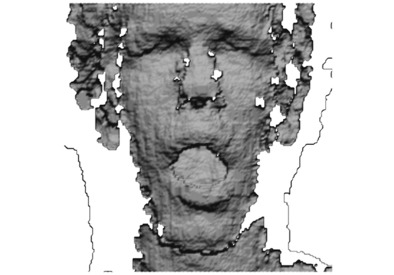}
\includegraphics[trim={1.65cm 0 1.65cm 0},clip,width=\linewidth]{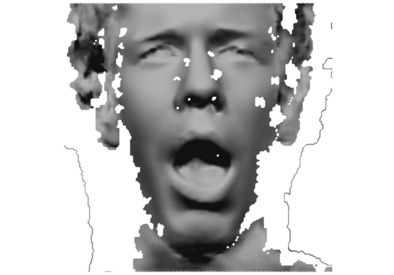}
\end{subfigure}
\begin{subfigure}{\szd\linewidth}
\includegraphics[width=\linewidth]{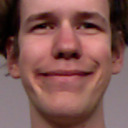}
\includegraphics[trim={1.65cm 0 1.65cm 0},clip,width=\linewidth]{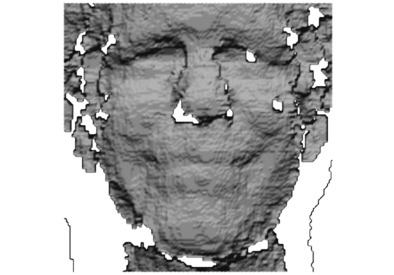}
\includegraphics[trim={1.65cm 0 1.65cm 0},clip,width=\linewidth]{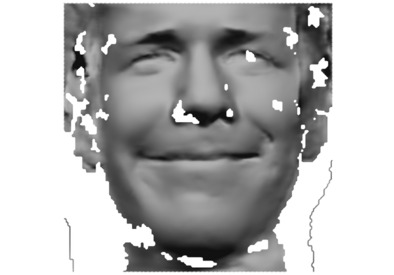}
\end{subfigure}
\begin{subfigure}{\szd\linewidth}
\includegraphics[width=\linewidth]{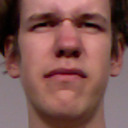}
\includegraphics[trim={1.65cm 0 1.65cm 0},clip,width=\linewidth]{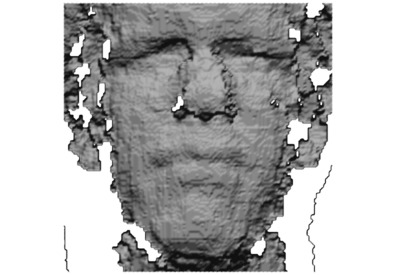}
\includegraphics[trim={1.65cm 0 1.65cm 0},clip,width=\linewidth]{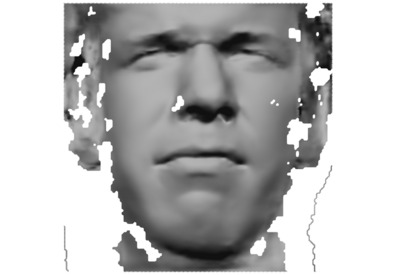}
\end{subfigure}
\begin{subfigure}{\szd\linewidth}
\includegraphics[width=\linewidth]{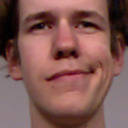}
\includegraphics[trim={1.65cm 0 1.65cm 0},clip,width=\linewidth]{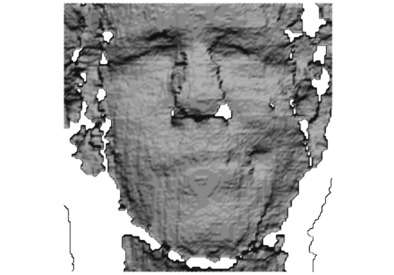}
\includegraphics[trim={1.65cm 0 1.65cm 0},clip,width=\linewidth]{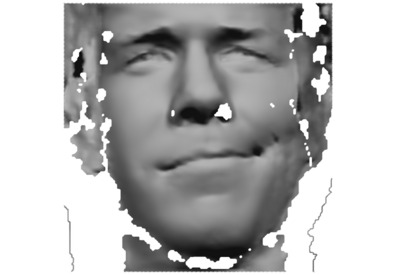}
\end{subfigure}
\begin{subfigure}{\szd\linewidth}
\includegraphics[width=\linewidth]{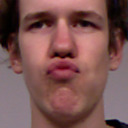}
\includegraphics[trim={1.65cm 0 1.65cm 0},clip,width=\linewidth]{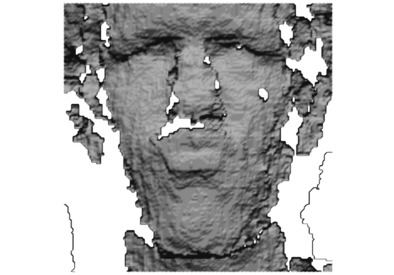}
\includegraphics[trim={1.65cm 0 1.65cm 0},clip,width=\linewidth]{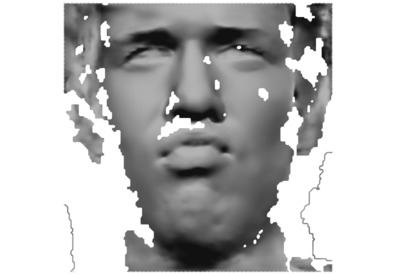}
\end{subfigure}
\begin{subfigure}{\szd\linewidth}
\includegraphics[width=\linewidth]{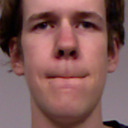}
\includegraphics[trim={1.65cm 0 1.65cm 0},clip,width=\linewidth]{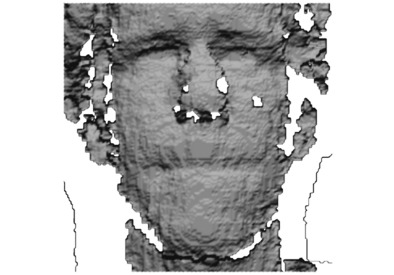}
\includegraphics[trim={1.65cm 0 1.65cm 0},clip,width=\linewidth]{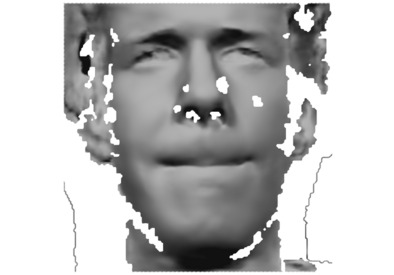}
\end{subfigure}
\begin{subfigure}{\szd\linewidth}
\includegraphics[width=\linewidth]{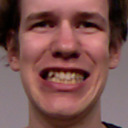}
\includegraphics[trim={1.65cm 0 1.65cm 0},clip,width=\linewidth]{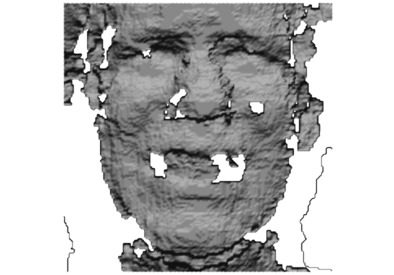}
\includegraphics[trim={1.65cm 0 1.65cm 0},clip,width=\linewidth]{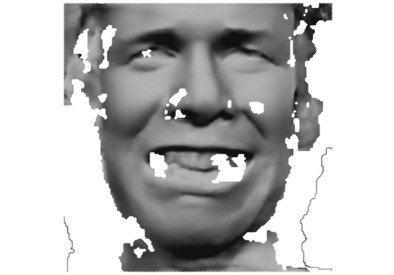}
\end{subfigure}
\begin{subfigure}{\szd\linewidth}
\includegraphics[width=\linewidth]{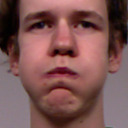}
\includegraphics[trim={1.65cm 0 1.65cm 0},clip,width=\linewidth]{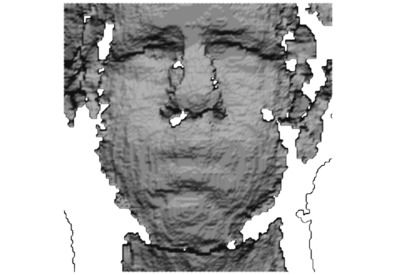}
\includegraphics[trim={1.65cm 0 1.65cm 0},clip,width=\linewidth]{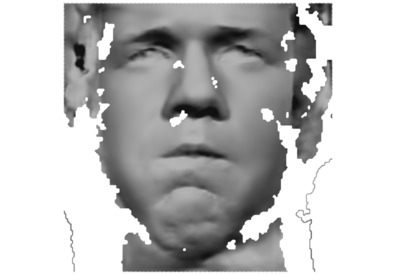}
\end{subfigure}
\begin{subfigure}{\szd\linewidth}
\includegraphics[width=\linewidth]{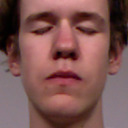}
\includegraphics[trim={1.65cm 0 1.65cm 0},clip,width=\linewidth]{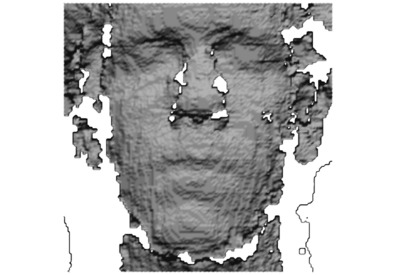}
\includegraphics[trim={1.65cm 0 1.65cm 0},clip,width=\linewidth]{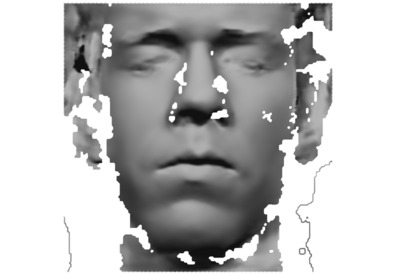}
\end{subfigure}

\vspace{15pt}

\begin{subfigure}{\szd\linewidth}
\includegraphics[width=\linewidth]{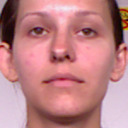}
\includegraphics[trim={1.65cm 0.15cm 1.65cm 0.15cm},clip,width=\linewidth]{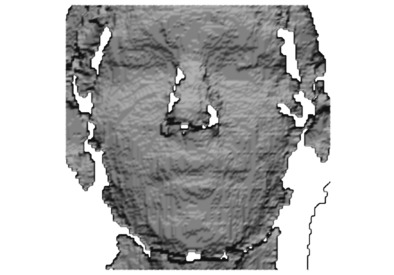}
\includegraphics[trim={1.65cm 0.15cm 1.65cm 0.15cm},clip,width=\linewidth]{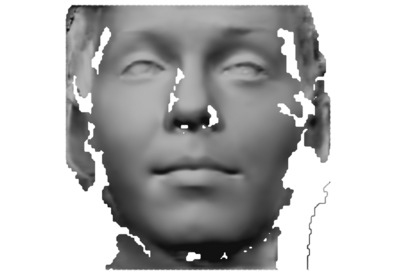}
\end{subfigure}
\begin{subfigure}{\szd\linewidth}
\includegraphics[width=\linewidth]{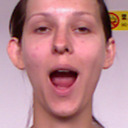}
\includegraphics[trim={1.65cm 0.15cm 1.65cm 0.15cm},clip,width=\linewidth]{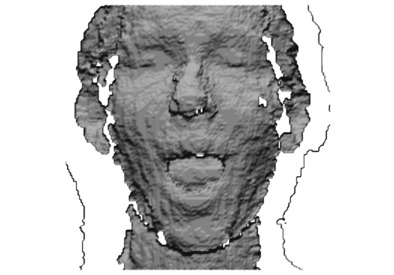}
\includegraphics[trim={1.65cm 0.15cm 1.65cm 0.15cm},clip,width=\linewidth]{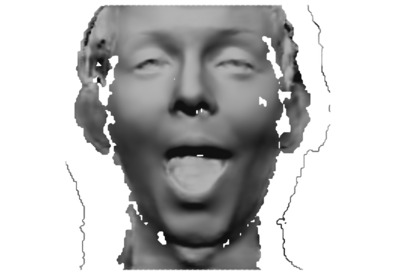}
\end{subfigure}
\begin{subfigure}{\szd\linewidth}
\includegraphics[width=\linewidth]{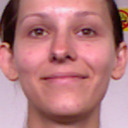}
\includegraphics[trim={1.65cm 0.15cm 1.65cm 0.15cm},clip,width=\linewidth]{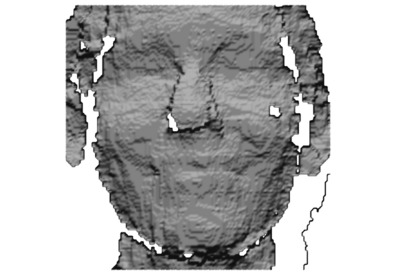}
\includegraphics[trim={1.65cm 0.15cm 1.65cm 0.15cm},clip,width=\linewidth]{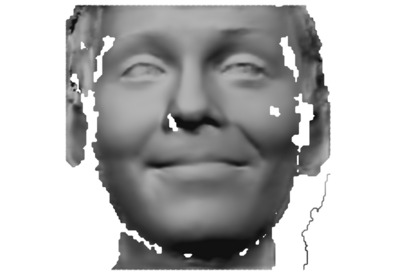}
\end{subfigure}
\begin{subfigure}{\szd\linewidth}
\includegraphics[width=\linewidth]{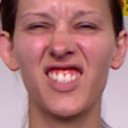}
\includegraphics[trim={1.65cm 0.15cm 1.65cm 0.15cm},clip,width=\linewidth]{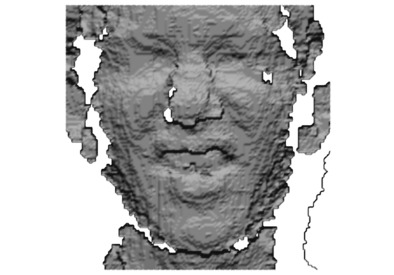}
\includegraphics[trim={1.65cm 0.15cm 1.65cm 0.15cm},clip,width=\linewidth]{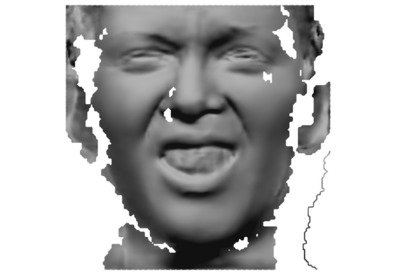}
\end{subfigure}
\begin{subfigure}{\szd\linewidth}
\includegraphics[width=\linewidth]{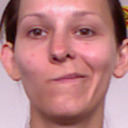}
\includegraphics[trim={1.65cm 0.15cm 1.65cm 0.15cm},clip,width=\linewidth]{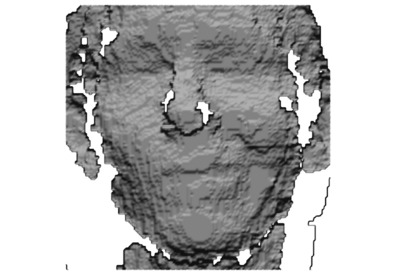}
\includegraphics[trim={1.65cm 0.15cm 1.65cm 0.15cm},clip,width=\linewidth]{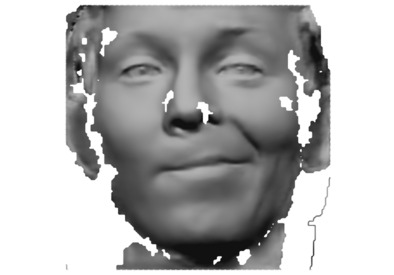}
\end{subfigure}
\begin{subfigure}{\szd\linewidth}
\includegraphics[width=\linewidth]{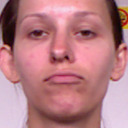}
\includegraphics[trim={1.65cm 0.15cm 1.65cm 0.15cm},clip,width=\linewidth]{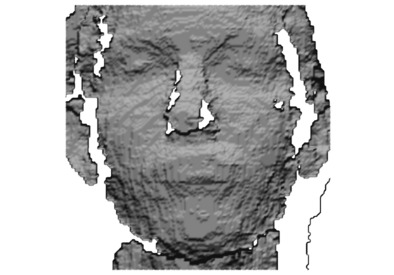}
\includegraphics[trim={1.65cm 0.15cm 1.65cm 0.15cm},clip,width=\linewidth]{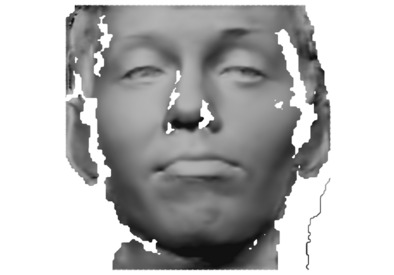}
\end{subfigure}
\begin{subfigure}{\szd\linewidth}
\includegraphics[width=\linewidth]{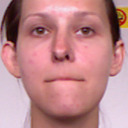}
\includegraphics[trim={1.65cm 0.15cm 1.65cm 0.15cm},clip,width=\linewidth]{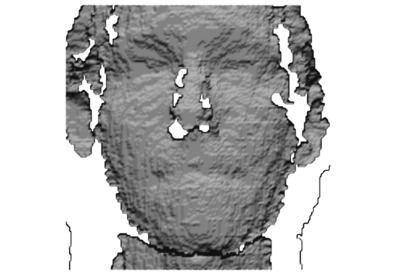}
\includegraphics[trim={1.65cm 0.15cm 1.65cm 0.15cm},clip,width=\linewidth]{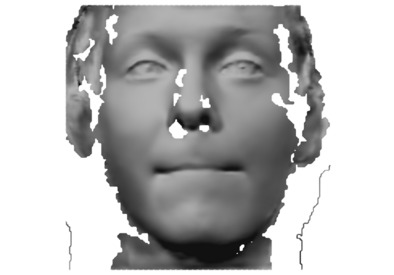}
\end{subfigure}
\begin{subfigure}{\szd\linewidth}
\includegraphics[width=\linewidth]{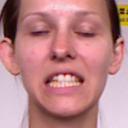}
\includegraphics[trim={1.65cm 0.15cm 1.65cm 0.15cm},clip,width=\linewidth]{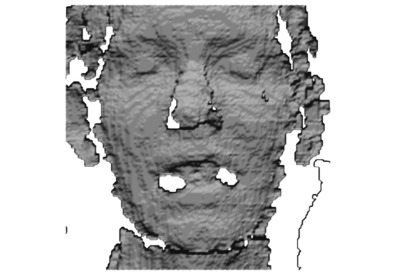}
\includegraphics[trim={1.65cm 0.15cm 1.65cm 0.15cm},clip,width=\linewidth]{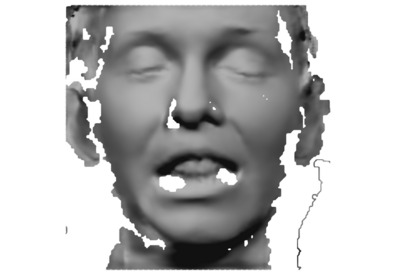}
\end{subfigure}
\begin{subfigure}{\szd\linewidth}
\includegraphics[width=\linewidth]{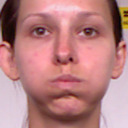}
\includegraphics[trim={1.65cm 0.15cm 1.65cm 0.15cm},clip,width=\linewidth]{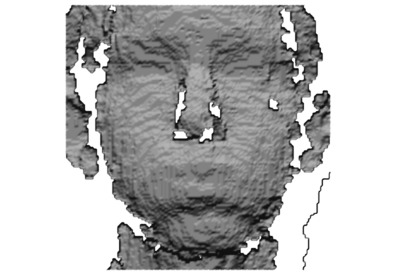}
\includegraphics[trim={1.65cm 0.15cm 1.65cm 0.15cm},clip,width=\linewidth]{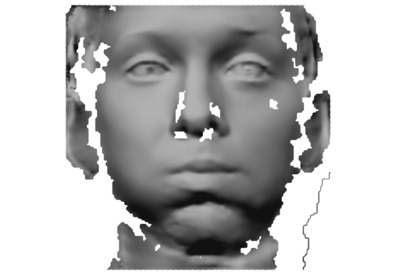}
\end{subfigure}
\begin{subfigure}{\szd\linewidth}
\includegraphics[width=\linewidth]{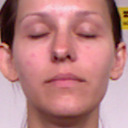}
\includegraphics[trim={1.65cm 0.15cm 1.65cm 0.15cm},clip,width=\linewidth]{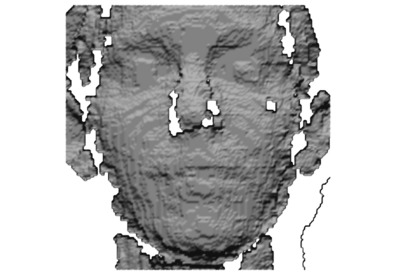}
\includegraphics[trim={1.65cm 0.15cm 1.65cm 0.15cm},clip,width=\linewidth]{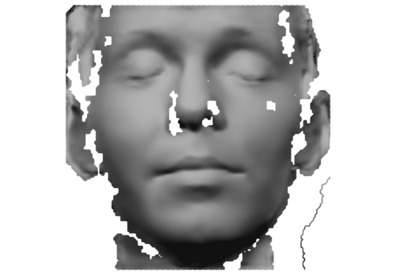}
\end{subfigure}

\centering
\caption{Raw Kinect depth enhancement using our normals on the Facewarehouse dataset \cite{Cao13}.}
\label{fig:kinect}

\end{figure*}

\vspace{-5pt}

\paragraph{Supplementary qualitative comparisons:}
Figs.~\ref{fig:comparison_mesh2} and \ref{fig:comparison_norm2} show additional predictions from our model in comparison to competing methods on the 300-W dataset~\cite{Sagonas13} in both the normal and geometry domains.
 
\newcommand{\sze}{0.11}
\newcommand{\szf}{1.5}

\clearpage
\begin{figure*}[h!]
\centering

\begin{subfigure}{\sze\linewidth}
\includegraphics[width=\linewidth]{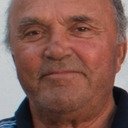}
\centering
\end{subfigure}
\begin{subfigure}{\sze\linewidth}
\includegraphics[trim={\szf cm 0 \szf cm 0},clip,width=\linewidth]{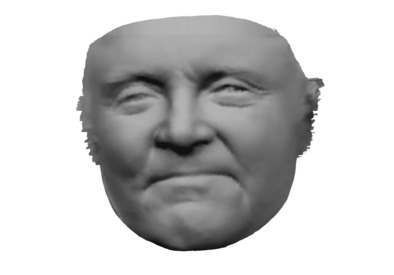}
\centering
\end{subfigure}
\begin{subfigure}{\sze\linewidth}
\includegraphics[trim={\szf cm 0 \szf cm 0},clip,width=\linewidth]{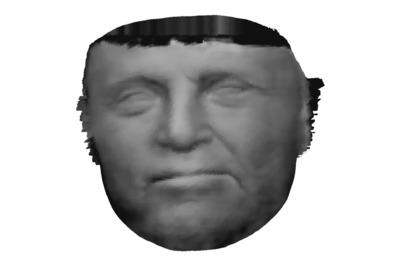}
\centering
\end{subfigure}
\begin{subfigure}{\sze\linewidth}
\includegraphics[trim={\szf cm 0 \szf cm 0},clip,width=\linewidth]{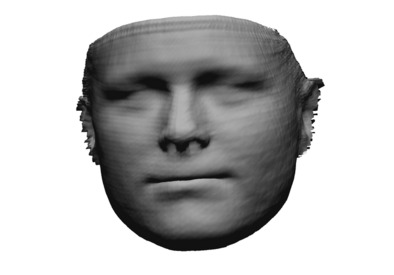}
\centering
\end{subfigure}
\begin{subfigure}{\sze\linewidth}
\includegraphics[trim={\szf cm 0 \szf cm 0},clip,width=\linewidth]{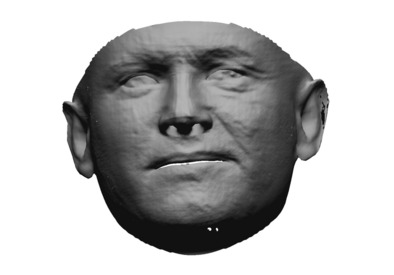}
\centering
\end{subfigure}
\begin{subfigure}{\sze\linewidth}
\includegraphics[trim={\szf cm 0 \szf cm 0},clip,width=\linewidth]{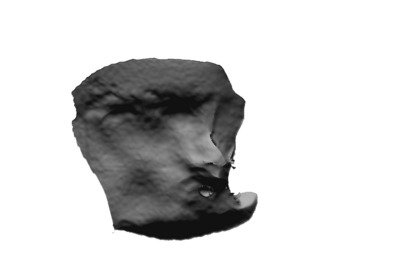}
\centering
\end{subfigure}
\begin{subfigure}{\sze\linewidth}
\includegraphics[trim={\szf cm 0 \szf cm 0},clip,width=\linewidth]{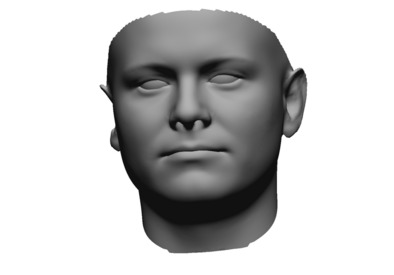}
\centering
\end{subfigure}

\begin{subfigure}{\sze\linewidth}
\includegraphics[width=\linewidth]{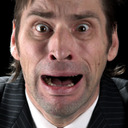}
\centering
\end{subfigure}
\begin{subfigure}{\sze\linewidth}
\includegraphics[trim={2cm 0.5cm 2cm 0.5cm},clip,width=\linewidth]{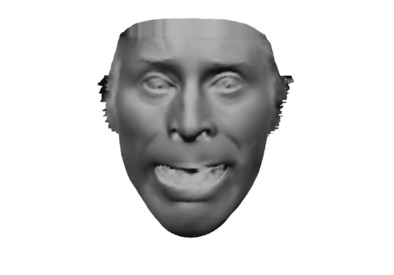}
\centering
\end{subfigure}
\begin{subfigure}{\sze\linewidth}
\includegraphics[trim={2cm 0.5cm 2cm 0.5cm},clip,width=\linewidth]{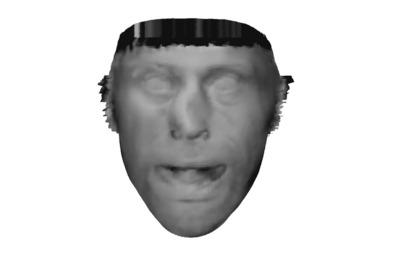}
\centering
\end{subfigure}
\begin{subfigure}{\sze\linewidth}
\includegraphics[trim={2cm 0.5cm 2cm 0.5cm},clip,width=\linewidth]{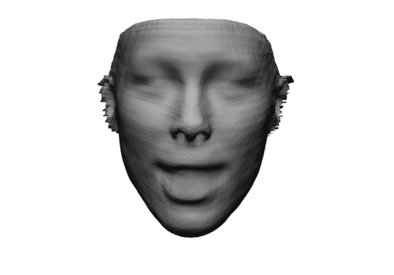}
\centering
\end{subfigure}
\begin{subfigure}{\sze\linewidth}
\includegraphics[trim={2cm 0.5cm 2cm 0.5cm},clip,width=\linewidth]{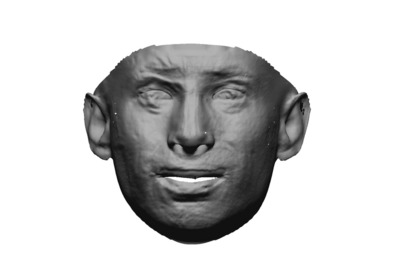}
\centering
\end{subfigure}
\begin{subfigure}{\sze\linewidth}
\includegraphics[trim={1.5cm 0.5cm 1.5cm 0.5cm},clip,width=\linewidth]{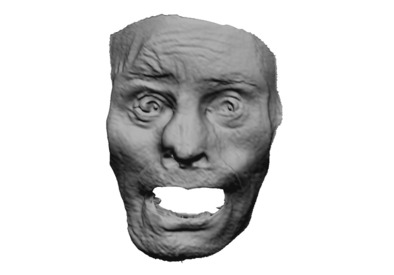}
\centering
\end{subfigure}
\begin{subfigure}{\sze\linewidth}
\includegraphics[trim={2cm 0.5cm 2cm 0.5cm},clip,width=\linewidth]{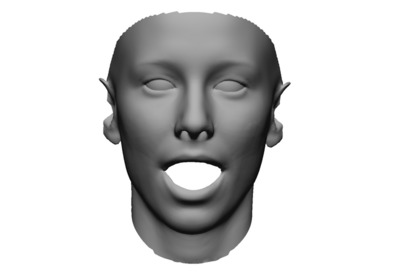}
\centering
\end{subfigure}


\begin{subfigure}{\sze\linewidth}
\includegraphics[width=\linewidth]{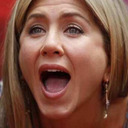}
\centering
\end{subfigure}
\begin{subfigure}{\sze\linewidth}
\includegraphics[trim={1.5cm 0.5cm 2cm 0cm},clip,width=\linewidth]{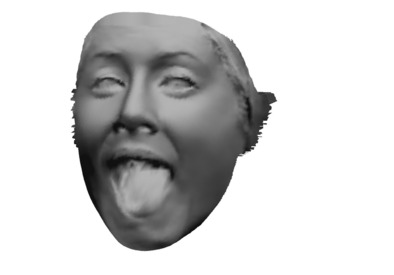}
\centering
\end{subfigure}
\begin{subfigure}{\sze\linewidth}
\includegraphics[trim={1.5cm 0.5cm 2cm 0cm},clip,width=\linewidth]{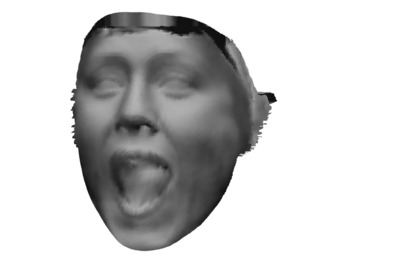}
\centering
\end{subfigure}
\begin{subfigure}{\sze\linewidth}
\includegraphics[trim={1.5cm 0.5cm 2cm 0cm},clip,width=\linewidth]{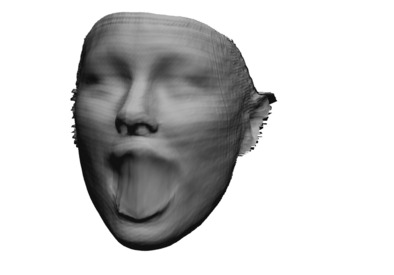}
\centering
\end{subfigure}
\begin{subfigure}{\sze\linewidth}
\includegraphics[trim={\szf cm 0 \szf cm 0},clip,width=\linewidth]{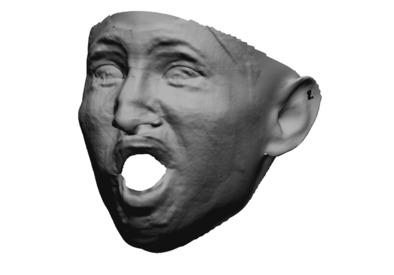}
\centering
\end{subfigure}
\begin{subfigure}{\sze\linewidth}
\includegraphics[trim={\szf cm 0 \szf cm 0},clip,width=\linewidth]{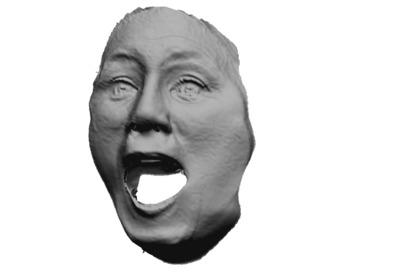}
\centering
\end{subfigure}
\begin{subfigure}{\sze\linewidth}
\includegraphics[trim={1.5cm 0.5cm 2.5cm 0.5cm},clip,width=\linewidth]{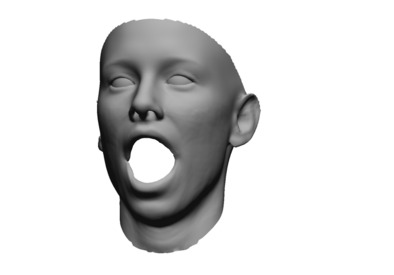}
\centering
\end{subfigure}

\begin{subfigure}{\sze\linewidth}
\includegraphics[width=\linewidth]{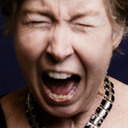}
\centering
\end{subfigure}
\begin{subfigure}{\sze\linewidth}
\includegraphics[trim={\szf cm 0 \szf cm 0},clip,width=\linewidth]{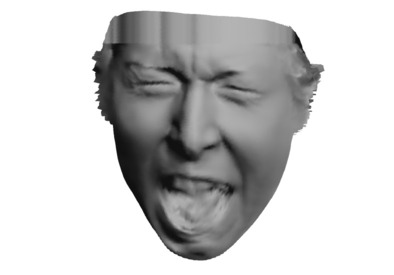}
\centering
\end{subfigure}
\begin{subfigure}{\sze\linewidth}
\includegraphics[trim={\szf cm 0 \szf cm 0},clip,width=\linewidth]{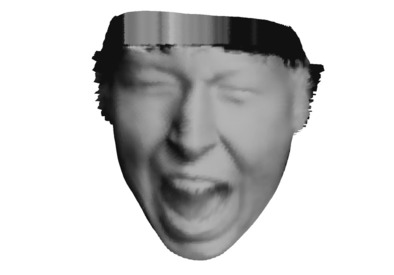}
\centering
\end{subfigure}
\begin{subfigure}{\sze\linewidth}
\includegraphics[trim={\szf cm 0 \szf cm 0},clip,width=\linewidth]{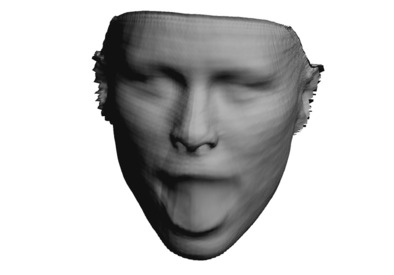}
\centering
\end{subfigure}
\begin{subfigure}{\sze\linewidth}
\includegraphics[trim={\szf cm 0 \szf cm 0},clip,width=\linewidth]{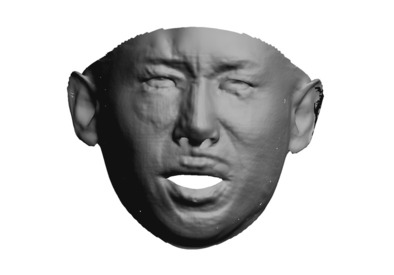}
\centering
\end{subfigure}
\begin{subfigure}{\sze\linewidth}
\includegraphics[trim={\szf cm 0 \szf cm 0},clip,width=\linewidth]{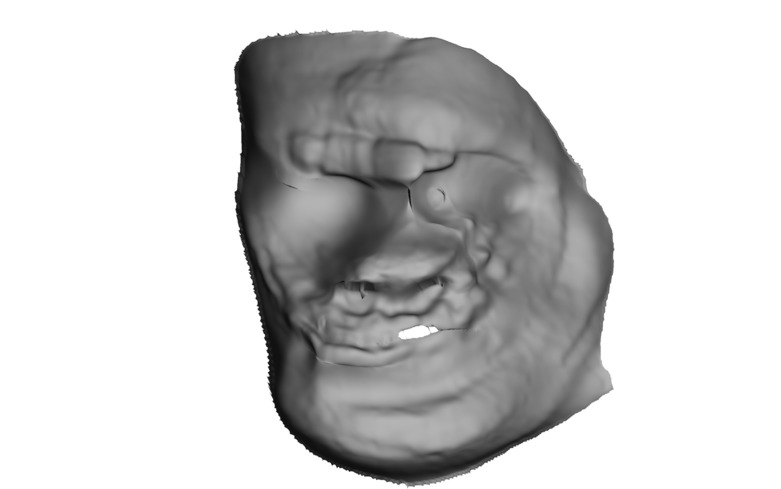}
\centering
\end{subfigure}
\begin{subfigure}{\sze\linewidth}
\includegraphics[trim={\szf cm 0 \szf cm 0},clip,width=\linewidth]{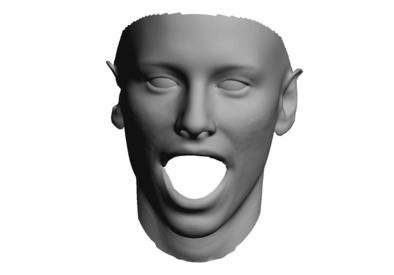}
\centering
\end{subfigure}

\begin{subfigure}{\sze\linewidth}
\includegraphics[width=\linewidth]{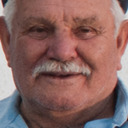}
\centering
\end{subfigure}
\begin{subfigure}{\sze\linewidth}
\includegraphics[trim={\szf cm 0 \szf cm 0},clip,width=\linewidth]{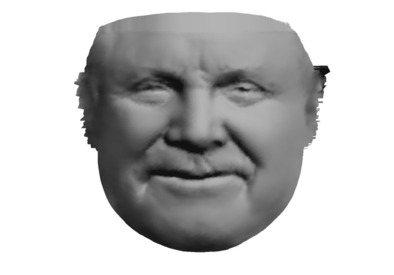}
\centering
\end{subfigure}
\begin{subfigure}{\sze\linewidth}
\includegraphics[trim={\szf cm 0 \szf cm 0},clip,width=\linewidth]{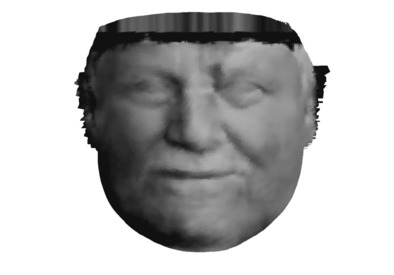}
\centering
\end{subfigure}
\begin{subfigure}{\sze\linewidth}
\includegraphics[trim={\szf cm 0 \szf cm 0},clip,width=\linewidth]{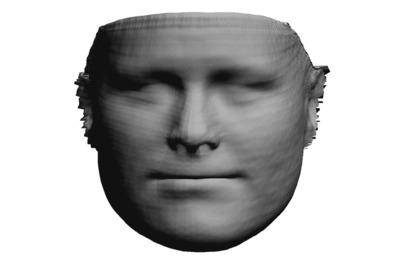}
\centering
\end{subfigure}
\begin{subfigure}{\sze\linewidth}
\includegraphics[trim={\szf cm 0 \szf cm 0},clip,width=\linewidth]{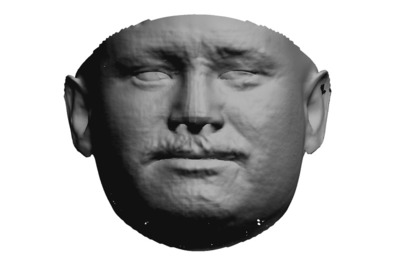}
\centering
\end{subfigure}
\begin{subfigure}{\sze\linewidth}
\includegraphics[trim={\szf cm 0 \szf cm 0},clip,width=\linewidth]{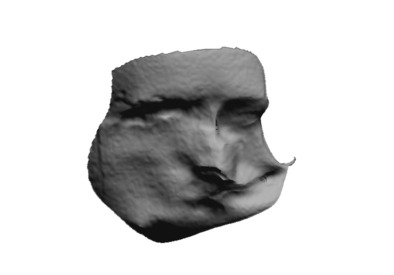}
\centering
\end{subfigure}
\begin{subfigure}{\sze\linewidth}
\includegraphics[trim={\szf cm 0 \szf cm 0},clip,width=\linewidth]{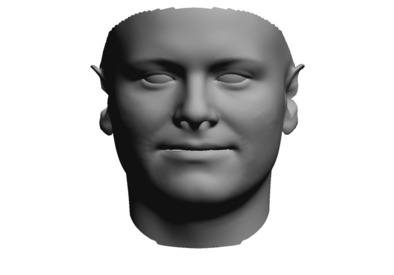}
\centering
\end{subfigure}

\begin{subfigure}{\sze\linewidth}
\includegraphics[width=\linewidth]{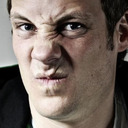}
\centering
\end{subfigure}
\begin{subfigure}{\sze\linewidth}
\includegraphics[trim={\szf cm 0 \szf cm 0},clip,width=\linewidth]{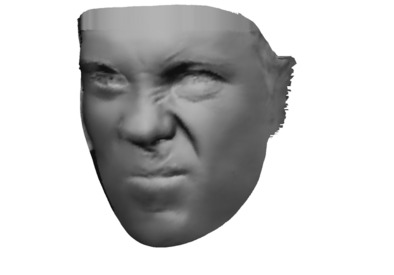}
\centering
\end{subfigure}
\begin{subfigure}{\sze\linewidth}
\includegraphics[trim={\szf cm 0 \szf cm 0},clip,width=\linewidth]{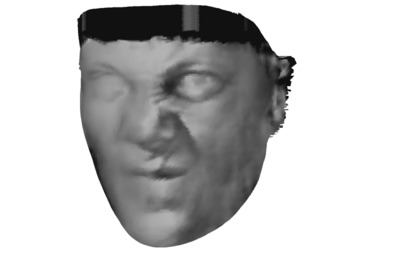}
\centering
\end{subfigure}
\begin{subfigure}{\sze\linewidth}
\includegraphics[trim={\szf cm 0 \szf cm 0},clip,width=\linewidth]{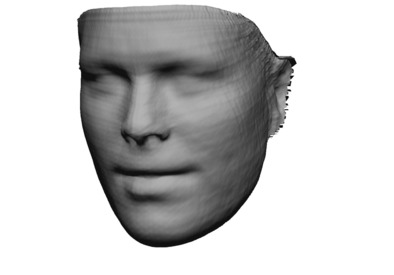}
\centering
\end{subfigure}
\begin{subfigure}{\sze\linewidth}
\includegraphics[trim={\szf cm 0 \szf cm 0},clip,width=\linewidth]{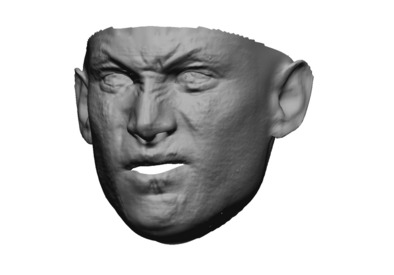}
\centering
\end{subfigure}
\begin{subfigure}{\sze\linewidth}
\includegraphics[trim={\szf cm 0 \szf cm 0},clip,width=\linewidth]{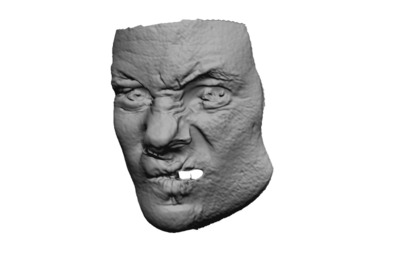}
\centering
\end{subfigure}
\begin{subfigure}{\sze\linewidth}
\includegraphics[trim={\szf cm 0 \szf cm 0},clip,width=\linewidth]{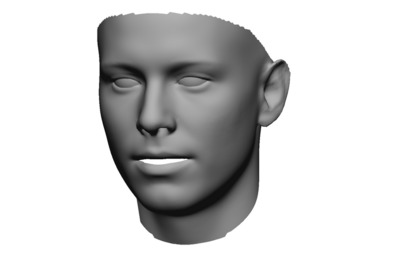}
\centering
\end{subfigure}

\begin{subfigure}{\sze\linewidth}
\includegraphics[width=\linewidth]{figures/im/25}
\centering
\end{subfigure}
\begin{subfigure}{\sze\linewidth}
\includegraphics[trim={\szf cm 0 \szf cm 0},clip,width=\linewidth]{figures/res/ours/25}
\centering
\end{subfigure}
\begin{subfigure}{\sze\linewidth}
\includegraphics[trim={\szf cm 0 \szf cm 0},clip,width=\linewidth]{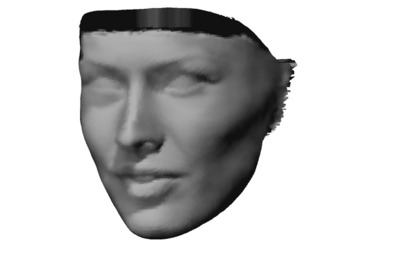}
\centering
\end{subfigure}
\begin{subfigure}{\sze\linewidth}
\includegraphics[trim={\szf cm 0 \szf cm 0},clip,width=\linewidth]{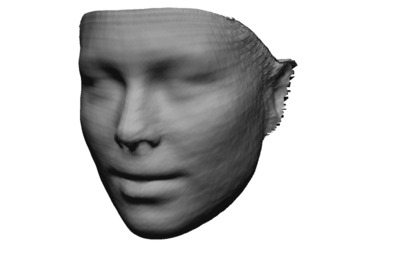}
\centering
\end{subfigure}
\begin{subfigure}{\sze\linewidth}
\includegraphics[trim={\szf cm 0 \szf cm 0},clip,width=\linewidth]{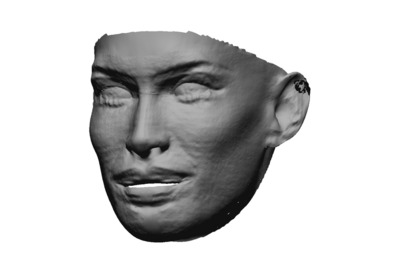}
\centering
\end{subfigure}
\begin{subfigure}{\sze\linewidth}
\includegraphics[trim={\szf cm 0 \szf cm 0},clip,width=\linewidth]{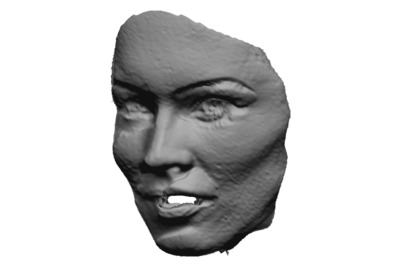}
\centering
\end{subfigure}
\begin{subfigure}{\sze\linewidth}
\includegraphics[trim={\szf cm 0 \szf cm 0},clip,width=\linewidth]{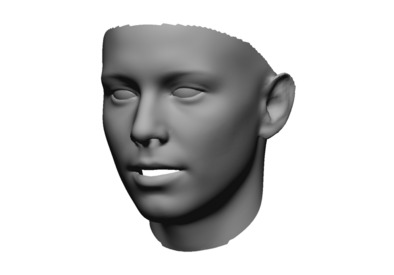}
\centering
\end{subfigure}

\begin{subfigure}{\sze\linewidth}
\includegraphics[width=\linewidth]{figures/im/29}
\centering
\end{subfigure}
\begin{subfigure}{\sze\linewidth}
\includegraphics[trim={\szf cm 0 \szf cm 0},clip,width=\linewidth]{figures/res/ours/29}
\centering
\end{subfigure}
\begin{subfigure}{\sze\linewidth}
\includegraphics[trim={\szf cm 0 \szf cm 0},clip,width=\linewidth]{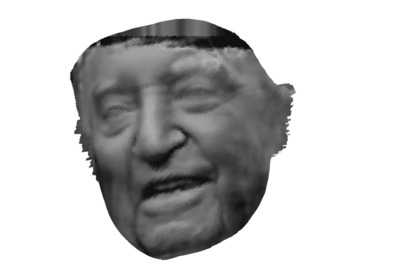}
\centering
\end{subfigure}
\begin{subfigure}{\sze\linewidth}
\includegraphics[trim={\szf cm 0 \szf cm 0},clip,width=\linewidth]{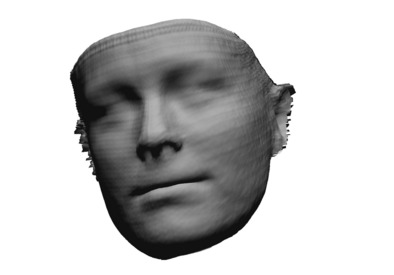}
\centering
\end{subfigure}
\begin{subfigure}{\sze\linewidth}
\includegraphics[trim={\szf cm 0 \szf cm 0},clip,width=\linewidth]{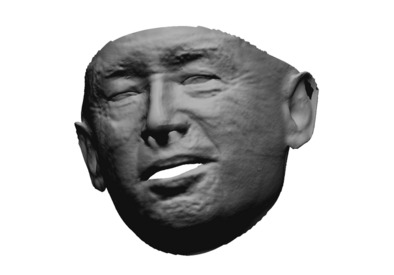}
\centering
\end{subfigure}
\begin{subfigure}{\sze\linewidth}
\includegraphics[trim={\szf cm 0 \szf cm 0},clip,width=\linewidth]{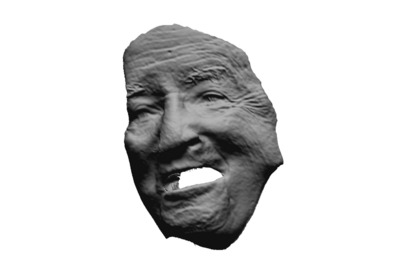}
\centering
\end{subfigure}
\begin{subfigure}{\sze\linewidth}
\includegraphics[trim={\szf cm 0 \szf cm 0},clip,width=\linewidth]{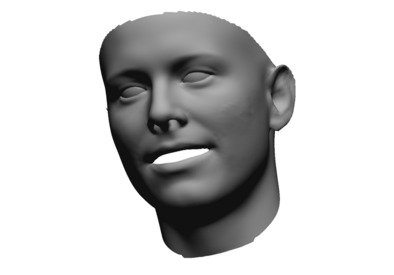}
\centering
\end{subfigure}

\begin{subfigure}{\sze\linewidth}
\includegraphics[width=\linewidth]{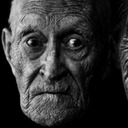}
\centering
\end{subfigure}
\begin{subfigure}{\sze\linewidth}
\includegraphics[trim={\szf cm 0 \szf cm 0},clip,width=\linewidth]{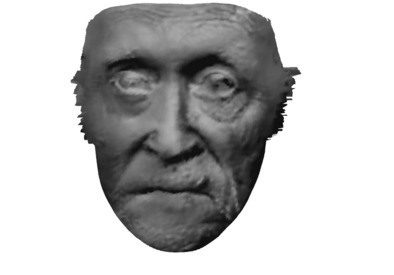}
\centering
\end{subfigure}
\begin{subfigure}{\sze\linewidth}
\includegraphics[trim={\szf cm 0 \szf cm 0},clip,width=\linewidth]{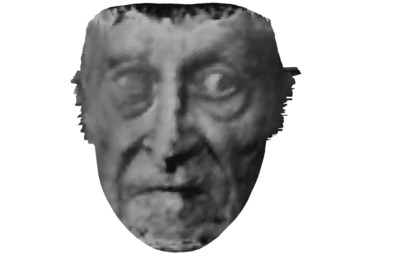}
\centering
\end{subfigure}
\begin{subfigure}{\sze\linewidth}
\includegraphics[trim={\szf cm 0 \szf cm 0},clip,width=\linewidth]{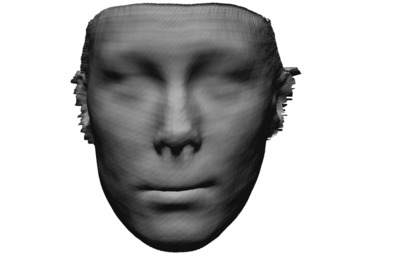}
\centering
\end{subfigure}
\begin{subfigure}{\sze\linewidth}
\includegraphics[trim={\szf cm 0 \szf cm 0},clip,width=\linewidth]{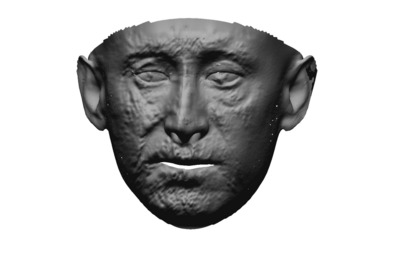}
\centering
\end{subfigure}
\begin{subfigure}{\sze\linewidth}
\includegraphics[trim={\szf cm 0 \szf cm 0},clip,width=\linewidth]{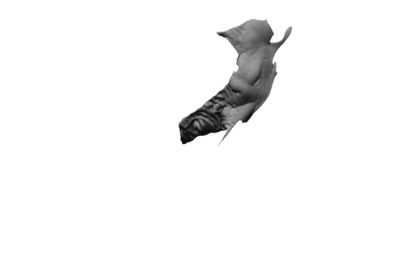}
\centering
\end{subfigure}
\begin{subfigure}{\sze\linewidth}
\includegraphics[trim={\szf cm 0 \szf cm 0},clip,width=\linewidth]{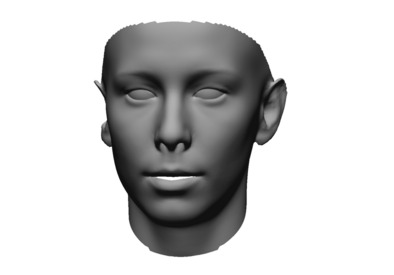}
\centering
\end{subfigure}

\begin{subfigure}{\sze\linewidth}
\includegraphics[width=\linewidth]{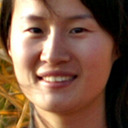}
\centering
\end{subfigure}
\begin{subfigure}{\sze\linewidth}
\includegraphics[trim={\szf cm 0 \szf cm 0},clip,width=\linewidth]{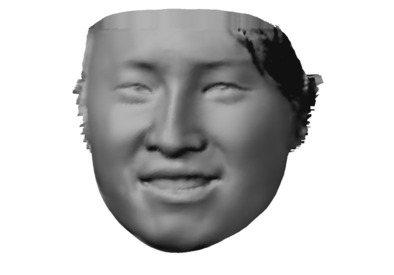}
\centering
\end{subfigure}
\begin{subfigure}{\sze\linewidth}
\includegraphics[trim={\szf cm 0 \szf cm 0},clip,width=\linewidth]{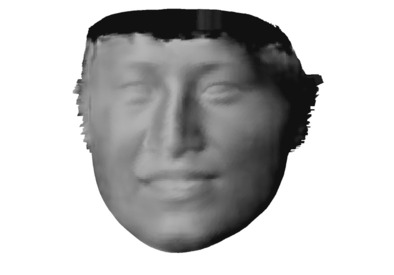}
\centering
\end{subfigure}
\begin{subfigure}{\sze\linewidth}
\includegraphics[trim={\szf cm 0 \szf cm 0},clip,width=\linewidth]{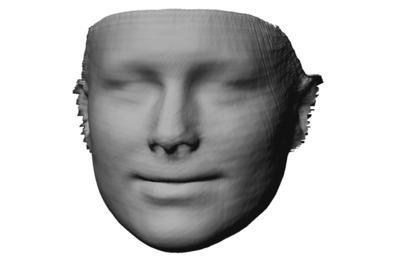}
\centering
\end{subfigure}
\begin{subfigure}{\sze\linewidth}
\includegraphics[trim={\szf cm 0 \szf cm 0},clip,width=\linewidth]{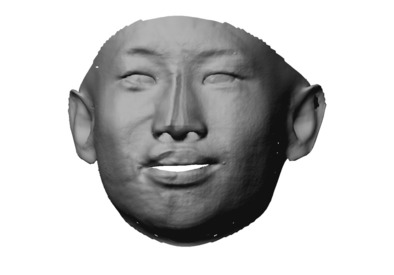}
\centering
\end{subfigure}
\begin{subfigure}{\sze\linewidth}
\includegraphics[trim={\szf cm 0 \szf cm 0},clip,width=\linewidth]{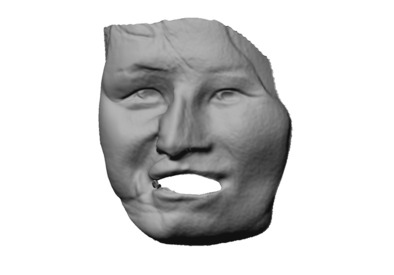}
\centering
\end{subfigure}
\begin{subfigure}{\sze\linewidth}
\includegraphics[trim={\szf cm 0 \szf cm 0},clip,width=\linewidth]{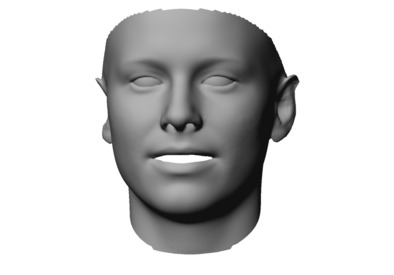}
\centering
\end{subfigure}

\begin{subfigure}{\sze\linewidth}
\includegraphics[width=\linewidth]{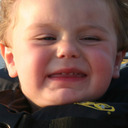}
\centering
\captionsetup{labelformat=empty}
\caption{\scriptsize Input}
\end{subfigure}
\begin{subfigure}{\sze\linewidth}
\includegraphics[trim={\szf cm 0 \szf cm 0},clip,width=\linewidth]{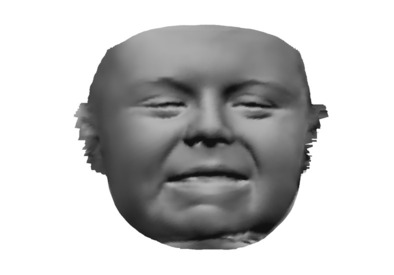}
\centering
\captionsetup{labelformat=empty}
\caption{\scriptsize Ours+PRN}
\end{subfigure}
\begin{subfigure}{\sze\linewidth}
\includegraphics[trim={\szf cm 0 \szf cm 0},clip,width=\linewidth]{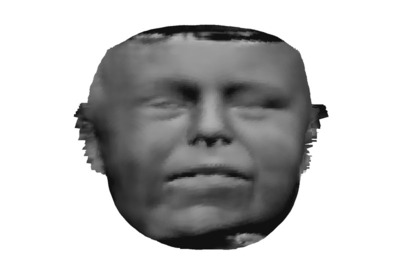}
\centering
\captionsetup{labelformat=empty}
\caption{\scriptsize SfSNet+PRN}
\end{subfigure}
\begin{subfigure}{\sze\linewidth}
\includegraphics[trim={\szf cm 0 \szf cm 0},clip,width=\linewidth]{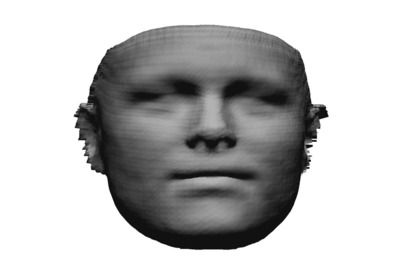}
\centering
\captionsetup{labelformat=empty}
\caption{\scriptsize PRN}
\end{subfigure}
\begin{subfigure}{\sze\linewidth}
\includegraphics[trim={\szf cm 0 \szf cm 0},clip,width=\linewidth]{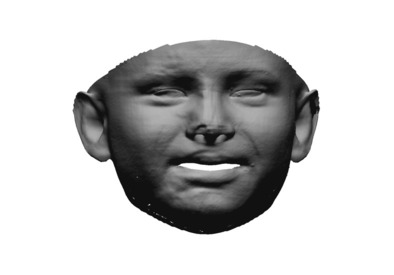}
\centering
\captionsetup{labelformat=empty}
\caption{\scriptsize Extreme}
\end{subfigure}
\begin{subfigure}{\sze\linewidth}
\includegraphics[trim={\szf cm 0 \szf cm 0},clip,width=\linewidth]{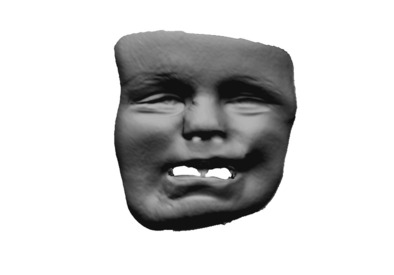}
\centering
\captionsetup{labelformat=empty}
\caption{\scriptsize Pix2V}
\end{subfigure}
\begin{subfigure}{\sze\linewidth}
\includegraphics[trim={\szf cm 0 \szf cm 0},clip,width=\linewidth]{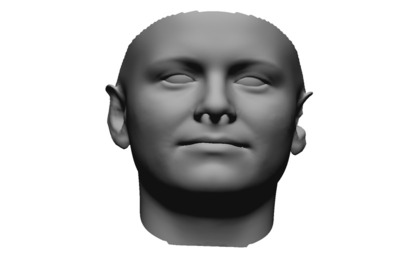}
\centering
\captionsetup{labelformat=empty}
\caption{\scriptsize 3DDFA}
\end{subfigure}

\end{figure*}
%
%
\begin{figure*}[h!]\ContinuedFloat
\centering

\begin{subfigure}{\sze\linewidth}
\includegraphics[width=\linewidth]{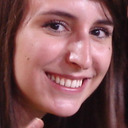}
\centering
\end{subfigure}
\begin{subfigure}{\sze\linewidth}
\includegraphics[trim={\szf cm 0 \szf cm 0},clip,width=\linewidth]{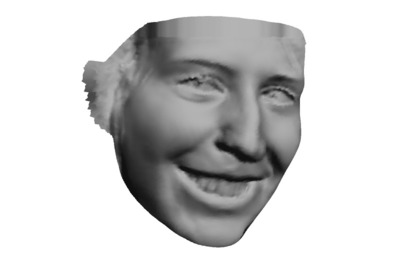}
\centering
\end{subfigure}
\begin{subfigure}{\sze\linewidth}
\includegraphics[trim={\szf cm 0 \szf cm 0},clip,width=\linewidth]{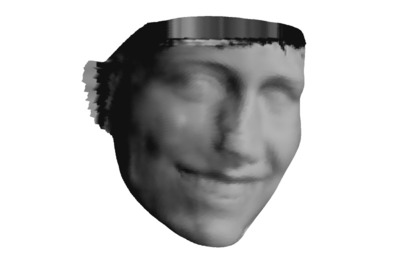}
\centering
\end{subfigure}
\begin{subfigure}{\sze\linewidth}
\includegraphics[trim={\szf cm 0 \szf cm 0},clip,width=\linewidth]{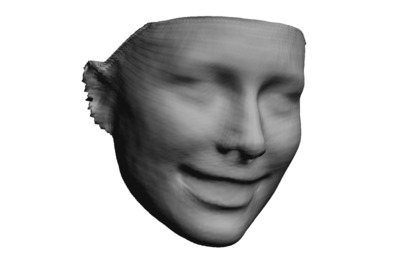}
\centering
\end{subfigure}
\begin{subfigure}{\sze\linewidth}
\includegraphics[trim={\szf cm 0 \szf cm 0},clip,width=\linewidth]{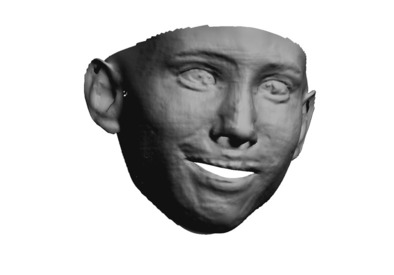}
\centering
\end{subfigure}
\begin{subfigure}{\sze\linewidth}
\includegraphics[trim={\szf cm 0 \szf cm 0},clip,width=\linewidth]{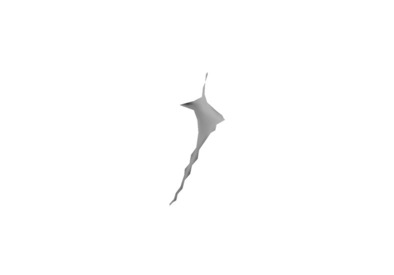}
\centering
\end{subfigure}
\begin{subfigure}{\sze\linewidth}
\includegraphics[trim={\szf cm 0 \szf cm 0},clip,width=\linewidth]{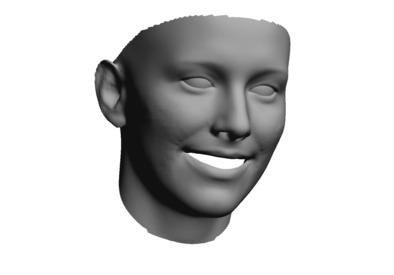}
\centering
\end{subfigure}

\begin{subfigure}{\sze\linewidth}
\includegraphics[width=\linewidth]{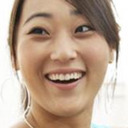}
\centering
\end{subfigure}
\begin{subfigure}{\sze\linewidth}
\includegraphics[trim={\szf cm 0 \szf cm 0},clip,width=\linewidth]{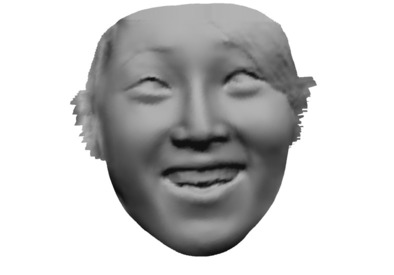}
\centering
\end{subfigure}
\begin{subfigure}{\sze\linewidth}
\includegraphics[trim={\szf cm 0 \szf cm 0},clip,width=\linewidth]{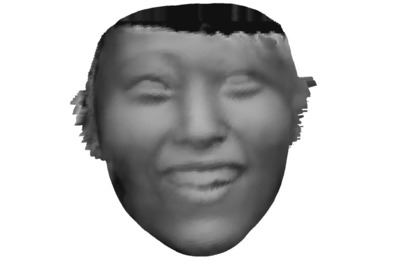}
\centering
\end{subfigure}
\begin{subfigure}{\sze\linewidth}
\includegraphics[trim={\szf cm 0 \szf cm 0},clip,width=\linewidth]{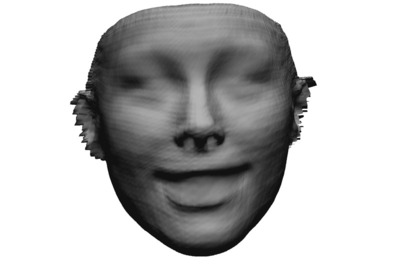}
\centering
\end{subfigure}
\begin{subfigure}{\sze\linewidth}
\includegraphics[trim={1.9cm 0.5cm 1.9cm 0.5cm},clip,width=\linewidth]{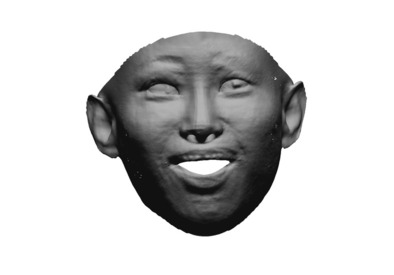}
\centering
\end{subfigure}
\begin{subfigure}{\sze\linewidth}
\includegraphics[trim={\szf cm 0 \szf cm 0},clip,width=\linewidth]{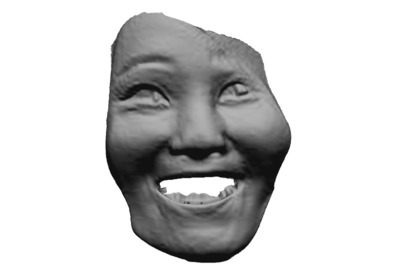}
\centering
\end{subfigure}
\begin{subfigure}{\sze\linewidth}
\includegraphics[trim={\szf cm 0 \szf cm 0},clip,width=\linewidth]{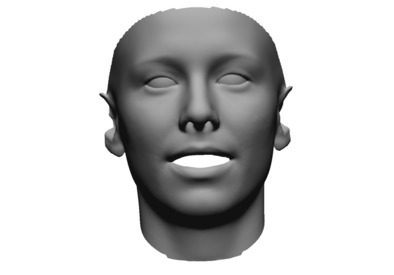}
\centering
\end{subfigure}

\begin{subfigure}{\sze\linewidth}
\includegraphics[width=\linewidth]{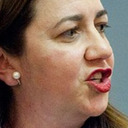}
\centering
\end{subfigure}
\begin{subfigure}{\sze\linewidth}
\includegraphics[trim={1.75cm 0.5cm 1.75cm 0.5cm},clip,width=\linewidth]{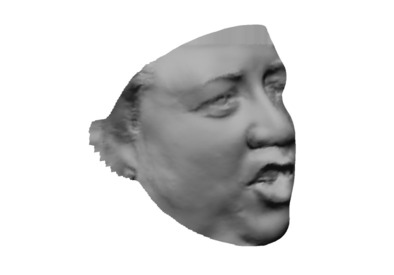}
\centering
\end{subfigure}
\begin{subfigure}{\sze\linewidth}
\includegraphics[trim={1.75cm 0.5cm 1.75cm 0.5cm},clip,width=\linewidth]{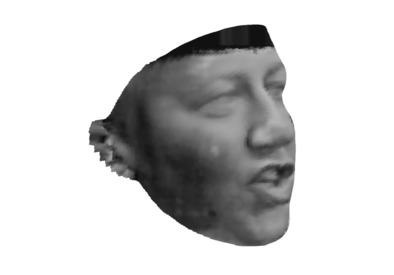}
\centering
\end{subfigure}
\begin{subfigure}{\sze\linewidth}
\includegraphics[trim={1.75cm 0.5cm 1.75cm 0.5cm},clip,width=\linewidth]{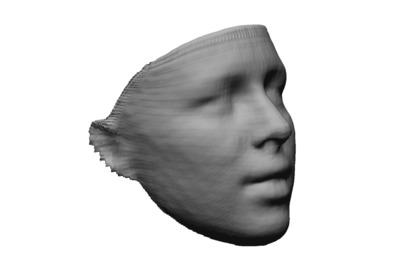}
\centering
\end{subfigure}
\begin{subfigure}{\sze\linewidth}
\includegraphics[trim={1.75cm 0.5cm 1.75cm 0.5cm},clip,width=\linewidth]{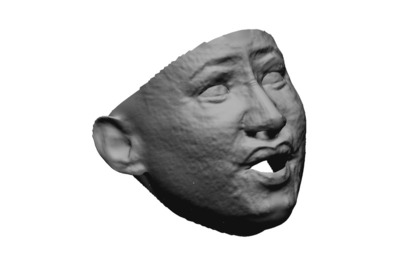}
\centering
\end{subfigure}
\begin{subfigure}{\sze\linewidth}
\includegraphics[trim={\szf cm 0 \szf cm 0},clip,width=\linewidth]{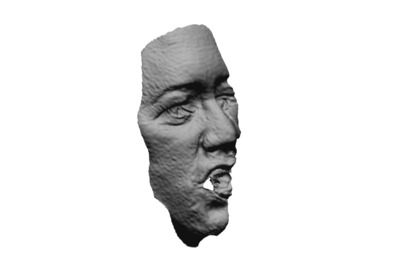}
\centering
\end{subfigure}
\begin{subfigure}{\sze\linewidth}
\includegraphics[trim={1.6cm 0.5cm 1.6cm 0.5cm},clip,width=\linewidth]{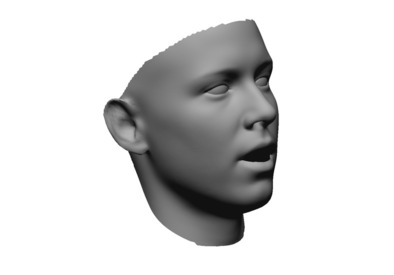}
\centering
\end{subfigure}

\begin{subfigure}{\sze\linewidth}
\includegraphics[width=\linewidth]{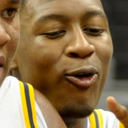}
\centering
\end{subfigure}
\begin{subfigure}{\sze\linewidth}
\includegraphics[trim={\szf cm 0 \szf cm 0},clip,width=\linewidth]{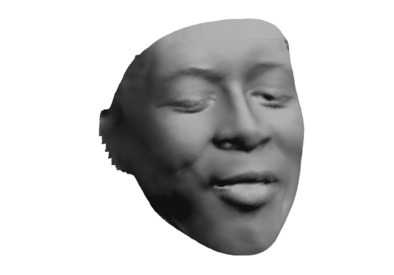}
\centering
\end{subfigure}
\begin{subfigure}{\sze\linewidth}
\includegraphics[trim={\szf cm 0 \szf cm 0},clip,width=\linewidth]{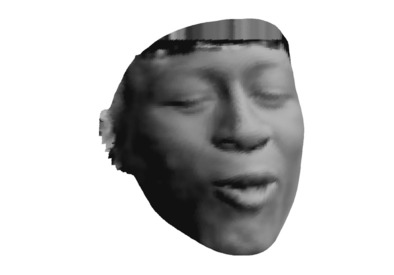}
\centering
\end{subfigure}
\begin{subfigure}{\sze\linewidth}
\includegraphics[trim={\szf cm 0 \szf cm 0},clip,width=\linewidth]{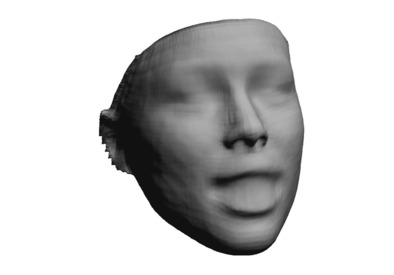}
\centering
\end{subfigure}
\begin{subfigure}{\sze\linewidth}
\includegraphics[trim={\szf cm 0 \szf cm 0},clip,width=\linewidth]{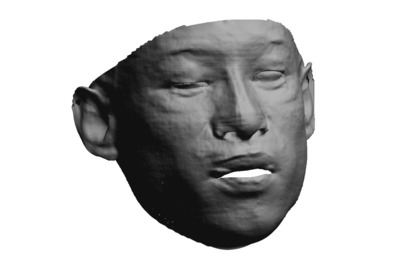}
\centering
\end{subfigure}
\begin{subfigure}{\sze\linewidth}
\includegraphics[trim={\szf cm 0 \szf cm 0},clip,width=\linewidth]{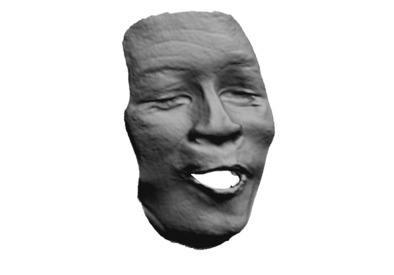}
\centering
\end{subfigure}
\begin{subfigure}{\sze\linewidth}
\includegraphics[trim={\szf cm 0 \szf cm 0},clip,width=\linewidth]{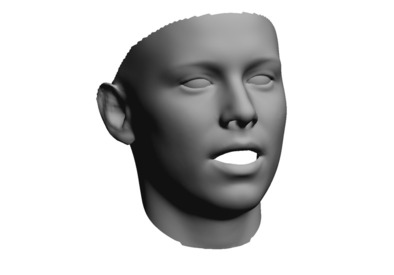}
\centering
\end{subfigure}

\begin{subfigure}{\sze\linewidth}
\includegraphics[width=\linewidth]{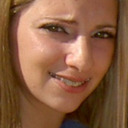}
\centering
\end{subfigure}
\begin{subfigure}{\sze\linewidth}
\includegraphics[trim={\szf cm 0 \szf cm 0},clip,width=\linewidth]{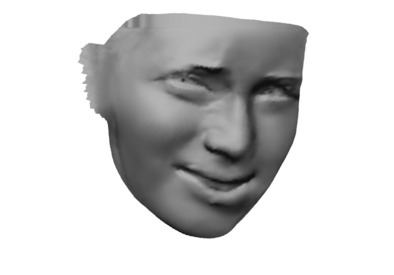}
\centering
\end{subfigure}
\begin{subfigure}{\sze\linewidth}
\includegraphics[trim={\szf cm 0 \szf cm 0},clip,width=\linewidth]{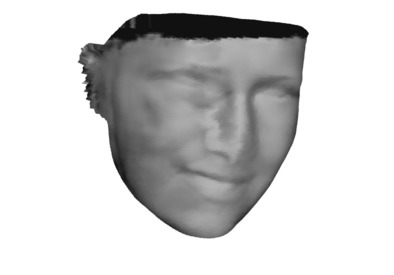}
\centering
\end{subfigure}
\begin{subfigure}{\sze\linewidth}
\includegraphics[trim={\szf cm 0 \szf cm 0},clip,width=\linewidth]{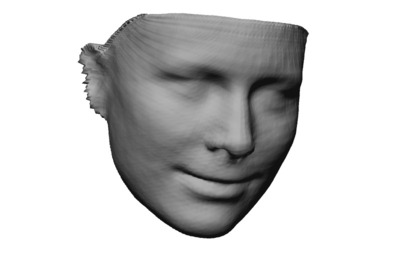}
\centering
\end{subfigure}
\begin{subfigure}{\sze\linewidth}
\includegraphics[trim={\szf cm 0 \szf cm 0},clip,width=\linewidth]{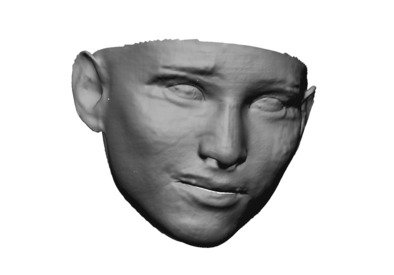}
\centering
\end{subfigure}
\begin{subfigure}{\sze\linewidth}
\includegraphics[trim={\szf cm 0 \szf cm 0},clip,width=\linewidth]{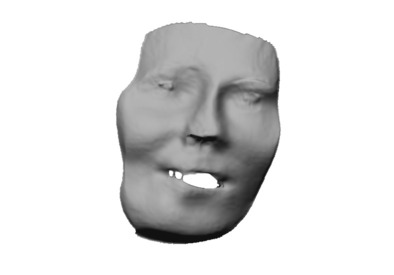}
\centering
\end{subfigure}
\begin{subfigure}{\sze\linewidth}
\includegraphics[trim={\szf cm 0 \szf cm 0},clip,width=\linewidth]{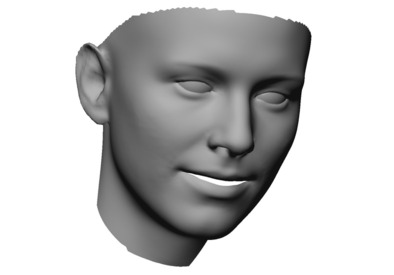}
\centering
\end{subfigure}

\begin{subfigure}{\sze\linewidth}
\includegraphics[width=\linewidth]{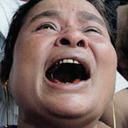}
\centering
\end{subfigure}
\begin{subfigure}{\sze\linewidth}
\includegraphics[trim={\szf cm 0 \szf cm 0},clip,width=\linewidth]{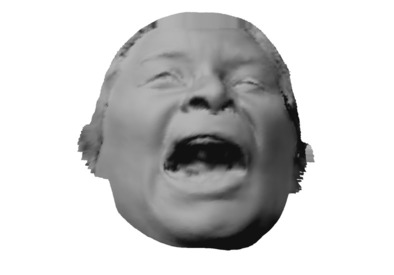}
\centering
\end{subfigure}
\begin{subfigure}{\sze\linewidth}
\includegraphics[trim={\szf cm 0 \szf cm 0},clip,width=\linewidth]{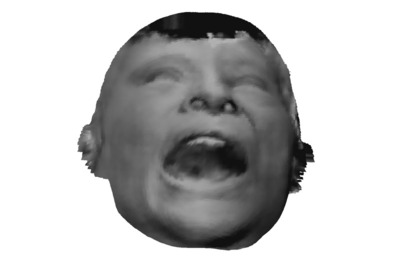}
\centering
\end{subfigure}
\begin{subfigure}{\sze\linewidth}
\includegraphics[trim={\szf cm 0 \szf cm 0},clip,width=\linewidth]{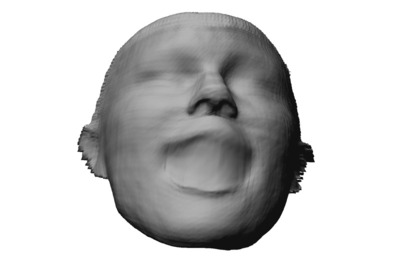}
\centering
\end{subfigure}
\begin{subfigure}{\sze\linewidth}
\includegraphics[trim={\szf cm 0 \szf cm 0},clip,width=\linewidth]{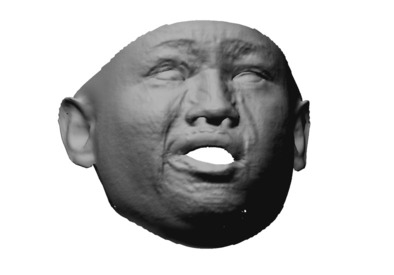}
\centering
\end{subfigure}
\begin{subfigure}{\sze\linewidth}
\includegraphics[trim={\szf cm 0 \szf cm 0},clip,width=\linewidth]{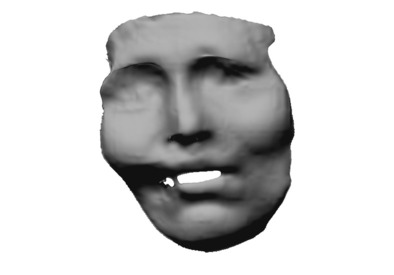}
\centering
\end{subfigure}
\begin{subfigure}{\sze\linewidth}
\includegraphics[trim={\szf cm 0 \szf cm 0},clip,width=\linewidth]{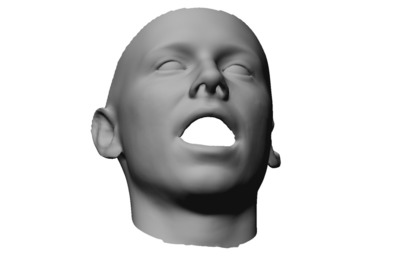}
\centering
\end{subfigure}

\begin{subfigure}{\sze\linewidth}
\includegraphics[width=\linewidth]{figures/im/11}
\centering
\end{subfigure}
\begin{subfigure}{\sze\linewidth}
\includegraphics[trim={\szf cm 0 \szf cm 0},clip,width=\linewidth]{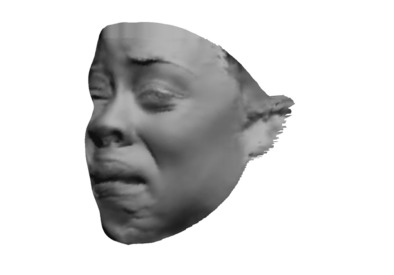}
\centering
\end{subfigure}
\begin{subfigure}{\sze\linewidth}
\includegraphics[trim={\szf cm 0 \szf cm 0},clip,width=\linewidth]{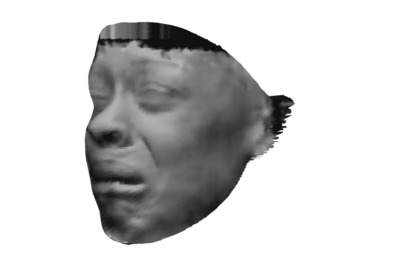}
\centering
\end{subfigure}
\begin{subfigure}{\sze\linewidth}
\includegraphics[trim={\szf cm 0 \szf cm 0},clip,width=\linewidth]{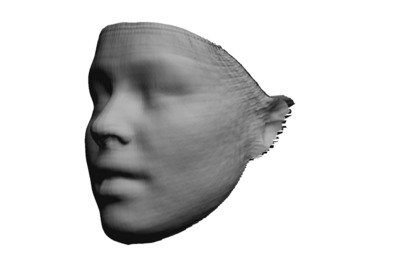}
\centering
\end{subfigure}
\begin{subfigure}{\sze\linewidth}
\includegraphics[trim={1.4cm 0 1.4cm 0},clip,width=\linewidth]{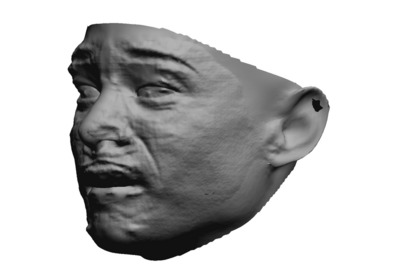}
\centering
\end{subfigure}
\begin{subfigure}{\sze\linewidth}
\includegraphics[trim={\szf cm 0 \szf cm 0},clip,width=\linewidth]{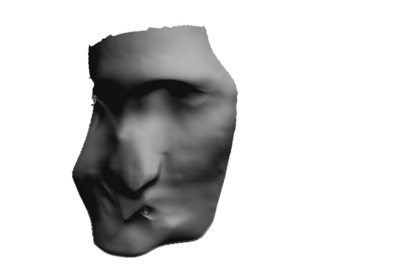}
\centering
\end{subfigure}
\begin{subfigure}{\sze\linewidth}
\includegraphics[trim={\szf cm 0 \szf cm 0},clip,width=\linewidth]{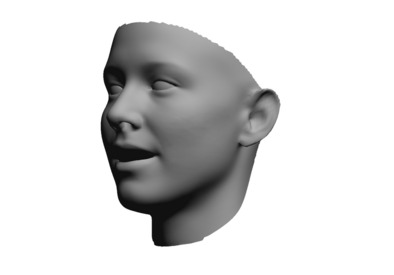}
\centering
\end{subfigure}

\begin{subfigure}{\sze\linewidth}
\includegraphics[width=\linewidth]{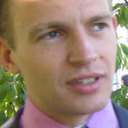}
\centering
\end{subfigure}
\begin{subfigure}{\sze\linewidth}
\includegraphics[trim={\szf cm 0 \szf cm 0},clip,width=\linewidth]{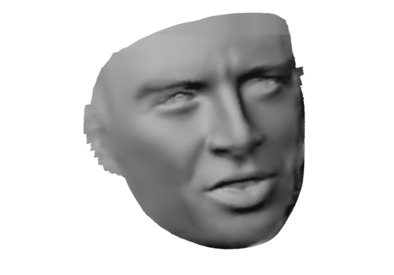}
\centering
\end{subfigure}
\begin{subfigure}{\sze\linewidth}
\includegraphics[trim={\szf cm 0 \szf cm 0},clip,width=\linewidth]{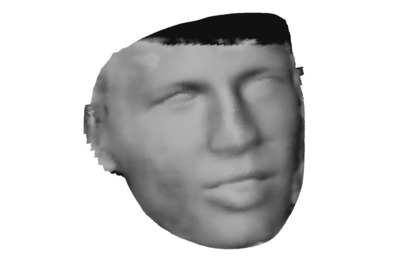}
\centering
\end{subfigure}
\begin{subfigure}{\sze\linewidth}
\includegraphics[trim={\szf cm 0 \szf cm 0},clip,width=\linewidth]{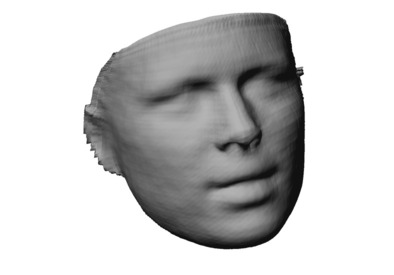}
\centering
\end{subfigure}
\begin{subfigure}{\sze\linewidth}
\includegraphics[trim={\szf cm 0 \szf cm 0},clip,width=\linewidth]{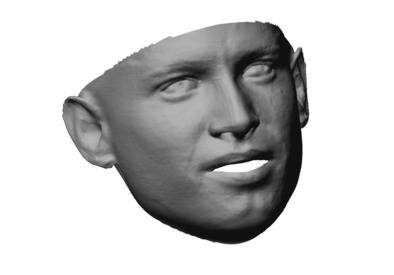}
\centering
\end{subfigure}
\begin{subfigure}{\sze\linewidth}
\includegraphics[trim={\szf cm 0 \szf cm 0},clip,width=\linewidth]{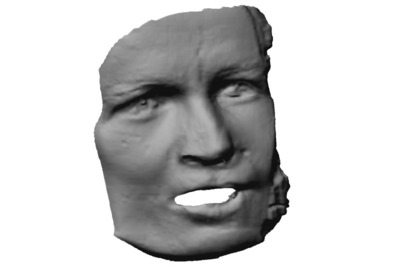}
\centering
\end{subfigure}
\begin{subfigure}{\sze\linewidth}
\includegraphics[trim={\szf cm 0 \szf cm 0},clip,width=\linewidth]{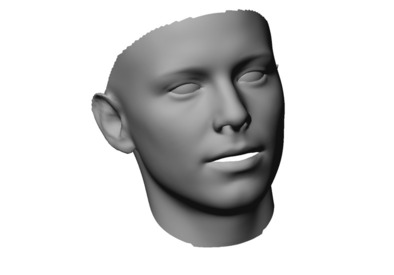}
\centering
\end{subfigure}

\begin{subfigure}{\sze\linewidth}
\includegraphics[width=\linewidth]{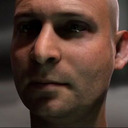}
\centering
\end{subfigure}
\begin{subfigure}{\sze\linewidth}
\includegraphics[trim={\szf cm 0 \szf cm 0},clip,width=\linewidth]{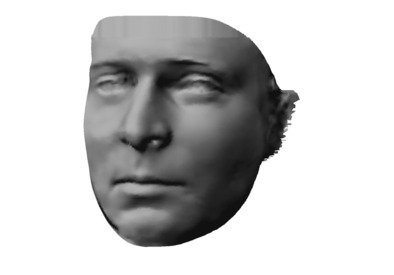}
\centering
\end{subfigure}
\begin{subfigure}{\sze\linewidth}
\includegraphics[trim={\szf cm 0 \szf cm 0},clip,width=\linewidth]{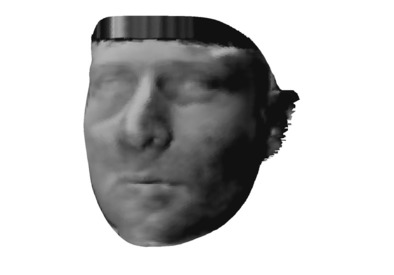}
\centering
\end{subfigure}
\begin{subfigure}{\sze\linewidth}
\includegraphics[trim={\szf cm 0 \szf cm 0},clip,width=\linewidth]{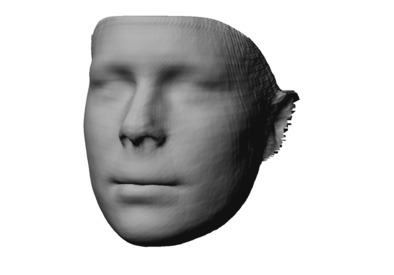}
\centering
\end{subfigure}
\begin{subfigure}{\sze\linewidth}
\includegraphics[trim={\szf cm 0 \szf cm 0},clip,width=\linewidth]{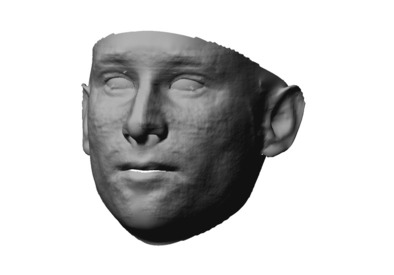}
\centering
\end{subfigure}
\begin{subfigure}{\sze\linewidth}
\includegraphics[trim={\szf cm 0 \szf cm 0},clip,width=\linewidth]{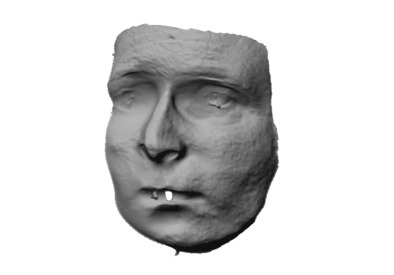}
\centering
\end{subfigure}
\begin{subfigure}{\sze\linewidth}
\includegraphics[trim={\szf cm 0 \szf cm 0},clip,width=\linewidth]{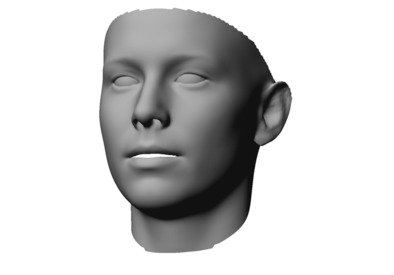}
\centering
\end{subfigure}

\begin{subfigure}{\sze\linewidth}
\includegraphics[width=\linewidth]{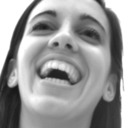}
\centering
\end{subfigure}
\begin{subfigure}{\sze\linewidth}
\includegraphics[trim={\szf cm 0 \szf cm 0},clip,width=\linewidth]{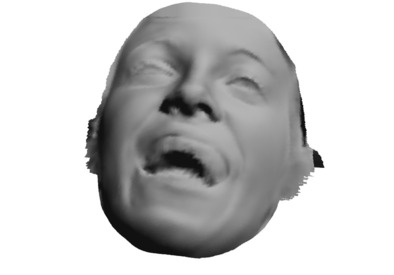}
\centering
\end{subfigure}
\begin{subfigure}{\sze\linewidth}
\includegraphics[trim={\szf cm 0 \szf cm 0},clip,width=\linewidth]{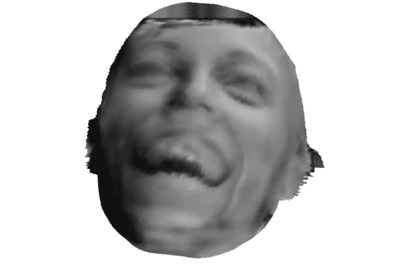}
\centering
\end{subfigure}
\begin{subfigure}{\sze\linewidth}
\includegraphics[trim={\szf cm 0 \szf cm 0},clip,width=\linewidth]{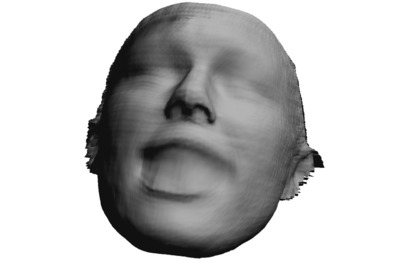}
\centering
\end{subfigure}
\begin{subfigure}{\sze\linewidth}
\includegraphics[trim={\szf cm 0 \szf cm 0},clip,width=\linewidth]{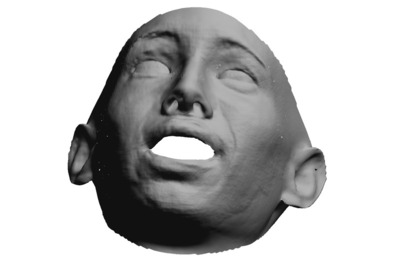}
\centering
\end{subfigure}
\begin{subfigure}{\sze\linewidth}
\includegraphics[trim={\szf cm 0 \szf cm 0},clip,width=\linewidth]{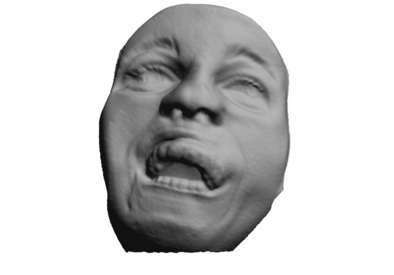}
\centering
\end{subfigure}
\begin{subfigure}{\sze\linewidth}
\includegraphics[trim={\szf cm 0 \szf cm 0},clip,width=\linewidth]{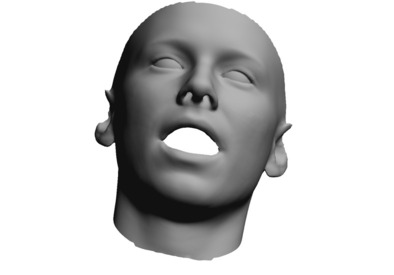}
\centering
\end{subfigure}

\begin{subfigure}{\sze\linewidth}
\includegraphics[width=\linewidth]{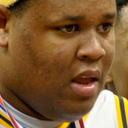}
\centering
\captionsetup{labelformat=empty}
\caption{\scriptsize Input}
\end{subfigure}
\begin{subfigure}{\sze\linewidth}
\includegraphics[trim={\szf cm 0 \szf cm 0},clip,width=\linewidth]{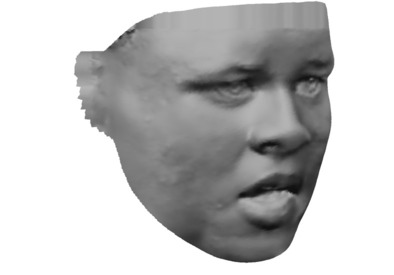}
\centering
\captionsetup{labelformat=empty}
\caption{\scriptsize Ours+PRN}
\end{subfigure}
\begin{subfigure}{\sze\linewidth}
\includegraphics[trim={\szf cm 0 \szf cm 0},clip,width=\linewidth]{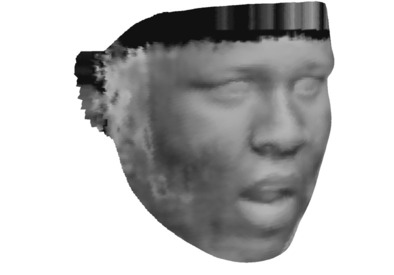}
\centering
\captionsetup{labelformat=empty}
\caption{\scriptsize SfSNet+PRN}
\end{subfigure}
\begin{subfigure}{\sze\linewidth}
\includegraphics[trim={\szf cm 0 \szf cm 0},clip,width=\linewidth]{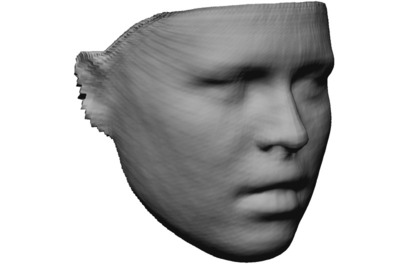}
\centering
\captionsetup{labelformat=empty}
\caption{\scriptsize PRN}
\end{subfigure}
\begin{subfigure}{\sze\linewidth}
\includegraphics[trim={1cm 0 1.5cm 0},clip,width=\linewidth]{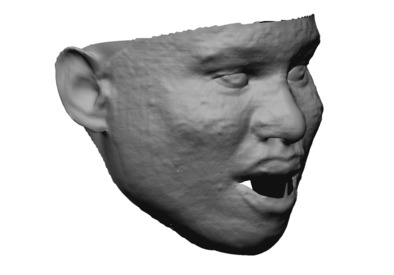}
\centering
\captionsetup{labelformat=empty}
\caption{\scriptsize Extreme}
\end{subfigure}
\begin{subfigure}{\sze\linewidth}
\includegraphics[trim={\szf cm 0 \szf cm 0},clip,width=\linewidth]{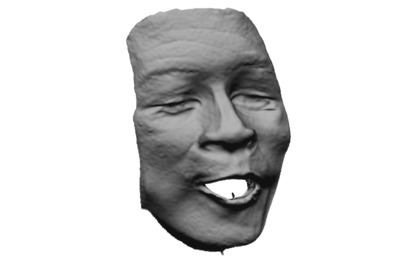}
\centering
\captionsetup{labelformat=empty}
\caption{\scriptsize Pix2V}
\end{subfigure}
\begin{subfigure}{\sze\linewidth}
\includegraphics[trim={\szf cm 0 \szf cm 0},clip,width=\linewidth]{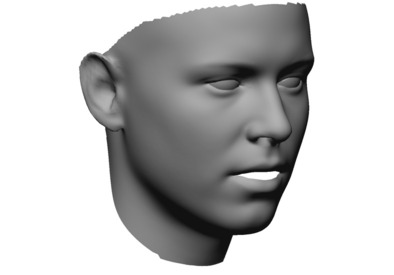}
\centering
\captionsetup{labelformat=empty}
\caption{\scriptsize 3DDFA}
\end{subfigure}

\caption{Qualitative comparisons with geometries in the 300-W dataset~\cite{Sagonas13}.}
\label{fig:comparison_mesh2}
\end{figure*}

\clearpage
\begin{figure*}[h!]
\centering

\begin{subfigure}{\sze\linewidth}
\includegraphics[width=\linewidth]{figures/im/41}
\centering
\end{subfigure}
\begin{subfigure}{\sze\linewidth}
\includegraphics[width=\linewidth]{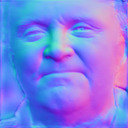}
\centering
\end{subfigure}
\begin{subfigure}{\sze\linewidth}
\includegraphics[width=\linewidth]{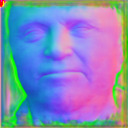}
\centering
\end{subfigure}
\begin{subfigure}{\sze\linewidth}
\includegraphics[width=\linewidth]{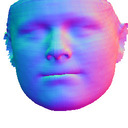}
\centering
\end{subfigure}
\begin{subfigure}{\sze\linewidth}
\includegraphics[width=\linewidth]{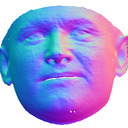}
\centering
\end{subfigure}
\begin{subfigure}{\sze\linewidth}
\includegraphics[width=\linewidth]{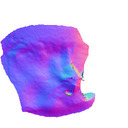}
\centering
\end{subfigure}
\begin{subfigure}{\sze\linewidth}
\includegraphics[width=\linewidth]{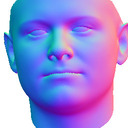}
\centering
\end{subfigure}

\begin{subfigure}{\sze\linewidth}
\includegraphics[width=\linewidth]{figures/im/3}
\centering
\end{subfigure}
\begin{subfigure}{\sze\linewidth}
\includegraphics[width=\linewidth]{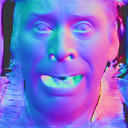}
\centering
\end{subfigure}
\begin{subfigure}{\sze\linewidth}
\includegraphics[width=\linewidth]{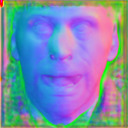}
\centering
\end{subfigure}
\begin{subfigure}{\sze\linewidth}
\includegraphics[width=\linewidth]{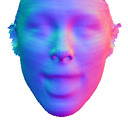}
\centering
\end{subfigure}
\begin{subfigure}{\sze\linewidth}
\includegraphics[width=\linewidth]{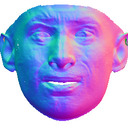}
\centering
\end{subfigure}
\begin{subfigure}{\sze\linewidth}
\includegraphics[width=\linewidth]{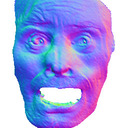}
\centering
\end{subfigure}
\begin{subfigure}{\sze\linewidth}
\includegraphics[width=\linewidth]{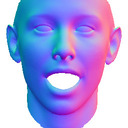}
\centering
\end{subfigure}

\begin{subfigure}{\sze\linewidth}
\includegraphics[width=\linewidth]{figures/im/5}
\centering
\end{subfigure}
\begin{subfigure}{\sze\linewidth}
\includegraphics[width=\linewidth]{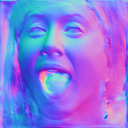}
\centering
\end{subfigure}
\begin{subfigure}{\sze\linewidth}
\includegraphics[width=\linewidth]{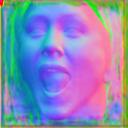}
\centering
\end{subfigure}
\begin{subfigure}{\sze\linewidth}
\includegraphics[width=\linewidth]{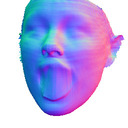}
\centering
\end{subfigure}
\begin{subfigure}{\sze\linewidth}
\includegraphics[width=\linewidth]{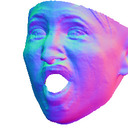}
\centering
\end{subfigure}
\begin{subfigure}{\sze\linewidth}
\includegraphics[width=\linewidth]{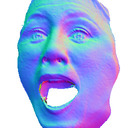}
\centering
\end{subfigure}
\begin{subfigure}{\sze\linewidth}
\includegraphics[width=\linewidth]{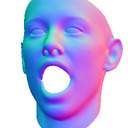}
\centering
\end{subfigure}

\begin{subfigure}{\sze\linewidth}
\includegraphics[width=\linewidth]{figures/im/6}
\centering
\end{subfigure}
\begin{subfigure}{\sze\linewidth}
\includegraphics[width=\linewidth]{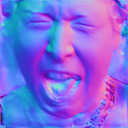}
\centering
\end{subfigure}
\begin{subfigure}{\sze\linewidth}
\includegraphics[width=\linewidth]{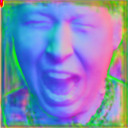}
\centering
\end{subfigure}
\begin{subfigure}{\sze\linewidth}
\includegraphics[width=\linewidth]{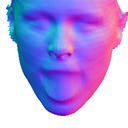}
\centering
\end{subfigure}
\begin{subfigure}{\sze\linewidth}
\includegraphics[width=\linewidth]{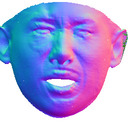}
\centering
\end{subfigure}
\begin{subfigure}{\sze\linewidth}
\includegraphics[width=\linewidth]{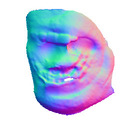}
\centering
\end{subfigure}
\begin{subfigure}{\sze\linewidth}
\includegraphics[width=\linewidth]{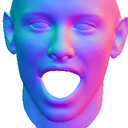}
\centering
\end{subfigure}

\begin{subfigure}{\sze\linewidth}
\includegraphics[width=\linewidth]{figures/im/12}
\centering
\end{subfigure}
\begin{subfigure}{\sze\linewidth}
\includegraphics[width=\linewidth]{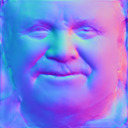}
\centering
\end{subfigure}
\begin{subfigure}{\sze\linewidth}
\includegraphics[width=\linewidth]{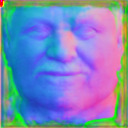}
\centering
\end{subfigure}
\begin{subfigure}{\sze\linewidth}
\includegraphics[width=\linewidth]{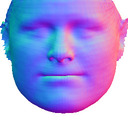}
\centering
\end{subfigure}
\begin{subfigure}{\sze\linewidth}
\includegraphics[width=\linewidth]{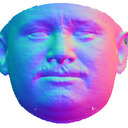}
\centering
\end{subfigure}
\begin{subfigure}{\sze\linewidth}
\includegraphics[width=\linewidth]{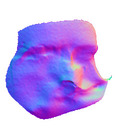}
\centering
\end{subfigure}
\begin{subfigure}{\sze\linewidth}
\includegraphics[width=\linewidth]{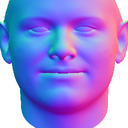}
\centering
\end{subfigure}

\begin{subfigure}{\sze\linewidth}
\includegraphics[width=\linewidth]{figures/im/30}
\centering
\end{subfigure}
\begin{subfigure}{\sze\linewidth}
\includegraphics[width=\linewidth]{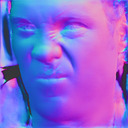}
\centering
\end{subfigure}
\begin{subfigure}{\sze\linewidth}
\includegraphics[width=\linewidth]{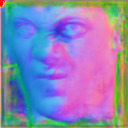}
\centering
\end{subfigure}
\begin{subfigure}{\sze\linewidth}
\includegraphics[width=\linewidth]{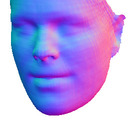}
\centering
\end{subfigure}
\begin{subfigure}{\sze\linewidth}
\includegraphics[width=\linewidth]{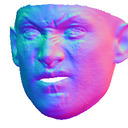}
\centering
\end{subfigure}
\begin{subfigure}{\sze\linewidth}
\includegraphics[width=\linewidth]{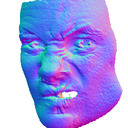}
\centering
\end{subfigure}
\begin{subfigure}{\sze\linewidth}
\includegraphics[width=\linewidth]{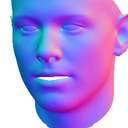}
\centering
\end{subfigure}

\begin{subfigure}{\sze\linewidth}
\includegraphics[width=\linewidth]{figures/im/25}
\centering
\end{subfigure}
\begin{subfigure}{\sze\linewidth}
\includegraphics[width=\linewidth]{figures/resn/ours/25}
\centering
\end{subfigure}
\begin{subfigure}{\sze\linewidth}
\includegraphics[width=\linewidth]{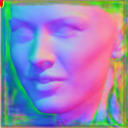}
\centering
\end{subfigure}
\begin{subfigure}{\sze\linewidth}
\includegraphics[width=\linewidth]{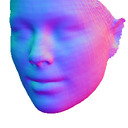}
\centering
\end{subfigure}
\begin{subfigure}{\sze\linewidth}
\includegraphics[width=\linewidth]{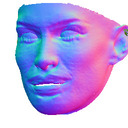}
\centering
\end{subfigure}
\begin{subfigure}{\sze\linewidth}
\includegraphics[width=\linewidth]{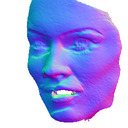}
\centering
\end{subfigure}
\begin{subfigure}{\sze\linewidth}
\includegraphics[width=\linewidth]{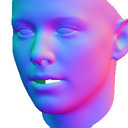}
\centering
\end{subfigure}

\begin{subfigure}{\sze\linewidth}
\includegraphics[width=\linewidth]{figures/im/29}
\centering
\end{subfigure}
\begin{subfigure}{\sze\linewidth}
\includegraphics[width=\linewidth]{figures/resn/ours/29}
\centering
\end{subfigure}
\begin{subfigure}{\sze\linewidth}
\includegraphics[width=\linewidth]{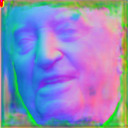}
\centering
\end{subfigure}
\begin{subfigure}{\sze\linewidth}
\includegraphics[width=\linewidth]{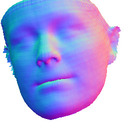}
\centering
\end{subfigure}
\begin{subfigure}{\sze\linewidth}
\includegraphics[width=\linewidth]{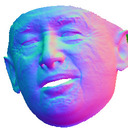}
\centering
\end{subfigure}
\begin{subfigure}{\sze\linewidth}
\includegraphics[width=\linewidth]{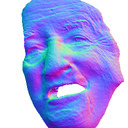}
\centering
\end{subfigure}
\begin{subfigure}{\sze\linewidth}
\includegraphics[width=\linewidth]{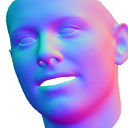}
\centering
\end{subfigure}

\begin{subfigure}{\sze\linewidth}
\includegraphics[width=\linewidth]{figures/im/21}
\centering
\end{subfigure}
\begin{subfigure}{\sze\linewidth}
\includegraphics[width=\linewidth]{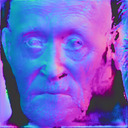}
\centering
\end{subfigure}
\begin{subfigure}{\sze\linewidth}
\includegraphics[width=\linewidth]{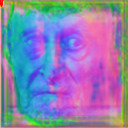}
\centering
\end{subfigure}
\begin{subfigure}{\sze\linewidth}
\includegraphics[width=\linewidth]{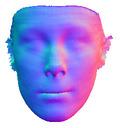}
\centering
\end{subfigure}
\begin{subfigure}{\sze\linewidth}
\includegraphics[width=\linewidth]{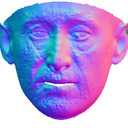}
\centering
\end{subfigure}
\begin{subfigure}{\sze\linewidth}
\includegraphics[width=\linewidth]{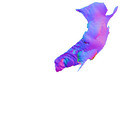}
\centering
\end{subfigure}
\begin{subfigure}{\sze\linewidth}
\includegraphics[width=\linewidth]{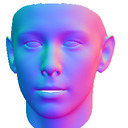}
\centering
\end{subfigure}

\begin{subfigure}{\sze\linewidth}
\includegraphics[width=\linewidth]{figures/im/27}
\centering
\end{subfigure}
\begin{subfigure}{\sze\linewidth}
\includegraphics[width=\linewidth]{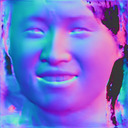}
\centering
\end{subfigure}
\begin{subfigure}{\sze\linewidth}
\includegraphics[width=\linewidth]{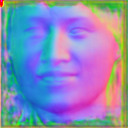}
\centering
\end{subfigure}
\begin{subfigure}{\sze\linewidth}
\includegraphics[width=\linewidth]{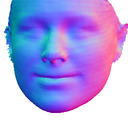}
\centering
\end{subfigure}
\begin{subfigure}{\sze\linewidth}
\includegraphics[width=\linewidth]{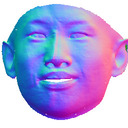}
\centering
\end{subfigure}
\begin{subfigure}{\sze\linewidth}
\includegraphics[width=\linewidth]{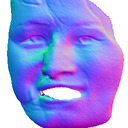}
\centering
\end{subfigure}
\begin{subfigure}{\sze\linewidth}
\includegraphics[width=\linewidth]{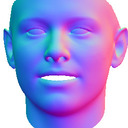}
\centering
\end{subfigure}

\begin{subfigure}{\sze\linewidth}
\includegraphics[width=\linewidth]{figures/im/24}
\centering
\captionsetup{labelformat=empty}
\caption{\scriptsize Input}
\end{subfigure}
\begin{subfigure}{\sze\linewidth}
\includegraphics[width=\linewidth]{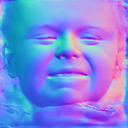}
\centering
\captionsetup{labelformat=empty}
\caption{\scriptsize Ours+PRN}
\end{subfigure}
\begin{subfigure}{\sze\linewidth}
\includegraphics[width=\linewidth]{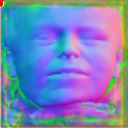}
\centering
\captionsetup{labelformat=empty}
\caption{\scriptsize SfSNet+PRN}
\end{subfigure}
\begin{subfigure}{\sze\linewidth}
\includegraphics[width=\linewidth]{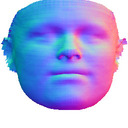}
\centering
\captionsetup{labelformat=empty}
\caption{\scriptsize PRN}
\end{subfigure}
\begin{subfigure}{\sze\linewidth}
\includegraphics[width=\linewidth]{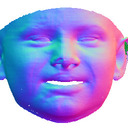}
\centering
\captionsetup{labelformat=empty}
\caption{\scriptsize Extreme}
\end{subfigure}
\begin{subfigure}{\sze\linewidth}
\includegraphics[width=\linewidth]{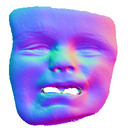}
\centering
\captionsetup{labelformat=empty}
\caption{\scriptsize Pix2V}
\end{subfigure}
\begin{subfigure}{\sze\linewidth}
\includegraphics[width=\linewidth]{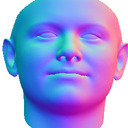}
\centering
\captionsetup{labelformat=empty}
\caption{\scriptsize 3DDFA}
\end{subfigure}

\end{figure*}
%
%
\begin{figure*}[h!]\ContinuedFloat
\centering

\begin{subfigure}{\sze\linewidth}
\includegraphics[width=\linewidth]{figures/im2/24}
\centering
\end{subfigure}
\begin{subfigure}{\sze\linewidth}
\includegraphics[width=\linewidth]{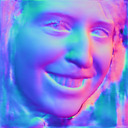}
\centering
\end{subfigure}
\begin{subfigure}{\sze\linewidth}
\includegraphics[width=\linewidth]{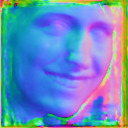}
\centering
\end{subfigure}
\begin{subfigure}{\sze\linewidth}
\includegraphics[width=\linewidth]{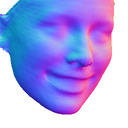}
\centering
\end{subfigure}
\begin{subfigure}{\sze\linewidth}
\includegraphics[width=\linewidth]{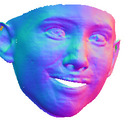}
\centering
\end{subfigure}
\begin{subfigure}{\sze\linewidth}
\includegraphics[width=\linewidth]{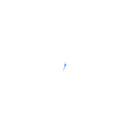}
\centering
\end{subfigure}
\begin{subfigure}{\sze\linewidth}
\includegraphics[width=\linewidth]{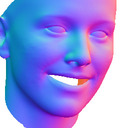}
\centering
\end{subfigure}

\begin{subfigure}{\sze\linewidth}
\includegraphics[width=\linewidth]{figures/im2/52}
\centering
\end{subfigure}
\begin{subfigure}{\sze\linewidth}
\includegraphics[width=\linewidth]{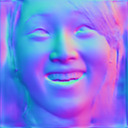}
\centering
\end{subfigure}
\begin{subfigure}{\sze\linewidth}
\includegraphics[width=\linewidth]{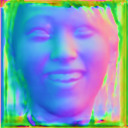}
\centering
\end{subfigure}
\begin{subfigure}{\sze\linewidth}
\includegraphics[width=\linewidth]{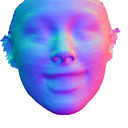}
\centering
\end{subfigure}
\begin{subfigure}{\sze\linewidth}
\includegraphics[width=\linewidth]{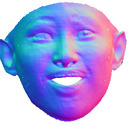}
\centering
\end{subfigure}
\begin{subfigure}{\sze\linewidth}
\includegraphics[width=\linewidth]{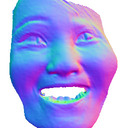}
\centering
\end{subfigure}
\begin{subfigure}{\sze\linewidth}
\includegraphics[width=\linewidth]{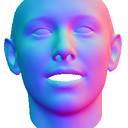}
\centering
\end{subfigure}

\begin{subfigure}{\sze\linewidth}
\includegraphics[width=\linewidth]{figures/im2/5}
\centering
\end{subfigure}
\begin{subfigure}{\sze\linewidth}
\includegraphics[width=\linewidth]{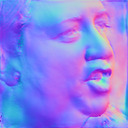}
\centering
\end{subfigure}
\begin{subfigure}{\sze\linewidth}
\includegraphics[width=\linewidth]{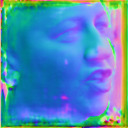}
\centering
\end{subfigure}
\begin{subfigure}{\sze\linewidth}
\includegraphics[width=\linewidth]{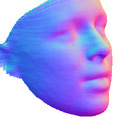}
\centering
\end{subfigure}
\begin{subfigure}{\sze\linewidth}
\includegraphics[width=\linewidth]{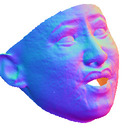}
\centering
\end{subfigure}
\begin{subfigure}{\sze\linewidth}
\includegraphics[width=\linewidth]{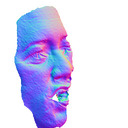}
\centering
\end{subfigure}
\begin{subfigure}{\sze\linewidth}
\includegraphics[width=\linewidth]{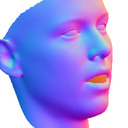}
\centering
\end{subfigure}

\begin{subfigure}{\sze\linewidth}
\includegraphics[width=\linewidth]{figures/im2/7}
\centering
\end{subfigure}
\begin{subfigure}{\sze\linewidth}
\includegraphics[width=\linewidth]{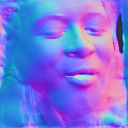}
\centering
\end{subfigure}
\begin{subfigure}{\sze\linewidth}
\includegraphics[width=\linewidth]{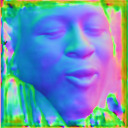}
\centering
\end{subfigure}
\begin{subfigure}{\sze\linewidth}
\includegraphics[width=\linewidth]{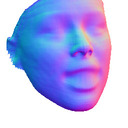}
\centering
\end{subfigure}
\begin{subfigure}{\sze\linewidth}
\includegraphics[width=\linewidth]{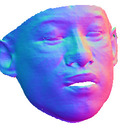}
\centering
\end{subfigure}
\begin{subfigure}{\sze\linewidth}
\includegraphics[width=\linewidth]{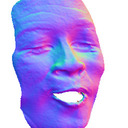}
\centering
\end{subfigure}
\begin{subfigure}{\sze\linewidth}
\includegraphics[width=\linewidth]{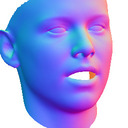}
\centering
\end{subfigure}

\begin{subfigure}{\sze\linewidth}
\includegraphics[width=\linewidth]{figures/im2/33}
\centering
\end{subfigure}
\begin{subfigure}{\sze\linewidth}
\includegraphics[width=\linewidth]{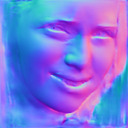}
\centering
\end{subfigure}
\begin{subfigure}{\sze\linewidth}
\includegraphics[width=\linewidth]{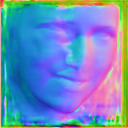}
\centering
\end{subfigure}
\begin{subfigure}{\sze\linewidth}
\includegraphics[width=\linewidth]{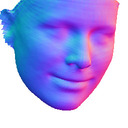}
\centering
\end{subfigure}
\begin{subfigure}{\sze\linewidth}
\includegraphics[width=\linewidth]{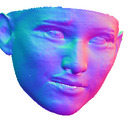}
\centering
\end{subfigure}
\begin{subfigure}{\sze\linewidth}
\includegraphics[width=\linewidth]{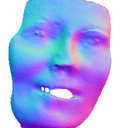}
\centering
\end{subfigure}
\begin{subfigure}{\sze\linewidth}
\includegraphics[width=\linewidth]{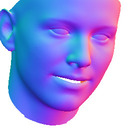}
\centering
\end{subfigure}

\begin{subfigure}{\sze\linewidth}
\includegraphics[width=\linewidth]{figures/im2/35}
\centering
\end{subfigure}
\begin{subfigure}{\sze\linewidth}
\includegraphics[width=\linewidth]{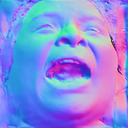}
\centering
\end{subfigure}
\begin{subfigure}{\sze\linewidth}
\includegraphics[width=\linewidth]{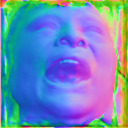}
\centering
\end{subfigure}
\begin{subfigure}{\sze\linewidth}
\includegraphics[width=\linewidth]{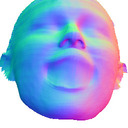}
\centering
\end{subfigure}
\begin{subfigure}{\sze\linewidth}
\includegraphics[width=\linewidth]{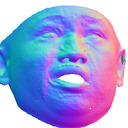}
\centering
\end{subfigure}
\begin{subfigure}{\sze\linewidth}
\includegraphics[width=\linewidth]{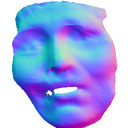}
\centering
\end{subfigure}
\begin{subfigure}{\sze\linewidth}
\includegraphics[width=\linewidth]{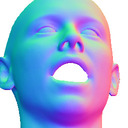}
\centering
\end{subfigure}

\begin{subfigure}{\sze\linewidth}
\includegraphics[width=\linewidth]{figures/im/11}
\centering
\end{subfigure}
\begin{subfigure}{\sze\linewidth}
\includegraphics[width=\linewidth]{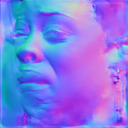}
\centering
\end{subfigure}
\begin{subfigure}{\sze\linewidth}
\includegraphics[width=\linewidth]{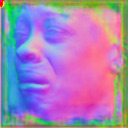}
\centering
\end{subfigure}
\begin{subfigure}{\sze\linewidth}
\includegraphics[width=\linewidth]{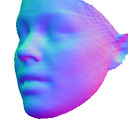}
\centering
\end{subfigure}
\begin{subfigure}{\sze\linewidth}
\includegraphics[width=\linewidth]{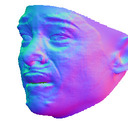}
\centering
\end{subfigure}
\begin{subfigure}{\sze\linewidth}
\includegraphics[width=\linewidth]{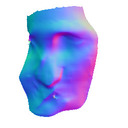}
\centering
\end{subfigure}
\begin{subfigure}{\sze\linewidth}
\includegraphics[width=\linewidth]{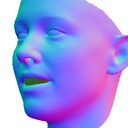}
\centering
\end{subfigure}

\begin{subfigure}{\sze\linewidth}
\includegraphics[width=\linewidth]{figures/im2/46}
\centering
\end{subfigure}
\begin{subfigure}{\sze\linewidth}
\includegraphics[width=\linewidth]{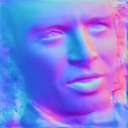}
\centering
\end{subfigure}
\begin{subfigure}{\sze\linewidth}
\includegraphics[width=\linewidth]{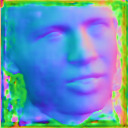}
\centering
\end{subfigure}
\begin{subfigure}{\sze\linewidth}
\includegraphics[width=\linewidth]{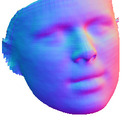}
\centering
\end{subfigure}
\begin{subfigure}{\sze\linewidth}
\includegraphics[width=\linewidth]{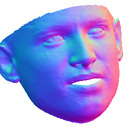}
\centering
\end{subfigure}
\begin{subfigure}{\sze\linewidth}
\includegraphics[width=\linewidth]{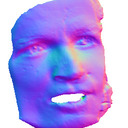}
\centering
\end{subfigure}
\begin{subfigure}{\sze\linewidth}
\includegraphics[width=\linewidth]{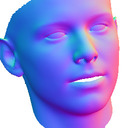}
\centering
\end{subfigure}

\begin{subfigure}{\sze\linewidth}
\includegraphics[width=\linewidth]{figures/im2/60}
\centering
\end{subfigure}
\begin{subfigure}{\sze\linewidth}
\includegraphics[width=\linewidth]{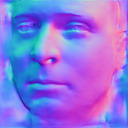}
\centering
\end{subfigure}
\begin{subfigure}{\sze\linewidth}
\includegraphics[width=\linewidth]{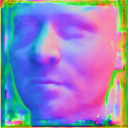}
\centering
\end{subfigure}
\begin{subfigure}{\sze\linewidth}
\includegraphics[width=\linewidth]{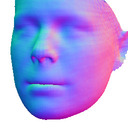}
\centering
\end{subfigure}
\begin{subfigure}{\sze\linewidth}
\includegraphics[width=\linewidth]{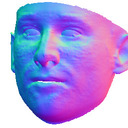}
\centering
\end{subfigure}
\begin{subfigure}{\sze\linewidth}
\includegraphics[width=\linewidth]{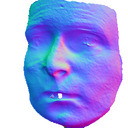}
\centering
\end{subfigure}
\begin{subfigure}{\sze\linewidth}
\includegraphics[width=\linewidth]{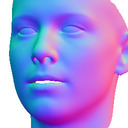}
\centering
\end{subfigure}

\begin{subfigure}{\sze\linewidth}
\includegraphics[width=\linewidth]{figures/im2/62}
\centering
\end{subfigure}
\begin{subfigure}{\sze\linewidth}
\includegraphics[width=\linewidth]{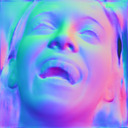}
\centering
\end{subfigure}
\begin{subfigure}{\sze\linewidth}
\includegraphics[width=\linewidth]{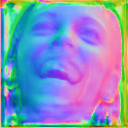}
\centering
\end{subfigure}
\begin{subfigure}{\sze\linewidth}
\includegraphics[width=\linewidth]{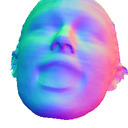}
\centering
\end{subfigure}
\begin{subfigure}{\sze\linewidth}
\includegraphics[width=\linewidth]{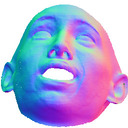}
\centering
\end{subfigure}
\begin{subfigure}{\sze\linewidth}
\includegraphics[width=\linewidth]{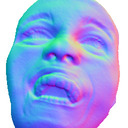}
\centering
\end{subfigure}
\begin{subfigure}{\sze\linewidth}
\includegraphics[width=\linewidth]{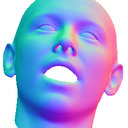}
\centering
\end{subfigure}

\begin{subfigure}{\sze\linewidth}
\includegraphics[width=\linewidth]{figures/im2/63}
\centering
\captionsetup{labelformat=empty}
\caption{\scriptsize Input}
\end{subfigure}
\begin{subfigure}{\sze\linewidth}
\includegraphics[width=\linewidth]{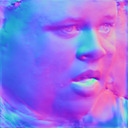}
\centering
\captionsetup{labelformat=empty}
\caption{\scriptsize Ours+PRN}
\end{subfigure}
\begin{subfigure}{\sze\linewidth}
\includegraphics[width=\linewidth]{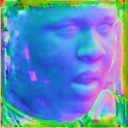}
\centering
\captionsetup{labelformat=empty}
\caption{\scriptsize SfSNet+PRN}
\end{subfigure}
\begin{subfigure}{\sze\linewidth}
\includegraphics[width=\linewidth]{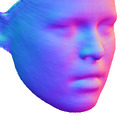}
\centering
\captionsetup{labelformat=empty}
\caption{\scriptsize PRN}
\end{subfigure}
\begin{subfigure}{\sze\linewidth}
\includegraphics[width=\linewidth]{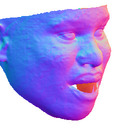}
\centering
\captionsetup{labelformat=empty}
\caption{\scriptsize Extreme}
\end{subfigure}
\begin{subfigure}{\sze\linewidth}
\includegraphics[width=\linewidth]{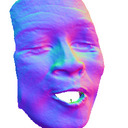}
\centering
\captionsetup{labelformat=empty}
\caption{\scriptsize Pix2V}
\end{subfigure}
\begin{subfigure}{\sze\linewidth}
\includegraphics[width=\linewidth]{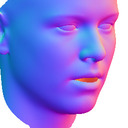}
\centering
\captionsetup{labelformat=empty}
\caption{\scriptsize 3DDFA}
\end{subfigure}

\caption{Qualitative comparisons with normals in the 300-W dataset~\cite{Sagonas13}.}
\label{fig:comparison_norm2}
\end{figure*}


\endgroup

\end{document}